%% file: solar_neurips2026.tex
\documentclass{article}

\PassOptionsToPackage{numbers,compress}{natbib}
\usepackage[main,final]{neurips_2026}

\usepackage[utf8]{inputenc}
\usepackage[T1]{fontenc}
\usepackage{hyperref}
\hypersetup{hidelinks}
\usepackage{url}
\usepackage{booktabs}
\usepackage{amsfonts}
\usepackage{nicefrac}
\usepackage{microtype}
\usepackage{xcolor}
\usepackage{graphicx}
\usepackage{amsmath}
\usepackage{wrapfig}
\usepackage{amssymb}
\usepackage{mathrsfs}
\usepackage{latexsym}

\IfFileExists{newtxmath.sty}{\usepackage{newtxmath}}{}

\makeatletter
\newcommand{\mathbbm}[1]{\mathbf{#1}}
\makeatother

\usepackage{bm}

\usepackage{colortbl}
\usepackage{multirow}
\usepackage{makecell}
\usepackage{multicol}
\usepackage{pifont}
\usepackage{caption}

\usepackage{tcolorbox}
\tcbuselibrary{breakable, skins}
\tcbuselibrary{skins}

\usepackage{CJKutf8}

\definecolor{lightpink}{RGB}{255,182,193}
\definecolor{lightgreen}{RGB}{144,238,144}
\definecolor{verylightgray}{RGB}{220,220,220}
\definecolor{rankone}{RGB}{181,224,210}
\definecolor{ranktwo}{RGB}{250,228,218}
\newcommand{\best}[1]{\cellcolor{rankone}{#1}}
\newcommand{\sbest}[1]{\cellcolor{ranktwo}{#1}}

\newif\ifshowstd
\showstdfalse

\definecolor{redinfigure}{RGB}{196, 86, 54}

\usepackage{pgf}
\definecolor{gradgreen}{RGB}{98,190,160}
\definecolor{gradred}{RGB}{235,155,120}
\newcommand{\maxdeltaval}{0.75}%
\newcommand{\goodc}[1]{%
  \pgfmathtruncatemacro{\gvpctint}{round(min(sqrt(abs(#1)/\maxdeltaval)*100,100))}%
  \ifnum\gvpctint>0\relax
    \begingroup\edef\x{\endgroup\noexpand\cellcolor{gradgreen!\gvpctint!white}}\x
  \fi}
\newcommand{\badc}[1]{%
  \pgfmathtruncatemacro{\gvpctint}{round(min(sqrt(abs(#1)/\maxdeltaval)*100,100))}%
  \ifnum\gvpctint>0\relax
    \begingroup\edef\x{\endgroup\noexpand\cellcolor{gradred!\gvpctint!white}}\x
  \fi}
\newcommand{\dgood}[1]{\goodc{#1}#1}
\newcommand{\dbad}[1]{\badc{#1}#1}

\title{Soft Token Alignment for Cross-Lingual Reasoning}

\author{%
  Jiayi He \quad Jungsoo Park \quad Wei Xu \quad Alan Ritter\\
  Georgia Institute of Technology\\
  \texttt{jhe478@gatech.edu}\\
}

\begin{document}
\makeatletter
\setlength{\@fptop}{0pt}
\setlength{\@fpsep}{12pt}
\makeatother
\maketitle

\input{sections/0-abstract}
\input{sections/1-introduction}

\begin{figure*}[t]
\centering
\includegraphics[width=\textwidth]{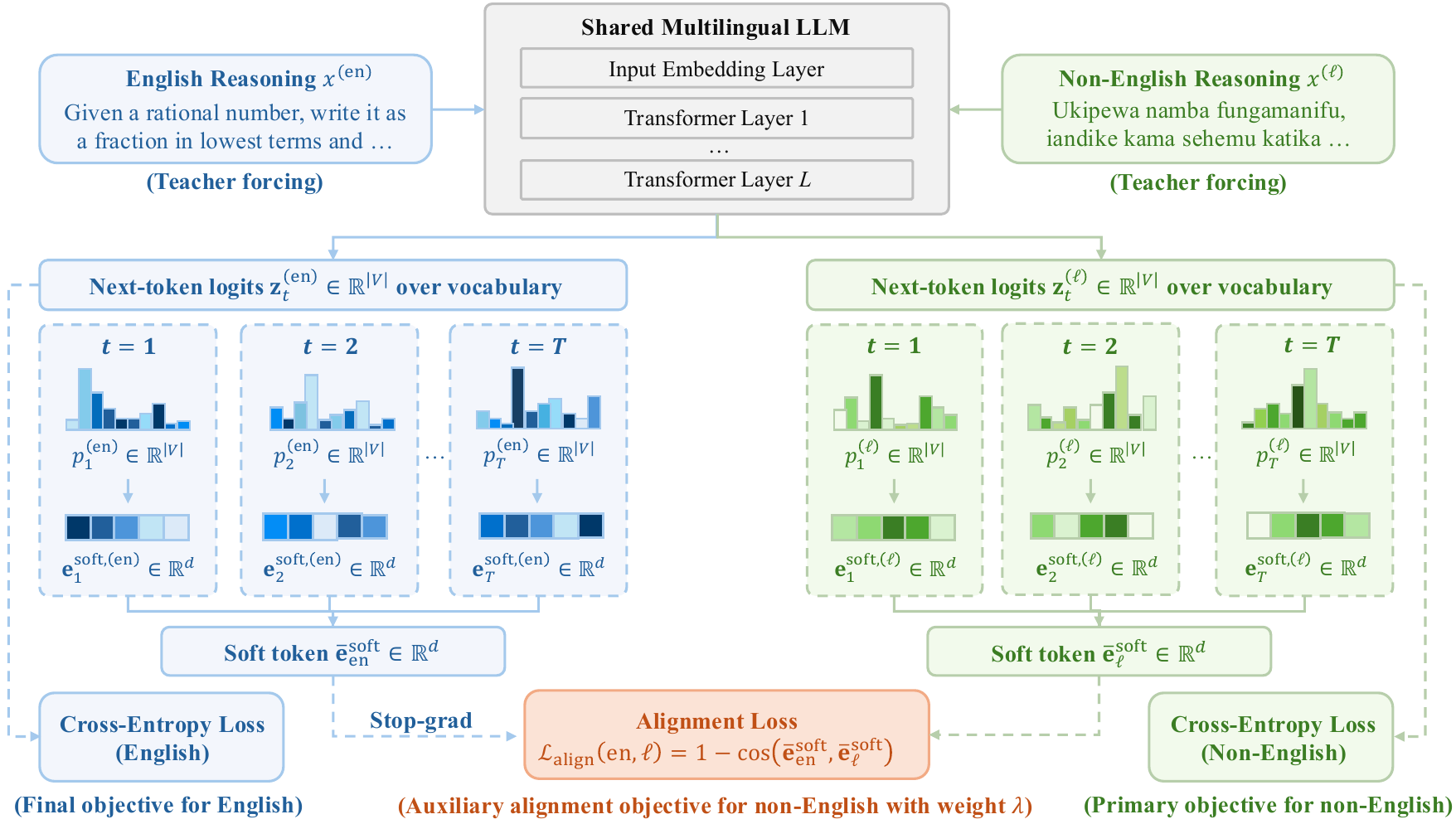}
\caption{SOLAR framework. Parallel English and non-English inputs are processed by a shared LLM under teacher forcing, producing a per-position soft token $\mathbf{e}^{\,\text{soft}}_t$ that mixes vocabulary embeddings under the model's next-token distribution. Soft tokens are mean-pooled across response positions into a per-language summary vector ($\bar{\mathbf{e}}^{\,\text{soft}}_{\text{en}}$, $\bar{\mathbf{e}}^{\,\text{soft}}_{\ell}$). The alignment objective $\mathcal{L}_{\text{align}}$ is the cosine distance between the English and non-English summaries; stop-gradient on the English side restricts alignment gradients to the non-English branch. Both branches are supervised by cross-entropy ($\mathcal{L}^{(\text{en})}_{\text{CE}}$, $\mathcal{L}^{(\ell)}_{\text{CE}}$), and the non-English objective adds $\mathcal{L}_{\text{align}}$ weighted by $\lambda > 0$, which controls alignment strength.}
\label{fig:solar}
\end{figure*}

\section{SOLAR: soft token alignment for cross-lingual reasoning}
In this section, we first review discrete and soft thinking as two paradigms for constructing input representations during autoregressive reasoning
(\S\ref{sec:background}), then describe how SOLAR repurposes soft-token embeddings as a cross-lingual alignment signal during SFT (\S\ref{sec:alignment}).

\subsection{Background: discrete vs. soft thinking}\label{sec:background}

In Chain-of-Thought (CoT)~\citep{NEURIPS2022_9d560961}, which we call \textbf{Discrete Thinking}, reasoning proceeds by autoregressively generating discrete tokens. 
Let $x_{1:n}$ denote the input prompt tokens and $\hat{x}_{n+1:t-1}$ the previously generated reasoning tokens. Let $\mathbf{E}[\cdot]$ denote the token embedding lookup function, with embeddings $\mathbf{e}_i=\mathbf{E}[x_i]$ for $i\le n$ and $\mathbf{e}_i=\mathbf{E}[\hat{x}_i]$ for $i>n$.
The next token at step $t$ is decoded from $p_\theta(\cdot\mid \mathbf{e}_{1:t-1})$ (\textit{e.g.,} via greedy decoding or sampling):
\begin{equation}
\hat{x}_t \sim p_\theta(\cdot \mid \mathbf{e}_{1:t-1}),
\qquad
\mathbf{e}^{\,\text{discrete}}_t = \mathbf{E}[\hat{x}_t].
\label{eq:discrete}
\end{equation}

For a language model $f_\theta$ with vocabulary $V$ and token embedding matrix $\mathbf{E} \in \mathbb{R}^{|V| \times d}$, the logit vector $\mathbf{z}_t = f_\theta(\mathbf{e}_{1:t-1}) \in \mathbb{R}^{|V|}$ induces a distribution $p_\theta(\cdot \mid \mathbf{e}_{1:t-1})=\operatorname{softmax}(\mathbf{z}_t / \tau)$, where $\tau$ is the temperature. Eq.~\ref{eq:discrete} thus constitutes a hard projection from the distribution over the vocabulary $V$ to the embedding of a single discrete token $\hat{x}_t \in V$. That is, regardless of how probability mass is distributed, only $\mathbf{E}[\hat{x}_t]$ is fed as the input representation for the next step.

\textbf{Soft Thinking}~\cite{zhang2025soft} is an inference-time, autoregressive method that replaces this hard discrete projection during reasoning with a continuous representation formed from the next-token distribution. At step $t$, it defines a \emph{soft-token} embedding as the expectation over token embeddings $\mathbf{E}[v]$ for $v \in V$ under a truncated-and-renormalized distribution:
\begin{equation}
    \mathbf{e}^{\,\text{soft}}_t = \sum_{v \in V} \tilde{p}_\theta(v \mid \mathbf{e}_{1:t-1}) \cdot \mathbf{E}[v],
    \label{eq:soft-embed}
\end{equation}
where $\tilde{p}_\theta$ is constructed by applying top-$k$ filtering to the original distribution $p_\theta(v \mid \mathbf{e}_{1:t-1})=\left[\operatorname{softmax}\!\left(\mathbf{z}_t/\tau\right)\right]_v$. Specifically, let $F_{t} \subseteq V$ be the token
set obtained by retaining the $k$ highest-probability tokens.
The filtered distribution is
\begin{equation}
    \tilde{p}_\theta(v \mid \mathbf{e}_{1:t-1})=\frac{p_\theta(v \mid \mathbf{e}_{1:t-1})\mathbbm{1}[v \in F_{t}]}{\sum_{v' \in F_{t}}p_\theta(v' \mid \mathbf{e}_{1:t-1})}.
    \label{eq:filtered-prob}
\end{equation}
The resulting $\mathbf{e}^{\,\text{soft}}_t$ corresponds to no single discrete token and is fed back as the input embedding at step $t{+}1$, closing an autoregressive loop over continuous representations instead of discrete tokens.

\subsection{Aligning soft tokens for cross-lingual consistency}
\label{sec:alignment}

\paragraph{From inference-time decoding to training-time alignment.}
Soft tokens were originally introduced as a single-language, inference-time enhancement to CoT~\citep{zhang2025soft}, where they are fed back autoregressively during decoding.
For cross-lingual alignment, soft tokens mix shared vocabulary embeddings rather than committing to a single language-specific token or script.
Inference-only soft thinking can soften decoding decisions, but does not explicitly train semantically equivalent reasoning paths across languages to stay aligned.
SOLAR instead uses soft tokens as a training-time alignment signal during SFT: the forward pass stays teacher-forced on discrete reference tokens, soft tokens are used only to compute an auxiliary alignment loss, and non-English soft-token summaries are pulled toward an English pivot in the shared embedding space.

\paragraph{Soft-token summary and alignment.}
Given a mini-batch of parallel input-response pairs ${(x^{(\ell)}, y^{(\ell)})}_{\ell \in \mathcal{L}}$  across a set of languages $\mathcal{L}$ (with English $\text{en}\in\mathcal{L}$), a teacher-forced forward pass produces logits $\mathbf{z}_t$ at each response position $t \in \{1,\dots,T\}$.
After top-$k$ filtering and temperature scaling, $\tilde{p}_\theta(v \mid \mathbf{e}_{1:t-1})$ defines the soft-token embedding $\mathbf{e}^{\,\text{soft}}_t$ via Eq.~\ref{eq:soft-embed}.
We mean-pool these soft tokens into a summary vector $\bar{\mathbf{e}}^{\,\text{soft}}_\ell = \frac{1}{T}\sum_{t=1}^{T}\mathbf{e}^{\,\text{soft}}_t \in \mathbb{R}^d$.
The pooling range $t \in \{1,\dots,T\}$ covers \emph{all supervised response positions} and excludes prompt positions.
Unlike a discrete token embedding, this vector summarizes likely continuations in the shared embedding space instead of a single vocabulary item or a distribution over $\mathbb{R}^{|V|}$.
Responses can be compared across languages with disjoint vocabularies: SOLAR aligns \emph{semantic mixtures of embeddings}, not raw token-distribution support.

\paragraph{Cross-lingual alignment objective.}
Given a parallel training pair consisting of responses in two different languages~$\ell^*$ and~$\ell$, we let $\bar{\mathbf{e}}^{\,\text{soft}}_{\ell^*}$ and $\bar{\mathbf{e}}^{\,\text{soft}}_{\ell}$ denote the mean-pooled soft-token embeddings of the pivot and target responses. To encourage their response-level soft-token summaries to be close in the shared embedding space, we minimize their discrepancy using cosine distance:
\begin{equation}
    \mathcal{L}_{\text{align}}(\ell^*, \ell) = 1 - \cos\!\left(\bar{\mathbf{e}}^{\,\text{soft}}_{\ell^*},\; \bar{\mathbf{e}}^{\,\text{soft}}_{\ell}\right),
    \label{eq:pair-loss}
\end{equation}
where $\ell^*$ denotes the pivot language.
We form $(\ell^*,\ell)$ pairs for every non-pivot language $\ell$, with the pivot summary $\bar{\mathbf{e}}^{\,\text{soft}}_{\ell^*}$ treated as a fixed target so that alignment gradients flow exclusively through $\bar{\mathbf{e}}^{\,\text{soft}}_{\ell}$. This pulls non-pivot representations toward the pivot.
We set $\ell^* = \text{en}$ (English) in all main experiments, since English yields the highest MGSM accuracy for every model (Table~\ref{tab:mgsm-per-lang-main}).

\paragraph{Integration with supervised fine-tuning.}
In the SFT setting, the alignment objective $\mathcal{L}_{\text{align}}$ is added as an auxiliary loss for each non-English training sample
in language~$\ell$:
\begin{equation}
    \mathcal{L}^{(\ell)}_{\text{SFT}} = \mathcal{L}^{(\ell)}_{\text{CE}} + \lambda \cdot \mathcal{L}_{\text{align}}(\ell^*, \ell),
    \label{eq:sft-loss}
\end{equation}
where $\mathcal{L}^{(\ell)}_{\text{CE}}$ is the cross-entropy loss for the response in language~$\ell$, $\lambda > 0$ controls alignment strength, and soft tokens are computed from the same teacher-forced forward pass used to evaluate $\mathcal{L}_{\text{CE}}$.
For English samples, the alignment term is omitted, leaving $\mathcal{L}^{(\text{en})}_{\text{SFT}} = \mathcal{L}^{(\text{en})}_{\text{CE}}$.

\input{sections/4-experiments}
\input{sections/5-analysis}
\input{sections/6-related-work}
\input{sections/7-conclusion}

\begin{ack}
This research is supported in part by the NSF under grant numbers IIS-2144493, IIS-2052498, SMA-2418946, and NAIRR250217, in addition to a gift from Google. Any opinions, findings, and conclusions or recommendations expressed in this material are those of the author(s) and do not necessarily reflect the views of the National Science Foundation.
\end{ack}

\bibliographystyle{plainnat}
\bibliography{references}

\appendix

\input{appendix/0-limitations-impacts}
\input{appendix/1-related-work}
\input{appendix/2-implementation-details}
\input{appendix/3-additional-results}
\input{appendix/4-additional-analysis}

\clearpage
\input{checklist}

\end{document}

%% file: sections/0-abstract.tex
\begin{abstract}
Multilingual large language models often produce inconsistent reasoning and answers for semantically equivalent prompts in different languages. 
Prior work suggests that intermediate representations can be relatively language-agnostic, but generation becomes increasingly language-specific as models commit to discrete output tokens.
This is problematic because language-specific lexical choices can cause semantically equivalent reasoning paths to diverge across languages.
These divergences motivate searching for a cross-lingual alignment signal that is less tied to any single vocabulary item or script.
We propose \textbf{SOLAR}, an auxiliary objective for supervised fine-tuning that aligns \textit{soft-token} representations across languages, using English as a pivot. Soft tokens are probability-weighted mixtures over the vocabulary embeddings, yielding continuous representations that can aggregate information from semantically related tokens across languages.
We then align each non-English soft-token summary to its English counterpart in the shared embedding space.
Across four multilingual reasoning benchmarks, SOLAR improves accuracy by up to +17.7 points over the base model and +3.8 over standard supervised fine-tuning, with the largest gains on low-resource languages. SOLAR also strengthens final-layer cross-lingual similarity and substantially reduces language-cluster separability, suggesting that aligning soft-token representations helps preserve shared semantic structure during multilingual reasoning.
\end{abstract}

%% file: sections/1-introduction.tex
\section{Introduction}\label{sec:introduction}
Large language models (LLMs) increasingly serve multilingual users, yet often produce inconsistent reasoning and answers for semantically equivalent prompts in different languages~\cite{qi-etal-2023-cross,zhao2025comprehensiveevaluationmultilingualchainofthought}. Recent work studying the internal representations of multilingual models~\citep{wang-etal-2025-lost-multilinguality,lim2025languagespecific,pokharel2026crosslingualactivationsteeringmultilingual} reveals a consistent mechanism: intermediate Transformer layers encode largely language-agnostic semantics, while final layers become language-specialized as the model selects discrete tokens in the target language's script (\textit{e.g.,} Chinese characters or Thai glyphs) for generation. This late-stage divergence from shared semantics to language-specific surface forms is hypothesized to contribute to inconsistent reasoning across languages, even when intermediate representations may have converged on the same solution.

The transition from final-layer representations to discrete tokens can act as a bottleneck, collapsing probability mass over semantically related tokens into a single language-specific choice.
Continuous-space reasoning offers a way to soften this collapse: \textit{soft thinking}~\citep{zhang2025soft} replaces discrete next-token choices with probability-weighted mixtures over vocabulary embeddings. The resulting \textit{soft tokens} aggregate information across the vocabulary and are fed back as input embeddings for the next autoregressive step.
Because soft tokens are not tied to a single token or language, they may retain cross-lingual semantic content that discrete token selection discards.
However, prior soft-thinking methods use soft tokens mainly as an inference-time generation mechanism. They soften discrete decoding decisions but do not directly encourage semantically equivalent reasoning paths in different languages to stay aligned.
This gap motivates our hypothesis: \textit{soft-token representations can provide a better training-time signal for cross-lingual alignment than discrete tokens.}

We propose \textbf{SOLAR} (\underline{\textbf{So}}ft Token A\underline{\textbf{l}}ignment for Cross-Lingu\underline{\textbf{a}}l \underline{\textbf{R}}easoning), a lightweight auxiliary fine-tuning objective that aligns soft-token representations between English and non-English parallel reasoning traces during supervised fine-tuning (SFT) (Figure~\ref{fig:solar}). 
For each reasoning trace and final response, SOLAR summarizes the model's next-token distribution as an expectation over the embedding matrix. It then mean-pools across positions to obtain a single continuous vector that is independent of any individual token identity.
Using English as a pivot, SOLAR minimizes cosine distance between the English representation and each non-English counterpart, drawing all languages toward an English-anchored semantic space. Because soft tokens operate in the shared embedding space, they enable direct comparison across languages regardless of surface-form differences.

Our experiments show that SOLAR improves both task accuracy and cross-lingual consistency across four multilingual reasoning benchmarks. 
For Qwen3-4B~\cite{yang2025qwen3technicalreport}, SOLAR improves accuracy by up to +3.8 over SFT and cross-lingual consistency by up to +4.5 on AIME~2024, with the largest gains on low-resource languages such as Swahili. 
Representation analysis shows that soft tokens carry a stronger final-layer cross-lingual signal than discrete tokens on base models, and that SOLAR produces substantially stronger final-layer cross-lingual alignment than SFT and inference-time soft thinking alone. 
At the same time, behavioral analysis shows that these gains preserve native-language reasoning: on MGSM, SOLAR-tuned Qwen3-4B reasons in the target language's script at 98.13\% (vs.\ 98.81\% for the base), and SOLAR-tuned Qwen3-8B at 98.16\%. This reverses the base 8B model's collapse to English in non-Chinese languages (Ja/Th/Te below 5\%; 27.43\% four-language average).
To our knowledge, SOLAR is the first method to leverage soft tokens, originally introduced for inference-time continuous reasoning, in a cross-lingual alignment objective during SFT.

%% file: sections/4-experiments.tex
\section{Experiments}\label{sec:experiments}
We evaluate SOLAR on multilingual reasoning benchmarks to test whether soft-token alignment improves both accuracy and cross-lingual consistency. Further details are in Appendix~\ref{appendix:preliminary_implementation_details}.

\subsection{Experimental setup}
\paragraph{Training.}
We apply supervised fine-tuning to Qwen3-4B/8B and Gemma3-4B/12B~\cite{gemmateam2025gemma3technicalreport} using M-s1k~\cite{barua2025longchainofthoughtreasoninglanguages}.
M-s1k is a multilingual long CoT reasoning dataset translated from s1k~\cite{muennighoff-etal-2025-s1} by Gemini-2.0-Flash~\cite{google_gemini2flash}.  
s1k consists of 1,000 diverse, high-quality, and challenging questions paired with complex reasoning traces and solutions. 
We compute soft tokens with $\tau{=}1.0$ and top-$k{=}30$. 

\paragraph{Evaluation.}
We evaluate on MGSM~\cite{shi2023language} and three XReasoning benchmarks~\cite{qi-etal-2025-models}: AIME~2024~\cite{aime2024}, AIME~2025, and GPQA~Diamond~\cite{rein2024gpqa} (denoted GPQA hereafter).
Each benchmark covers 11 languages; we report the 7 overlapping with M-s1k: En, Fr, Ja, Sw, Te, Th, and Zh.
Following~\citet{zhang2025soft}, we decode with temperature $0.6$, $k{=}30$, $p{=}0.95$, and a maximum length of 32,768 tokens.
To reduce variance, we average over 5 runs for MGSM and GPQA and 10 for AIME.

\paragraph{Metrics.} We report (1) Task Accuracy ($\uparrow$): the proportion of questions answered correctly; and (2) Cross-Lingual Consistency (CLC $\uparrow$): the proportion of questions answered correctly in both English and a paired non-English language. We define CLC for language pair $(\ell_1, \ell_2)$ as $\mathrm{CLC}(\ell_1, \ell_2) = \frac{1}{N_q} \sum_{i=1}^{N_q} \mathbbm{1}\!\left[a_i^{\ell_1} = a_i^{\ell_2} = a_i^*\right]$, where $a_i^{\ell}$ is the model's answer for the $i$-th question in language $\ell$, $a_i^*$ is the ground-truth answer, and $N_q$ is the total number of questions.

\paragraph{Baselines.}
We compare SOLAR with several approaches: (1) base models with standard autoregressive decoding; (2) Soft Thinking~\cite{zhang2025soft}: inference-time replacement of discrete tokens with soft-token embeddings; (3) SFT: standard supervised fine-tuning without SOLAR's alignment objective; (4) MidAlign~\cite{liu-niehues-2025-middle}: a middle-layer representation alignment method based on contrastive learning; (5) MAPO~\cite{she-etal-2024-mapo}: a preference optimization method aligning non-English reasoning with English via translation model scores; and (6) AlignX~\cite{bu-etal-2025-alignx}: a two-stage framework combining intermediate-layer contrastive alignment with a language-matching classifier.
On MGSM, we evaluate English-pivot SFT on Qwen3-4B, pairing non-English questions with parallel English reasoning traces.

\begin{table*}[!t]
\centering
\caption{Average accuracy and cross-lingual consistency (\%) across four benchmarks for Qwen3-4B, Qwen3-8B, and their fine-tuned variants. \colorbox{rankone}{Best} and \colorbox{ranktwo}{second-best} within each model are highlighted.}
\label{tab:main-results}
\resizebox{\textwidth}{!}{%
\begin{tabular}{lcccccccccc}
\toprule
& \multicolumn{2}{c}{\textbf{MGSM}} & \multicolumn{2}{c}{\textbf{AIME 2024}} & \multicolumn{2}{c}{\textbf{AIME 2025}} & \multicolumn{2}{c}{\textbf{GPQA}} & \multicolumn{2}{c}{\textbf{Avg.}}\\
\cmidrule(lr){2-3} \cmidrule(lr){4-5} \cmidrule(lr){6-7} \cmidrule(lr){8-9} \cmidrule(lr){10-11}
\textbf{Model} & \textbf{Accuracy} & \textbf{CLC} & \textbf{Accuracy} & \textbf{CLC} & \textbf{Accuracy} & \textbf{CLC} & \textbf{Accuracy} & \textbf{CLC} & \textbf{Accuracy} & \textbf{CLC}\\
\midrule
Qwen3-4B         & 74.01 & 69.31 & 31.10 & 23.89 & 23.05 & 17.61 & 27.52 & 14.56 & 38.92 & 31.34\\
~~+ Soft Thinking & 74.23 & 70.37 & 31.86 & 24.44 & 23.33 & 18.22 & 27.34 & 14.31 & 39.19 & 31.84\\
\addlinespace[2pt]
~~+ SFT          & 76.25 & 72.29 & 34.76 & 25.17 & 27.90 & 20.44 & \sbest{43.75} & \sbest{33.15} & 45.67 & 37.76\\
~~+ MidAlign     & 76.45 & 73.11 & \sbest{36.67} & \sbest{27.11} & \sbest{29.05} & \sbest{23.22} & 43.00 & 32.47 & \sbest{46.29} & \sbest{38.98}\\
~~+ MAPO         & 77.52 & \sbest{73.90} & 34.76 & 25.89 & 25.81 & 18.94 & 40.32 & 30.15 & 44.60 & 37.22\\
~~+ AlignX       & \sbest{77.69} & 73.00 & 36.00 & 26.88 & 27.76 & 20.10 & 43.03 & 32.97 & 46.12 & 38.24\\
~~+ SOLAR & \best{78.27} & \best{75.61} & \best{38.52} & \best{29.67} & \best{30.95} & \best{24.51} & \best{45.24} & \best{36.26} & \best{48.25} & \best{41.51}\\
\midrule
Qwen3-8B    & 83.42 & 80.51 & 62.10 & 56.06 & 54.91 & 45.06 & 52.68 & 42.69 & 63.28 & 56.08\\
~~+ Soft Thinking & 83.04 & 80.01 & 62.52 & 56.91 & 54.71 & 44.22 & 53.06 & 43.42 & 63.33 & 56.14\\
\addlinespace[2pt]
~~+ SFT       & 86.90 & 84.71 & 65.67 & 58.67 & 58.38 & 50.13 & 63.64 & 55.73 & 68.65 & 62.31\\
~~+ MidAlign  & 87.17 & 85.20 & 66.05 & 59.96 & 59.05 & 52.32 & \sbest{63.82} & \sbest{56.48} & 69.02 & 63.49\\
~~+ MAPO      & \sbest{87.43} & \sbest{85.68} & \sbest{66.47} & \sbest{60.01} & \sbest{59.23} & \sbest{52.55} & 63.75 & 56.18 & \sbest{69.22} & \sbest{63.61}\\
~~+ AlignX    & 86.73 & 84.16 & 66.00 & 59.77 & 58.78 & 50.90 & 63.44 & 55.09 & 68.74 & 62.48\\
~~+ SOLAR & \best{88.27} & \best{87.11} & \best{67.52} & \best{60.61} & \best{60.10} & \best{53.02} & \best{64.96} & \best{57.27} & \best{70.21} & \best{64.50}\\
\bottomrule
\end{tabular}
}
\end{table*}

\begin{table*}[!t]
\centering
\caption{Per-language accuracy (\%) and average accuracy on MGSM for Qwen3-4B, Qwen3-8B, and their fine-tuned variants.}
\label{tab:mgsm-per-lang-main}
\footnotesize
\begin{tabular}{lcccccccc}
\toprule
\textbf{Model} & \textbf{En} & \textbf{Fr} & \textbf{Ja} & \textbf{Sw} & \textbf{Te} & \textbf{Th} & \textbf{Zh} & \textbf{Avg.} \\
\midrule
Qwen3-4B & 93.36 & 81.36 & 80.56 & 29.04 & 64.24 & \sbest{83.36} & 86.16 & 74.01 \\
~~+ Soft Thinking & \sbest{93.40} & 83.52 & \best{81.84} & 28.24 & 61.60 & \best{84.12} & 86.92 & 74.23 \\
~~+ SFT & \best{94.00} & 85.60 & 77.20 & 47.20 & 63.52 & 79.36 & 86.88 & 76.25 \\
~~+ English-pivot SFT & 91.20 & 78.32 & 64.16 & 26.24 & 51.76 & 70.40 & 74.96 & 65.29 \\
~~+ MidAlign & 93.24 & 85.16 & 79.92 & 48.72 & 63.04 & 77.60 & 87.44 & 76.45 \\
~~+ MAPO & 92.24 & \sbest{85.76} & 80.00 & \sbest{51.28} & \best{65.92} & 79.04 & \best{88.40} & 77.52 \\
~~+ AlignX & 93.12 & 84.56 & 80.32 & \best{53.48} & 64.32 & 80.40 & 87.60 & \sbest{77.69} \\
~~+ SOLAR & 93.32 & \best{86.32} & \sbest{81.00} & \best{53.48} & \sbest{65.36} & 80.56 & \sbest{87.84} & \best{78.27} \\
\midrule
Qwen3-8B & 94.88 & 86.80 & \best{84.96} & 59.36 & 80.08 & 89.36 & 88.48 & 83.42 \\
~~+ Soft Thinking & \sbest{94.96} & 86.80 & \sbest{84.88} & 58.32 & 81.44 & 88.00 & 86.88 & 83.04 \\
~~+ SFT & \best{95.44} & 89.60 & 82.24 & 79.76 & 80.56 & 91.20 & 89.52 & 86.90 \\
~~+ MidAlign & 94.80 & 90.88 & 82.40 & \sbest{80.80} & 81.20 & 90.64 & 89.44 & 87.17 \\
~~+ MAPO & 94.56 & \sbest{91.04} & 82.16 & 80.00 & 80.96 & \sbest{91.84} & \sbest{91.44} & \sbest{87.43} \\
~~+ AlignX & 94.48 & 89.84 & 82.64 & 79.68 & \sbest{81.60} & 90.32 & 88.56 & 86.73 \\
~~+ SOLAR & \sbest{94.96} & \best{91.28} & 82.72 & \best{81.36} & \best{83.20} & \best{92.80} & \best{91.60} & \best{88.27} \\
\bottomrule
\end{tabular}
\end{table*}

\subsection{Main results}\label{sec:main_results}
Table~\ref{tab:main-results} reports results across four benchmarks, and Table~\ref{tab:mgsm-per-lang-main} breaks down MGSM accuracy by language. Per-language results for AIME~2024, AIME~2025, and GPQA are in Appendix~\ref{appendix:experimental_results}.

\paragraph{SOLAR outperforms every baseline it is evaluated against, at each model size.} For Qwen3-4B, SOLAR improves accuracy over SFT by +1.5 to +3.8 and CLC by +3.1 to +4.5 across all benchmarks. On Qwen3-8B, it achieves +1.3 to +1.9 accuracy and +1.5 to +2.9 consistency gains over SFT. On Qwen3-4B, among cross-lingual alignment baselines, MidAlign is competitive on the harder benchmarks, reaching second-best accuracy on AIME~2024 (36.67\%) and AIME~2025 (29.05\%). MAPO is competitive on MGSM (77.52\%) but underperforms on AIME~2025 and GPQA, likely due to sparse preference pairs: Qwen3-4B answers only 1{,}148 of 9{,}000 non-English M-s1k questions correctly under 20-sample decoding. AlignX is comparable to MidAlign and MAPO (\textit{e.g.,} 77.69\% on MGSM). SOLAR also exceeds all three alignment baselines on Qwen3-8B across all four benchmarks and on Gemma3-4B on MGSM and GPQA (Appendix~\ref{appendix:alignment-baseline-comparison}). On Qwen3-4B, English-pivot SFT yields lower MGSM accuracy and CLC than native-language SFT (Appendix~\ref{appendix:english-pivot-results}). SOLAR retains native-language supervision while aligning soft-token summaries to an English pivot.

\paragraph{Low-resource languages benefit most, with the largest gains on Swahili.} Swahili accuracy on MGSM nearly doubles from 29.04\% to 53.48\% for Qwen3-4B and increases from 59.36\% to 81.36\% for Qwen3-8B, while high-resource languages maintain stable performance (Table~\ref{tab:mgsm-per-lang-main}).
This reflects the intuition that cross-lingual alignment is most helpful where cross-lingual divergence is greatest.

\paragraph{Inference-time soft thinking does not consistently improve cross-lingual reasoning.} For Qwen3-4B, accuracy changes stay within $\pm$0.8 points across all benchmarks, and CLC shifts are inconsistent (+1.06 on MGSM but $-$0.25 on GPQA). This indicates that simply replacing discrete tokens with soft tokens at inference time cannot substitute for explicit training-time alignment as in SOLAR.

%% file: sections/5-analysis.tex
\clearpage
\begin{figure*}[!ht]
  \centering
  \includegraphics[width=1\linewidth]{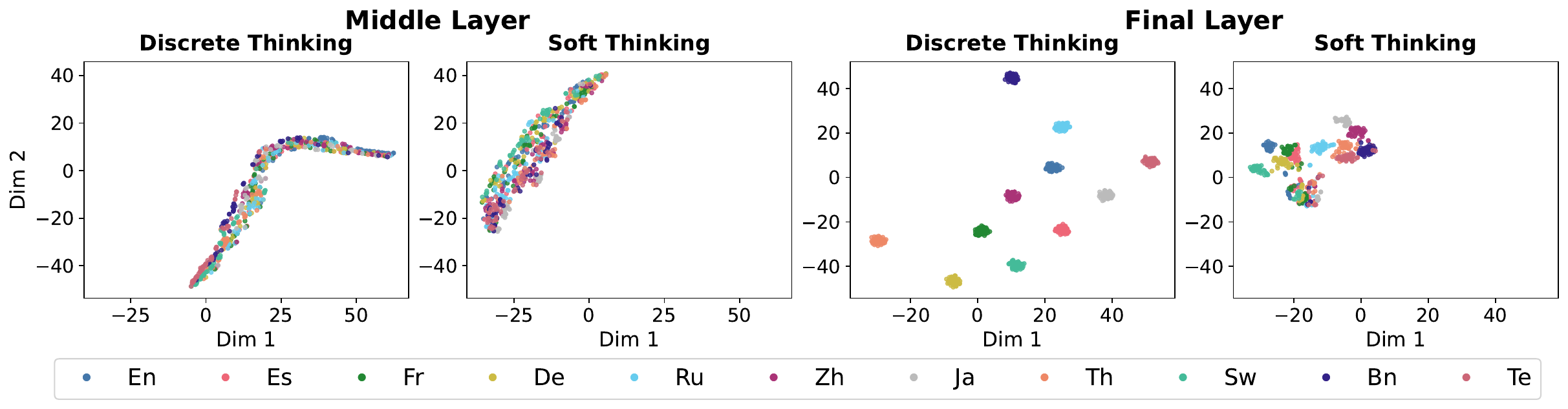}
  \caption{t-SNE of Gemma3-12B hidden states on MGSM at the middle (layer~24) and final (layer~48) layers. Language representations intermix at layer~24 under both thinking modes. At layer~48, discrete thinking yields separated clusters, while soft thinking substantially increases overlap.}
\label{fig:gemma_12b_tsne}
\end{figure*}

\section{Analysis}\label{sec:analysis}

This section analyzes the mechanisms behind SOLAR. We examine how SOLAR reshapes cross-lingual representations (\S\ref{sec:repr-analysis}), verify that SOLAR preserves target-language reasoning without shifting generation toward English (\S\ref{sec:behavioral}), isolate which component of the objective drives the gain (\S\ref{sec:ablation-design}), and analyze sensitivity to the alignment weight and soft-token temperature (\S\ref{sec:ablation}).

\subsection{Representation analysis}\label{sec:repr-analysis}
\begin{wrapfigure}[26]{r}{0.45\linewidth}
  \centering
  \vspace{-0.5cm}
  \includegraphics[width=0.9\linewidth]{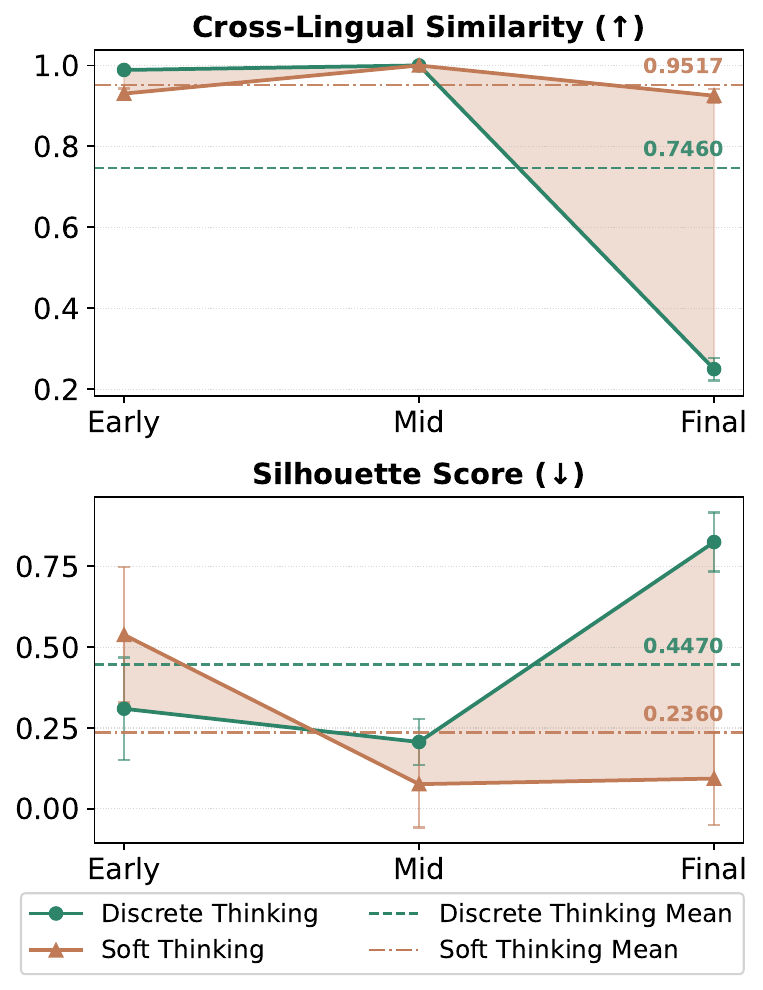}
  \caption{Layer-wise CLS and SS for Gemma3-12B on MGSM under discrete and soft thinking. CLS peaks in the middle layer and drops at the final layer, where soft thinking yields the largest gains.}
  \label{fig:layer-grid-gemma}
\end{wrapfigure}
We examine how hidden representations are organized across languages and how soft tokens affect cross-lingual alignment. To do so, we first quantify the effect of inference-time soft thinking (Table~\ref{tab:pilot_results}), then ask whether SOLAR further reshapes these representations on Qwen3-4B (Table~\ref{tab:repr-after-training}).
Let $\mathbf{h}_i^\ell \in \mathbb{R}^d$ be the hidden state for sample $i$ in language $\ell$, with language set $\mathcal{L}$, $L=|\mathcal{L}|$, and $N$ parallel samples per language.
We report two metrics in $[-1,+1]$: (1) \textbf{Cross-Lingual Similarity (CLS $\uparrow$)}, the average pairwise cosine similarity between parallel samples across all language pairs, where higher values indicate more language-agnostic representations; and (2) \textbf{Silhouette Score (SS $\downarrow$)}~\cite{ROUSSEEUW198753}, the standard cosine-distance silhouette score measuring how strongly hidden states cluster by language, where lower values indicate weaker language separability.
We probe early, middle, and final layers, using layer~1, the 50\%-depth layer, and the last layer for each model.
Metrics are computed on $N{=}50$ parallel MGSM instances across the seven training languages.
For soft-thinking representations, we first run a discrete-token forward pass to construct soft tokens, then run a second pass with these soft tokens as inputs to extract the corresponding hidden states.
Full results for Qwen3, Gemma3, and DeepSeek-R1-Distill~\cite{deepseekai2025deepseekr1incentivizingreasoningcapability} are in Appendix~\ref{appendix:repr-analysis}.

\begin{table*}[!t]
\centering
\caption{Cross-lingual representation alignment on MGSM under discrete (Disc-) vs.\ soft (Soft-) thinking. $\Delta$ columns are Soft$-$Disc. Mean over $N{=}50$ parallel samples.}
\footnotesize
\begin{tabular}{lccccccccc}
\toprule
& \multicolumn{3}{c}{\textbf{Qwen3-8B}} & \multicolumn{3}{c}{\textbf{Gemma3-12B}} & \multicolumn{3}{c}{\textbf{DeepSeek-R1-Distill-Qwen-7B}} \\
\cmidrule(lr){2-4} \cmidrule(lr){5-7} \cmidrule(lr){8-10}
\textbf{Metric} & \textbf{Early} & \textbf{Mid} & \textbf{Final} & \textbf{Early} & \textbf{Mid} & \textbf{Final} & \textbf{Early} & \textbf{Mid} & \textbf{Final} \\
\midrule
Disc-CLS & 0.82 & 0.97 & 0.23 & 0.99 & 1.00 & 0.25 & 0.74 & 1.00 & 0.24 \\
Soft-CLS & 0.83 & 0.94 & 0.36 & 0.93 & 1.00 & 0.93 & 0.73 & 0.97 & 0.25 \\
$\Delta$CLS & \dgood{+0.01} & \dbad{-0.03} & \dgood{+0.13} & \dbad{-0.06} & \dgood{+0.00} & \dgood{+0.68} & \dbad{-0.01} & \dbad{-0.03} & \dgood{+0.01} \\
\midrule
Disc-SS & 0.53 & 0.15 & 0.87 & 0.31 & 0.21 & 0.82 & 0.55 & 0.11 & 0.83 \\
Soft-SS & 0.60 & 0.27 & 0.83 & 0.54 & 0.08 & 0.09 & 0.59 & 0.20 & 0.82 \\
$\Delta$SS & \dbad{+0.07} & \dbad{+0.12} & \dgood{-0.04} & \dbad{+0.23} & \dgood{-0.13} & \dgood{-0.73} & \dbad{+0.04} & \dbad{+0.09} & \dgood{-0.01} \\
\bottomrule
\end{tabular}
\label{tab:pilot_results}
\end{table*}

\begin{table*}[!t]
\centering
\begin{minipage}[t]{0.49\linewidth}
\centering
\captionof{table}{Final-layer cross-lingual representation analysis on Qwen3-4B/8B and their fine-tuned variants. $\Delta$ rows are training effects. Results for early/middle layers in Appendix~\ref{appendix:repr-analysis}.}
\label{tab:repr-after-training}
\resizebox{\linewidth}{!}{%
\begin{tabular}{l cc cc}
\toprule
              & \multicolumn{2}{c}{\textbf{Discrete Thinking}}                   & \multicolumn{2}{c}{\textbf{Soft Thinking}}              \\
\cmidrule(lr){2-3}\cmidrule(lr){4-5}
              & \textbf{CLS} & \textbf{SS} & \textbf{CLS} & \textbf{SS} \\
\midrule
Qwen3-4B      & 0.14     & 0.89     & 0.24     & 0.88     \\
~~+ SFT       & 0.30     & 0.86     & 0.40     & 0.85     \\
\quad$\Delta$ & \dgood{+0.16}      & \dgood{-0.03}      & \dgood{+0.16}      & \dgood{-0.03}      \\
~~+ SOLAR     & 0.50     & 0.77     & 0.42     & 0.81     \\
\quad$\Delta$ & \goodc{+0.36}+0.36 & \goodc{-0.12}-0.12 & \dgood{+0.18}      & \dgood{-0.07}      \\
\midrule
Qwen3-8B      & 0.23     & 0.87     & 0.36     & 0.83     \\
~~+ SFT       & 0.50     & 0.86     & 0.56     & 0.86     \\
\quad$\Delta$ & \goodc{+0.27}+0.27 & \dgood{-0.01}      & \goodc{+0.20}+0.20 & \dbad{+0.03}       \\
~~+ SOLAR     & 0.76     & 0.77     & 0.87     & 0.76     \\
\quad$\Delta$ & \goodc{+0.53}+0.53 & \goodc{-0.10}-0.10 & \goodc{+0.51}+0.51 & \dgood{-0.07}      \\
\bottomrule
\end{tabular}}
\end{minipage}%
\hfill
\begin{minipage}[t]{0.49\linewidth}
\centering
\captionof{table}{Distribution-level target-script ratio (\%) on MGSM under top-$k$/top-$p$.}
\label{tab:target-script-ratio-main}
\resizebox{\linewidth}{!}{%
\begin{tabular}{lccccc}
\toprule
\textbf{Model} & \textbf{Zh} & \textbf{Ja} & \textbf{Th} & \textbf{Te} & \textbf{Avg.} \\
\midrule
\multicolumn{6}{l}{(a) Top-$k$ ($k{=}30$)} \\
\quad Qwen3-4B & \sbest{82.26} & 78.87 & 87.70 & \sbest{65.37} & 78.55 \\
\quad~~+ SFT & \best{91.29} & \best{92.26} & \best{91.91} & 65.21 & \sbest{85.17} \\
\quad~~+ SOLAR & 79.34 & \sbest{81.14} & \sbest{91.21} & \best{98.37} & \best{87.51} \\
\midrule
\quad Qwen3-8B & \sbest{92.49} & 19.73 & 27.99 & 11.97 & 38.05 \\
\quad~~+ SFT & \best{93.09} & \best{93.09} & \best{94.41} & \sbest{69.05} & \sbest{87.41} \\
\quad~~+ SOLAR & 82.83 & \sbest{77.79} & \sbest{91.75} & \best{98.99} & \best{87.84} \\
\midrule
\multicolumn{6}{l}{(b) Top-$p$ ($p{=}0.9$)} \\
\quad Qwen3-4B & 94.57 & 95.84 & 97.22 & 94.37 & 95.50 \\
\quad~~+ SFT & \sbest{95.69} & \best{97.35} & \sbest{98.41} & \sbest{98.75} & \sbest{97.55} \\
\quad~~+ SOLAR & \best{95.75} & \sbest{97.16} & \best{98.46} & \best{99.34} & \best{97.68} \\
\midrule
\quad Qwen3-8B & \best{97.99} & 21.75 & 32.59 & 13.15 & 41.37 \\
\quad~~+ SFT & \sbest{96.21} & \best{97.14} & \best{98.47} & \sbest{98.78} & \best{97.65} \\
\quad~~+ SOLAR & 95.68 & \sbest{96.05} & \sbest{97.38} & \best{99.47} & \sbest{97.14} \\
\bottomrule
\end{tabular}}
\end{minipage}
\end{table*}

\paragraph{Cross-lingual similarity follows an inverted-U (low-high-low) pattern across early, middle, and final layers.}

Table~\ref{tab:pilot_results} compares representations under discrete and inference-time soft thinking on MGSM for three base models, isolating the effect of soft tokens at inference time before introducing SOLAR's training-time alignment.
With discrete tokens, CLS is high in early and middle layers but drops sharply at the final layer. Figure~\ref{fig:layer-grid-gemma} visualizes this trend across the three probed layer depths for Gemma3-12B. SS shows a U-shaped pattern, reaching its lowest values at the middle layer before rising sharply at the final layer. This matches prior findings of more language-agnostic semantics mid-network and language-specific structure in final layers~\cite{lim2025languagespecific,wang-etal-2025-lost-multilinguality,xu-etal-2025-linguistic,tezuka-inoue-2025-transfer,pokharel2026crosslingualactivationsteeringmultilingual}.

\paragraph{Soft tokens improve final-layer alignment but do not dissolve language clusters.} Compared to discrete thinking, soft tokens raise final-layer CLS across all models, with the largest gain in Gemma3-12B ($\Delta$CLS{=}+0.68, Table~\ref{tab:pilot_results}). Early and middle layers see little benefit and can even degrade (\textit{e.g.,} Qwen3-8B has $\Delta$CLS{=}$-$0.03 at the middle layer). These layers already reach high CLS under discrete thinking, leaving little room to improve, and the distributional shift from soft tokens may slightly disrupt their alignment. SS improvements are limited and inconsistent: only Gemma3-12B shows a large decrease ($\Delta$SS{=}$-$0.73), while other models show negligible change or slight increases. Language clusters thus largely remain intact for most models (Figure~\ref{fig:gemma_12b_tsne}).

\paragraph{SOLAR produces the strongest cross-lingual representations.} Having shown that soft tokens carry useful cross-lingual signal but that inference-time use alone does not consistently improve performance (\S\ref{sec:main_results}), we examine whether fine-tuning reshapes representations more effectively (Table~\ref{tab:repr-after-training}; same 50 MGSM samples). The base Qwen3-4B shows the inverted-U pattern, with CLS dropping to 0.14 at the final layer. SFT raises final-layer CLS to 0.30 (+0.16), while SOLAR reaches 0.50 (+0.36, more than 2$\times$ the SFT gain) and substantially reduces final-layer SS from 0.89 to 0.77 ($-$0.12), indicating weaker language clustering. Applying soft-thinking inference to fine-tuned models improves SFT's CLS (0.30$\rightarrow$0.40) but slightly reduces SOLAR's (0.50$\rightarrow$0.42). This suggests that SOLAR internalizes soft tokens' alignment benefit, subsuming what inference-time soft thinking can recover post-hoc. On Qwen3-8B, SOLAR's discrete-thinking gain is larger ($\Delta$CLS\,=\,+0.53) and, unlike on 4B, remains nearly intact under soft thinking (+0.51).

\subsection{Behavioral analysis}\label{sec:behavioral}
We examine whether fine-tuning changes the model's reasoning behavior on MGSM (Zh, Ja, Th, Te) at the \textit{token} level (the generated tokens of each reasoning trace) and the \textit{distribution} level (the top-$k$/top-$p$ candidates in the output distribution during teacher-forced decoding). At both levels, every token is classified into one of three categories using Unicode code-point ranges: \textbf{target-script} (the prompt language's native writing system, \textit{e.g.,} CJK ideographs for Chinese), \textbf{Latin} (Unicode names containing \textsc{latin}, used as a proxy for English), and \textbf{neutral} (digits, punctuation, math symbols, and whitespace). We define the \emph{target-script ratio} as $\#\,\text{target} \,/\, (\#\,\text{target} + \#\,\text{Latin})$ to exclude neutral tokens so that script preference is not diluted by language-agnostic symbols.

\paragraph{Fine-tuning preserves native language choice at the token level.}
Qwen3-4B reasons almost entirely in the target script (98.81\%). SFT and SOLAR maintain comparable ratios (Figure~\ref{fig:target_script_ratio}). The ratio is nearly identical for correct and incorrect answers (average $\Delta < 0.5\%$; Table~\ref{tab:reasoning-language-mgsm}), indicating that fine-tuning improves task performance without shifting language choice toward English.

\paragraph{SFT and SOLAR achieve high target-script ratios through distinct mechanisms.}
\begin{wrapfigure}[13]{r}{0.5\linewidth}
  \centering
  \vspace{-1.5\baselineskip}
  \includegraphics[width=\linewidth]{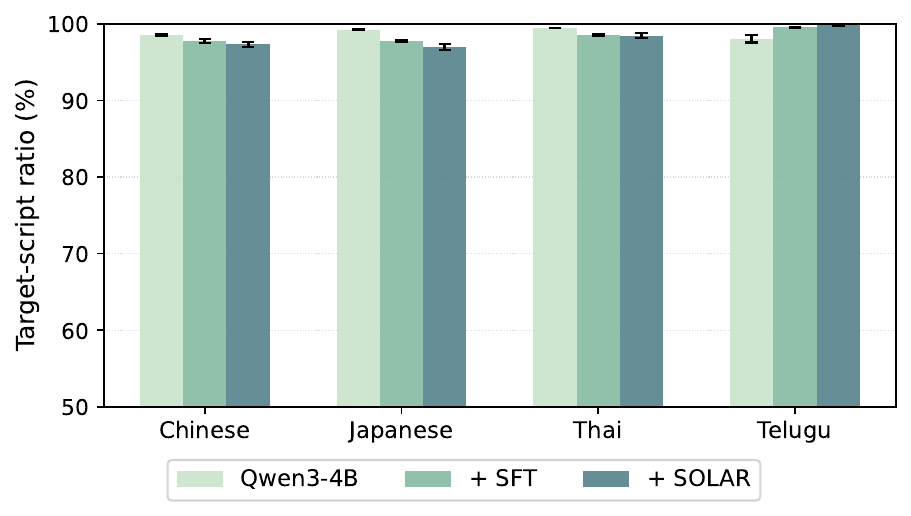}
  \addtocounter{figure}{1}\caption{Token-level target-script ratio (\%) on MGSM for Qwen3-4B and its fine-tuned variants.}
  \label{fig:target_script_ratio}
\end{wrapfigure}\addtocounter{figure}{-2}
The 8B base model shows a stronger English preference than 4B (\textit{e.g.,} Telugu 11.97\% vs.\ above 65\%; Figure~\ref{fig:combined-composition-main}, Table~\ref{tab:target-script-ratio-main}). SFT and SOLAR close this gap (8B average 38.05\%$\rightarrow$above 87\%) through different mechanisms: SFT raises target-script token density in the top-30 predictions, while SOLAR suppresses Latin competition. The contrast is clearest for Telugu, where SOLAR nearly eliminates Latin competition (98.37\% for 4B) but SFT leaves it largely unchanged (65.21\% vs.\ 65.37\%). Under top-$p$ sampling, all fine-tuned models reach 95--99\% target-script ratios, so remaining cross-lingual competition primarily concerns low-probability alternatives rather than the dominant script preference.

\begin{figure*}[!t]
  \centering
  \includegraphics[width=1\textwidth]{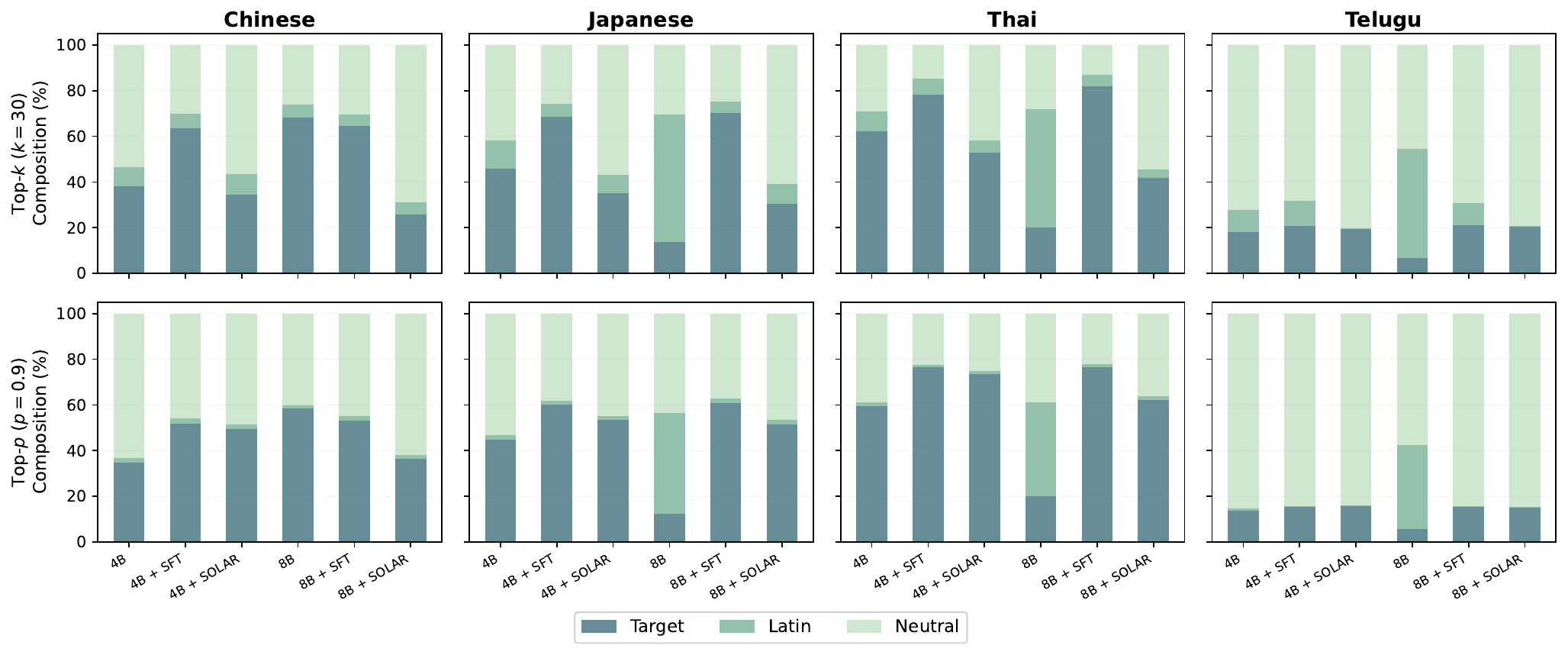}
  \caption{Distribution-level token composition by language script on MGSM for Qwen3-4B, Qwen3-8B, and their fine-tuned variants under top-$k$ ($k{=}30$, top row) and top-$p$ ($p{=}0.9$, bottom row). Each bar shows target-script (dark teal), Latin (medium green), and neutral (light green) tokens.}
  \label{fig:combined-composition-main}
\end{figure*}\addtocounter{figure}{1}

\subsection{What drives the alignment gain?}\label{sec:ablation-design}
To isolate SOLAR's alignment gains, we vary the alignment target, pooling granularity, top-$k$ filter, representation space, and pivot language one at a time, keeping other settings fixed.

\begin{table*}[t]
\centering
\caption{Alignment-target ablation for SOLAR with Qwen3-4B.}
\label{tab:ablation-target}
\resizebox{\textwidth}{!}{%
\begin{tabular}{lcccccccccc}
\toprule
& \multicolumn{2}{c}{\textbf{MGSM}} & \multicolumn{2}{c}{\textbf{AIME~2024}} & \multicolumn{2}{c}{\textbf{AIME~2025}} & \multicolumn{2}{c}{\textbf{GPQA}} & \multicolumn{2}{c}{\textbf{Avg.}}\\
\cmidrule(lr){2-3} \cmidrule(lr){4-5} \cmidrule(lr){6-7} \cmidrule(lr){8-9} \cmidrule(lr){10-11}
\textbf{Alignment target} & \textbf{Accuracy} & \textbf{CLC} & \textbf{Accuracy} & \textbf{CLC} & \textbf{Accuracy} & \textbf{CLC} & \textbf{Accuracy} & \textbf{CLC} & \textbf{Accuracy} & \textbf{CLC}\\
\midrule
Qwen3-4B $+$ SFT      & 76.25 & 72.29 & 34.76 & 25.17 & 27.90 & 20.44 & 43.75 & 33.15 & 45.67 & 37.76\\
~~+ Discrete          & 76.37 & 73.50 & 35.63 & 26.06 & \sbest{28.12} & 21.41 & 43.86 & 33.47 & 46.00 & 38.61\\
~~+ Final-hidden      & 76.97 & 73.79 & 35.47 & 25.91 & 27.87 & 20.05 & \sbest{44.05} & \sbest{34.51} & 46.09 & 38.57\\
~~+ Untruncated       & \sbest{77.19} & \sbest{74.72} & \sbest{37.03} & \sbest{27.63} & 27.93 & \sbest{22.93} & 44.04 & 34.46 & \sbest{46.55} & \sbest{39.94}\\
~~+ SOLAR ($k{=}30$)  & \best{78.27} & \best{75.61} & \best{38.52} & \best{29.67} & \best{30.95} & \best{24.51} & \best{45.24} & \best{36.26} & \best{48.25} & \best{41.51}\\
\bottomrule
\end{tabular}
}
\medskip
\centering
\caption{Pooling-granularity ablation for SOLAR with Qwen3-4B.}
\label{tab:ablation-pooling}
\resizebox{\textwidth}{!}{%
\begin{tabular}{lcccccccccc}
\toprule
& \multicolumn{2}{c}{\textbf{MGSM}} & \multicolumn{2}{c}{\textbf{AIME 2024}} & \multicolumn{2}{c}{\textbf{AIME 2025}} & \multicolumn{2}{c}{\textbf{GPQA}} & \multicolumn{2}{c}{\textbf{Avg.}}\\
\cmidrule(lr){2-3} \cmidrule(lr){4-5} \cmidrule(lr){6-7} \cmidrule(lr){8-9} \cmidrule(lr){10-11}
\textbf{Pooling granularity} & \textbf{Accuracy} & \textbf{CLC} & \textbf{Accuracy} & \textbf{CLC} & \textbf{Accuracy} & \textbf{CLC} & \textbf{Accuracy} & \textbf{CLC} & \textbf{Accuracy} & \textbf{CLC}\\
\midrule
Qwen3-4B $+$ SFT           & 76.25 & 72.29 & 34.76 & 25.17 & 27.90 & 20.44 & 43.75 & 33.15 & 45.67 & 37.76\\
~~+ Answer-only            & 76.00 & 71.91 & 34.93 & 25.92 & 27.92 & 20.94 & 44.95 & 34.96 & 45.95 & 38.43\\
~~+ Step-level             & \sbest{77.04} & \sbest{73.34} & \sbest{37.03} & \sbest{28.95} & \sbest{29.01} & \sbest{23.88} & \best{46.15} & \best{37.81} & \sbest{47.31} & \sbest{41.00}\\
~~+ Full (SOLAR)           & \best{78.27} & \best{75.61} & \best{38.52} & \best{29.67} & \best{30.95} & \best{24.51} & \sbest{45.24} & \sbest{36.26} & \best{48.25} & \best{41.51}\\
\bottomrule
\end{tabular}
}
\medskip
\centering
\caption{Top-$k$ ablation for SOLAR with Qwen3-4B.}
\label{tab:ablation-topk}
\resizebox{\textwidth}{!}{%
\begin{tabular}{lcccccccccc}
\toprule
& \multicolumn{2}{c}{\textbf{MGSM}} & \multicolumn{2}{c}{\textbf{AIME 2024}} & \multicolumn{2}{c}{\textbf{AIME 2025}} & \multicolumn{2}{c}{\textbf{GPQA}} & \multicolumn{2}{c}{\textbf{Avg.}}\\
\cmidrule(lr){2-3} \cmidrule(lr){4-5} \cmidrule(lr){6-7} \cmidrule(lr){8-9} \cmidrule(lr){10-11}
\textbf{Top-$k$} & \textbf{Accuracy} & \textbf{CLC} & \textbf{Accuracy} & \textbf{CLC} & \textbf{Accuracy} & \textbf{CLC} & \textbf{Accuracy} & \textbf{CLC} & \textbf{Accuracy} & \textbf{CLC}\\
\midrule
SFT                  & 76.25 & 72.29 & 34.76 & 25.17 & 27.90 & 20.44 & 43.75 & 33.15 & 45.67 & 37.76\\
$k{=}1$ (Discrete)   & 76.37 & 73.50 & 35.63 & 26.06 & 28.12 & 21.41 & 43.86 & 33.47 & 46.00 & 38.61\\
$k{=}5$              & 77.00 & 73.92 & 36.46 & 26.97 & 28.64 & 22.20 & 44.89 & 35.96 & 46.75 & 39.76\\
$k{=}10$             & \best{78.36} & \best{76.02} & \sbest{38.40} & \sbest{29.12} & 30.93 & 24.42 & 44.92 & 35.98 & 48.15 & \sbest{41.39}\\
$k{=}30$ (SOLAR)     & \sbest{78.27} & \sbest{75.61} & \best{38.52} & \best{29.67} & \sbest{30.95} & \sbest{24.51} & \sbest{45.24} & \sbest{36.26} & \best{48.25} & \best{41.51}\\
$k{=}50$             & 78.08 & 75.11 & 38.34 & 29.00 & \best{31.10} & \best{24.89} & \best{45.29} & \best{36.55} & \sbest{48.20} & \sbest{41.39}\\
\bottomrule
\end{tabular}
}
\end{table*}

\paragraph{Alignment target.}
We compare four auxiliary-loss targets: (1)~\textbf{Discrete}, the straight-through embedding of the top-1 token ($\mathbf{e}_t = \mathbf{E}[y_t]$); (2)~\textbf{Final-hidden}, the mean-pooled final-layer hidden state; (3)~\textbf{Untruncated}, the soft-token embedding over the full vocabulary; and (4)~\textbf{SOLAR}, the default top-$k{=}30$ soft-token embedding.
SOLAR performs best on Qwen3-4B (48.25\% average accuracy, 41.51\% CLC; Table~\ref{tab:ablation-target}). Discrete and Final-hidden stay near SFT (46.00\% and 46.09\% vs.\ 45.67\%), while Untruncated, the strongest alternative (46.55\%), still trails SOLAR. A single hard token recovers almost none of SOLAR's gain, indicating that soft mixtures and top-$k$ truncation contribute beyond compute savings. These results suggest that top-$k$ filtering suppresses cross-lingual noise in Untruncated's long tail and that embedding-space soft-token mixtures provide a more language-invariant alignment target than mean-pooled final-layer hidden states.

\paragraph{Pooling granularity.}
With the soft-token target fixed, we compare: (1)~\textbf{Answer-only}, pooling over the final \texttt{\textbackslash boxed\{\}} span; (2)~\textbf{Step-level}, pooling within steps split on \texttt{\textbackslash n\textbackslash n} and matched by relative position; and (3)~\textbf{Full}, pooling over the entire response.
Answer-only matches SFT (MGSM 76.00\% vs.\ 76.25\%, Table~\ref{tab:ablation-pooling}), locating the alignment signal in the reasoning trace. On MGSM, Step-level improves over Answer-only but trails Full (77.04\% vs.\ 78.27\%). On GPQA, however, Step-level exceeds Full in both accuracy (46.15\% vs.\ 45.24\%) and CLC (37.81\% vs.\ 36.26\%). We hypothesize that finer granularity helps only when steps correspond across languages: relative-position matching assumes parallel step structure, but equivalent reasoning may span different numbers of \texttt{\textbackslash n\textbackslash n}-delimited steps. Full-response pooling avoids this mismatch; finer-grained pooling likely requires content-aware segmentation to align steps across languages.

\begin{wraptable}[12]{r}{0.4\linewidth}
\centering
\vspace{\dimexpr-\baselineskip-2.4pt\relax}
\small
\caption{Representation-space and pivot-language ablation for SOLAR with Qwen3-4B on MGSM.}
\label{tab:ablation-space-pivot}
\begin{tabular}{lcc}
\toprule
\textbf{Configuration} & \textbf{Accuracy} & \textbf{CLC}\\
\midrule
\multicolumn{3}{l}{\textit{(a) Representation space}} \\
\quad $\mathbb{R}^d$ (emb.) & \best{78.27} & \best{75.61}\\
\quad $\mathbb{R}^{|V|}$ (prob.) & \sbest{60.27} & \sbest{44.76} \\
\midrule
\multicolumn{3}{l}{\textit{(b) Pivot language $\ell^*$}}\\
\quad English (En) & \best{78.27} & \best{75.61}\\
\quad Chinese (Zh) & \sbest{74.50} & \sbest{70.97}\\
\quad French (Fr) & 72.87 & 66.67\\
\bottomrule
\end{tabular}
\end{wraptable}
\paragraph{Top-$k$ sensitivity.} We test $k \in \{1, 5, 10, 30, 50\}$; $k{=}1$ recovers the Discrete target (Table~\ref{tab:ablation-topk}). Accuracy rises steeply to $k{=}10$, then plateaus near 48\% for $k \in \{10, 30, 50\}$, with no value dominating across benchmarks. Larger $k$ increases alignment-loss compute without improving accuracy. The overhead is small: cross-entropy already requires logits over $|V|$, so alignment adds only top-$k$ selection ($O(|V| \log k)$ vs.\ $O(|V| \cdot d)$ for logits) and a weighted sum over $k{=}30$ embedding rows per position. We use $k{=}30$; training overhead is in Appendix~\ref{appendix:training-cost}.

\paragraph{Alignment representation space.}

We compare alignment in $\mathbb{R}^d$ (soft-token embeddings) and $\mathbb{R}^{|V|}$ (probability distributions), fixing $\lambda{=}0.50$ and $\tau{=}1.0$. Probability-space alignment reduces accuracy to 60.3\% and CLC to 44.8\% (Table~\ref{tab:ablation-space-pivot}), suggesting that largely disjoint cross-lingual token support makes cosine distance less informative. The shared embedding space preserves cross-lingual correspondences.

\paragraph{Pivot language $\ell^*$.} With $\lambda{=}0.50$ and $\tau{=}1.0$ fixed, English outperforms Chinese and French pivots (Table~\ref{tab:ablation-space-pivot}). The gap concentrates in low-resource languages: MGSM Swahili accuracy drops from 53.48\% (En) to 40.56\% (Zh) and 36.80\% (Fr) (Table~\ref{tab:secondary-ablations}), suggesting that transfer depends on the pivot's pre-trained representation strength, with English the strongest anchor.

\subsection{Hyperparameter sensitivity}\label{sec:ablation}

\paragraph{Alignment weight $\lambda$.}
\begin{wraptable}[13]{r}{0.4\linewidth}
\centering
\vspace{\dimexpr-\baselineskip-2.4pt\relax}
\small
\caption{Hyperparameter ablation ($\lambda$, $\tau$) for SOLAR with Qwen3-4B on MGSM.}
\label{tab:ablation-main}
\begin{tabular}{lcc}
\toprule
\textbf{Configuration} & \textbf{Accuracy} & \textbf{CLC}\\
\midrule
\multicolumn{3}{l}{\textit{(a) Alignment weight $\lambda$}} \\
\quad 0.25 & \best{78.61} & \best{75.73} \\
\quad 0.50 & \sbest{78.27} & \sbest{75.61}\\
\quad 0.75 & 75.83 & 73.01 \\
\quad 1.00 & 69.25 & 63.67 \\
\midrule
\multicolumn{3}{l}{\textit{(b) Soft-token temperature $\tau$}} \\
\quad 0.6 & 60.48 & 55.43 \\
\quad 1.0 & \best{78.27} & \best{75.61}\\
\quad 2.0 & \sbest{76.41} & \sbest{73.13} \\
\bottomrule
\end{tabular}
\end{wraptable}
At $\tau{=}1.0$, we sweep $\lambda \in \{0.25, 0.50, 0.75, 1.00\}$.
Both $\lambda{=}0.25$ and $\lambda{=}0.50$ perform strongly on MGSM (Table~\ref{tab:ablation-main}). Larger weights hurt performance; $\lambda{=}1.00$ falls below the base model, suggesting that excessive alignment overwhelms cross-entropy.
We use $\lambda{=}0.50$ for all main experiments. Both settings ($\lambda \in \{0.25, 0.50\}$) outperform the strongest baseline on each benchmark, with their accuracies differing by at most 1.09 percentage points (Table~\ref{tab:lambda-ablation-all}).

\paragraph{Soft-token temperature $\tau$.} At $\lambda{=}0.50$, we sweep $\tau \in \{0.6, 1.0, 2.0\}$; lower $\tau$ sharpens the soft-token distribution, while higher $\tau$ smooths it. $\tau{=}1.0$ performs best (78.3\% accuracy, 75.6\% CLC; Table~\ref{tab:ablation-main}). At $\tau{=}0.6$, accuracy and CLC drop sharply to 60.5\% and 55.4\%, suggesting loss of cross-lingual signal; $\tau{=}2.0$ yields a moderate drop to 76.4\% and 73.1\%. We apply the default configuration to Qwen3-8B and Gemma3-4B/12B without model-specific retuning of either hyperparameter.

%% file: sections/6-related-work.tex
\section{Related work}
\paragraph{Multilingual reasoning.} 
Multilingual LLMs show inconsistent cross-lingual reasoning~\cite{shi2023language,qi-etal-2023-cross}. \citet{wang-etal-2025-lost-multilinguality} trace errors to the final-layer transition from language-agnostic representations to language-specific outputs. \citet{liu2025focusinglanguagerevealingexploiting} identify language-specific attention heads mediating cross-lingual attention transfer. \citet{lim2025languagespecific} find that larger models rely more on language-specific representations.
Methods operate during training or inference~\cite{ghosh-etal-2025-survey}.
Training-based methods study multilingual pre-training schedules~\cite{wang2026crosslingual}, parallel reasoning data during post-training~\cite{yang2025parallelscalinglawunveiling}, language-mixed CoT data~\cite{son2026pushingmultilingualreasoningmodels}, adapter merging~\cite{zhao-etal-2025-adamergex}, and representation alignment for multilingual capability transfer~\cite{bu-etal-2025-alignx} or safety~\cite{bu2026align}.
Recent multilingual reinforcement learning (RL) methods use an English pivot with translation-quality, language-consistency, or self-feedback rewards~\cite{zhao2026alignpivotdualalignment,huang2026tapotranslationaugmentedpolicy,zhang2026thinknativelyunlockingmultilingual}.
xCoT~\cite{chai2025xcot} uses English responses and aligns per-position output distributions across languages.
Inference approaches use in-context learning~\cite{tanwar-etal-2023-multilingual}, activation steering~\cite{pokharel2026crosslingualactivationsteeringmultilingual}, test-time scaling~\cite{yong2025crosslingualreasoningtesttimescaling}, and inference-time linear transformations between source-language and target-language hidden representations~\cite{wang-etal-2025-bridging}.

\paragraph{Continuous-space reasoning.}
CoT prompting~\cite{NEURIPS2022_9d560961} enhances reasoning with discrete tokens; latent reasoning explores alternatives to token sequences~\cite{chen2025reasoninglanguagecomprehensivesurvey,tan2025think}.
LLMs can learn continuous-space reasoning~\cite{xu-etal-2025-softcot,shen2025codicompressingchainofthoughtcontinuous}; COCONUT~\cite{hao2025training} feeds its last hidden state back as an input embedding for parallel reasoning.
\citet{butt2025softtokenshardtruths} train continuous CoTs with reinforcement learning and noise in the input embedding mixture for exploration; discrete decoding performs best after training.
Training-free Soft Thinking~\cite{zhang2025soft} replaces discrete embeddings with probability-weighted soft tokens.
Theoretical work shows efficient graph reachability with continuous CoT under its stated assumptions~\cite{zhu2025reasoningsuperpositiontheoreticalperspective}.
These results demonstrate practical latent-reasoning compression methods~\cite{tan2025think}.

%% file: sections/7-conclusion.tex
\section{Conclusion}
We introduce SOLAR, an auxiliary training objective that aligns soft-token representations across parallel multilingual reasoning traces. SOLAR improves task accuracy and cross-lingual consistency over standard SFT across four multilingual benchmarks, with the largest gains on low-resource languages. On MGSM, SOLAR improves final-layer CLS and reduces SS on Qwen3-4B/8B; under discrete thinking, its CLS gain on Qwen3-4B is more than twice that of SFT.

%% file: appendix/0-limitations-impacts.tex
\section{Further discussion, limitations, and future work}\label{sec:limitations}

\paragraph{Extensions to the alignment objective.}
SOLAR is evaluated as an auxiliary loss during SFT. Recent work reports stronger cross-lingual performance and unseen-language generalization with reinforcement learning than with SFT~\citep{huang2025englishcentrictrainingreinforcementlearning}, motivating integration with DPO~\cite{NEURIPS2023_a85b405e} or GRPO~\cite{shao2024deepseekmathpushinglimitsmathematical}.
SOLAR could complement the distinct signals used by PASMR~\cite{zhao2026alignpivotdualalignment} (pivot-derived self-feedback), M-Thinker~\cite{zhang2026thinknativelyunlockingmultilingual} (cross-lingual thinking alignment), and TAPO~\cite{huang2026tapotranslationaugmentedpolicy} (English-translation quality and final-answer correctness).
Symmetric alignment would remove the pivot stop-gradient in Eq.~\ref{eq:pair-loss}, allowing alignment gradients into the English branch. Its effect on pivot quality needs testing: transfer depends on the pivot (\S\ref{sec:ablation-design}), while current asymmetric alignment preserves English MGSM accuracy at 93.32\%, versus 93.36\% for the base model (Table~\ref{tab:mgsm-per-lang-main}).
Multi-pivot alignment could use a centroid of high-resource language summaries as the target in Eq.~\ref{eq:pair-loss}.

\paragraph{Evaluation scope.}
Our evaluation covers mathematical reasoning (MGSM, AIME~2024/2025) and scientific question answering (GPQA~Diamond), with short, verifiable answers.
Whether response-level mean-pooled soft-token summaries support open-ended generation, such as multilingual summarization~\cite{gao-etal-2025-mixed}, remains untested; step- or chunk-level alignment may better suit non-monotonic or retrieval-interleaved traces.
Our benchmarks also assume language-invariant correct answers (\S\ref{sec:experiments}). They cannot assess culturally dependent answers, for which the assumptions underlying the English pivot and CLC may fail. Culturally grounded multilingual evaluation is therefore needed.

\paragraph{Language and model coverage.}
M-s1k~\citep{barua2025longchainofthoughtreasoninglanguages} provides parallel long CoT traces in 10 languages; we evaluate the 7 overlapping with our benchmarks (Appendix~\ref{appendix:datasets}).
Broader coverage, especially of low-resource languages, requires costly translation and mathematical-fidelity verification. Quality-filtered translations or unpaired monolingual traces anchored through shared retrieval could relax the parallel-data requirement, but remain untested.
Fine-tuning is validated on Qwen3-4B/8B and Gemma3-4B/12B; the 10-model representation analysis additionally covers DeepSeek-R1-Distill (\S\ref{sec:repr-analysis}).
Larger models (30B+), mixture-of-experts architectures, and multimodal LLMs may differ in vocabulary geometry and pre-training composition; their alignment signal needs validation.

\section{Broader impacts}\label{sec:broader-impacts}
As multilingual LLMs serve users worldwide, ensuring consistent reasoning across languages is both a technical and an equity concern. Recent multilingual evaluations report substantial performance gaps between high- and low-resource languages~\cite{xuan-etal-2025-mmlu}. Regional medical benchmarks also reveal uneven performance across specialties and geographic contexts~\cite{nimo-etal-2025-afrimed}. Our work targets the cross-lingual performance gap by aligning soft-token representations across languages during SFT, with the largest gains on low-resource languages (\textit{e.g.,} Swahili MGSM accuracy rises from 29.04\% to 53.48\% for Qwen3-4B, and Telugu AIME~2024 from 5.67\% to 29.67\%), narrowing performance disparities while preserving native-language generation (Appendix~\ref{appendix:behavioral_analysis}).

Beyond accuracy, our analysis helps make multilingual model behavior more interpretable and controllable. Soft-token embeddings provide a continuous, vocabulary-shared signal in which cross-lingual reasoning can be inspected and shaped, complementing hidden-state probing and offering a lightweight, architecture-agnostic mechanism that can be combined with other multilingual training pipelines. Because SOLAR adds no architectural modification and reuses the model's own forward-pass logits, it is also accessible when compute or model-modification budgets are limited.

At the same time, anchoring all non-English summaries to an English pivot raises questions about cultural and linguistic representation. An English-anchored shared space may carry English-centric reasoning conventions, framing assumptions, or normative answers that surface only on culturally grounded tasks not covered by our benchmarks~\cite{gamboa-etal-2025-social}. Stronger multilingual reasoning also extends the reach of downstream misuse currently bottlenecked by English-only proficiency, and our work does not address multilingual safety alignment. We see multi-pivot or pivot-free alignment, evaluation on culturally grounded reasoning, and multilingual safety post-training as important complements to our work. Overall, we hope our work contributes to a more inclusive and principled approach to multilingual reasoning and helps narrow the cross-lingual capability gap in a way that is mindful of the linguistic communities these systems aim to serve.

%% file: appendix/1-related-work.tex
\section{Related work}
\subsection{Multilingual reasoning}
LLMs trained on multilingual corpora exhibit cross-lingual reasoning abilities~\cite{gemmateam2025gemma3technicalreport,yang2025qwen3technicalreport,zhao2025comprehensiveevaluationmultilingualchainofthought}, yet show inconsistent performance across different languages~\cite{qi-etal-2023-cross,wang-etal-2025-lost-multilinguality}. Specifically, given the same task, LLMs generally perform better when prompted in English, whereas their reasoning performance deteriorates when the task is expressed in low-resource languages. 
Recent work examines these disparities through internal model mechanisms. \citet{wang-etal-2025-lost-multilinguality} reveal that models encode factual knowledge in a language-agnostic concept space through intermediate layers, but errors arise during the final-layer transition to language-specific representations. \citet{liu2025focusinglanguagerevealingexploiting} identify language-specific attention heads that enable cross-lingual attention transfer. Recently, \citet{lim2025languagespecific} show that larger models increasingly rely on language-specific representations, hindering cross-lingual transfer. These mechanistic insights provide a foundation for understanding and improving cross-lingual reasoning.

\subsection{Cross-lingual alignment}
Multilingual reasoning methods operate during training or inference~\cite{ghosh-etal-2025-survey}.
Training-based approaches study multilingual pre-training data mixtures~\cite{foroutan2025revisitingmultilingualdatamixtures}, the role of parallel data~\cite{qorib-etal-2025-just,lin2025from}, and cross-lingual data scaling~\cite{wang2026crosslingual}.
Multilingual post-training includes instruction tuning~\cite{kew-etal-2024-turning,lai-nissim-2024-mcot,barua2025longchainofthoughtreasoninglanguages}, parallel reasoning data~\cite{yang2025parallelscalinglawunveiling}, and language-mixed CoT data~\cite{chen2025datatokensmultilingualunification,son2026pushingmultilingualreasoningmodels}.
Related approaches use translation-assisted information extraction~\cite{chen-etal-2025-translation}, adapter merging~\cite{zhao-etal-2025-adamergex}, representation alignment for transfer~\cite{liu-niehues-2025-middle,bu-etal-2025-alignx}, debiasing~\cite{peng-etal-2025-debiasing}, or safety~\cite{bu2026align}.
Other work studies multilingual post-training~\cite{she-etal-2024-mapo,wang-etal-2025-calm,lee-etal-2025-cross,faisal2025aligningmultilingualreasoningverifiable,li2025steering,liu2026optimizing,wu2026mgrpo,hwang2025learngloballyspeaklocally}.
\citet{huang2025englishcentrictrainingreinforcementlearning} report that RL outperforms SFT in cross-lingual reasoning and generalization, even with non-English training data.
PASMR~\cite{zhao2026alignpivotdualalignment} maps multilingual inputs to an English pivot via SFT, then uses self-feedback RL to reward cross-lingual reasoning consistency.
TAPO~\cite{huang2026tapotranslationaugmentedpolicy} combines translation-quality and reasoning rewards, using step-level relative advantages to separate credit assignment for translation and reasoning tokens.
M-Thinker~\cite{zhang2026thinknativelyunlockingmultilingual} combines language-consistency and thinking-alignment rewards, using an LLM judge against an English reasoning trace.
xCoT~\cite{chai2025xcot} pivots on English at the data level: translated questions share English responses, enabling per-position output-distribution alignment between parallel samples. SOLAR instead aligns mean-pooled soft-token embeddings in $\mathbb{R}^d$, retains each language's native trace supervision, and requires neither positional correspondence nor shared token support. These methods could complement SOLAR's alignment signal (Appendix~\ref{sec:limitations}).
Inference-based methods include in-context learning~\cite{tanwar-etal-2023-multilingual,yoo2025codeswitchingincontextlearningcrosslingual}, cross-lingual activation steering~\cite{10.1145/3626772.3657819,pokharel2026crosslingualactivationsteeringmultilingual}, test-time scaling~\cite{son-etal-2025-linguistic,yong2025crosslingualreasoningtesttimescaling}, and learned linear transformations between source- and target-language hidden states applied at inference~\cite{wang-etal-2025-bridging}. Studies show that prompting models to reason in English generally preserves or improves non-English task performance~\cite{barua2025longchainofthoughtreasoninglanguages,yong2025crosslingualreasoningtesttimescaling}.
SOLAR instead updates model parameters during SFT with an auxiliary soft-token alignment loss, requiring no per-language inference transformation or overhead beyond standard discrete decoding.

\subsection{Continuous-space reasoning}
CoT prompting~\cite{NEURIPS2022_9d560961} enhances reasoning through discrete tokens; latent reasoning draws on abstract concepts in human cognition~\cite{chen2025reasoninglanguagecomprehensivesurvey}.
Several methods train LLMs to reason in continuous space~\cite{xu-etal-2025-softcot,pham2025multimodal,shen2025codicompressingchainofthoughtcontinuous}. COCONUT~\cite{hao2025training} feeds its last hidden state back as an input embedding to encode multiple reasoning steps simultaneously.
\citet{bashir2026continuouscot} evaluate CODI-based latent reasoning on multilingual GSM8k and CommonsenseQA, with benefits particularly for Urdu and zero-shot transfer. They train continuous hidden-state reasoning via self-distillation and use latent decoding at inference, without explicit pairwise cross-lingual alignment. SOLAR aligns probability-weighted vocabulary embeddings during SFT and decodes discretely. Their results motivate multilingual latent reasoning but do not directly test SOLAR's soft-token alignment hypothesis.
Training-free Soft Thinking~\cite{zhang2025soft} replaces discrete embeddings with probability-weighted soft tokens for richer inference-time representations. \citet{wu2025llmssinglethreadedreasonersdemystifying} add noise via Dirichlet resampling and the Gumbel-Softmax trick.
\citet{butt2025softtokenshardtruths} train continuous CoTs with reinforcement learning and Gaussian noise in the input embedding mixture for exploration, without distilling ground-truth discrete traces. On math benchmarks, continuous-CoT training matches discrete training at pass@1 and surpasses it at pass@32; discrete decoding after continuous-CoT training performs best.
Under their respective theoretical assumptions, continuous reasoning supports parallel trace exploration and subset-sum constructions~\cite{gozeten2025continuouschainthoughtenables}, as well as efficient graph-reachability constructions~\cite{zhu2025reasoningsuperpositiontheoreticalperspective}.
These studies identify task-dependent computational advantages of continuous reasoning~\cite{xu2026formalcomparisonchainthought} and demonstrate practical latent-reasoning compression methods~\cite{tan2025think}.
SOLAR uses probability-weighted soft tokens~\cite{butt2025softtokenshardtruths,zhang2025soft} for cross-lingual alignment during supervised fine-tuning. We test whether mixtures over the shared vocabulary embedding matrix remain comparable across languages with disjoint surface tokens.

%% file: appendix/2-implementation-details.tex
\section{Implementation details}\label{appendix:preliminary_implementation_details}

\subsection{Models}\label{appendix:models}
For main experiments (\S\ref{sec:experiments}), we fine-tune Qwen3-4B/8B~\cite{yang2025qwen3technicalreport} and Gemma3-4B/12B~\cite{gemmateam2025gemma3technicalreport} via full-parameter SFT on M-s1k. We center the headline analysis on the Qwen3 family for its strong multilingual reasoning and wide adoption; full Gemma3 results (MGSM and GPQA) appear in \S\ref{appendix:experimental_results}. We use the 4B and 8B scales for Qwen3 because they balance reasoning capability against the compute cost of full-parameter fine-tuning on A40 GPUs (48\,GB).
For representation experiments (\S\ref{sec:repr-analysis}), we select 10 models from three families: Qwen3, Gemma3, and DeepSeek-R1-Distill~\cite{deepseekai2025deepseekr1incentivizingreasoningcapability}. These span multiple scales (1B--27B) and both base and instruction-tuned variants (\textit{e.g.,}\ Qwen3-4B-Base \textit{vs.}\ Qwen3-4B). Specifically, Qwen3 and Gemma3 are advanced multilingual models supporting 119 and 140+ languages, respectively.

\subsection{Training setup}\label{appendix:training}
All main-experiment models undergo full-parameter fine-tuning using the \texttt{LLaMA-Factory} framework~\cite{zheng-etal-2024-llamafactory}. Qwen3-4B is trained in non-thinking mode, while Qwen3-8B is trained in thinking mode. We train for 5 epochs with a batch size of 32. All runs use bfloat16 precision with a learning rate of $1 \times 10^{-5}$ that warms up linearly over the first 10\% of training and then decays to 0 following a cosine schedule. We use the AdamW optimizer~\cite{loshchilov2019decoupledweightdecayregularization} with $\beta_1 = 0.9$ and $\beta_2 = 0.999$. The maximum sequence length is 16{,}384 tokens. We use DeepSpeed ZeRO Stage~2~\cite{9355301} to reduce memory usage. All experiments are conducted on NVIDIA A40 GPUs (48\,GB).
For the response-only soft-token pooling in \S\ref{sec:alignment}, Qwen3-8B pools both the reasoning trace and the final response. Qwen3-4B produces no \texttt{<think>} block, so its pooling range covers the entire response.

For the SOLAR alignment objective, we compute soft tokens with $\tau{=}1.0$ and top-$k{=}30$. We ablate the alignment weight $\lambda \in \{0.25, 0.50, 0.75, 1.00\}$ and soft-token temperature $\tau \in \{0.6, 1.0, 2.0\}$ in \S\ref{sec:ablation}. We employ a group-aware batch sampler that places all translations of each problem contiguously, enabling co-occurrence within the same training step even under small per-device batch sizes. 

\subsection{Evaluation setup}\label{appendix:eval}
We use \texttt{SGLang} (v0.4.6)~\cite{zheng2024sglang} as the inference backend. For all evaluations, we use a sampling temperature of $0.6$, top-$k{=}30$, top-$p{=}0.95$, and a maximum generation length of 32,768 tokens. Note that this sampling temperature is used solely for next-token decoding at inference time and is distinct from the soft-token temperature $\tau$ used during SOLAR training (Eq.~\ref{eq:soft-embed}--\ref{eq:filtered-prob}). We run each evaluation multiple times and report the average to reduce estimation variance: five runs for MGSM and GPQA, and ten runs for AIME~2024 and AIME~2025. We count a model's answer as correct only if the extracted final answer (\textit{i.e.,} the content within \texttt{\textbackslash boxed\{\}}) matches the ground-truth label.

\subsection{Datasets}\label{appendix:datasets}
\paragraph{Training data.} M-s1k~\cite{barua2025longchainofthoughtreasoninglanguages} is a multilingual long CoT reasoning dataset translated from s1k~\cite{muennighoff-etal-2025-s1} by Gemini-2.0-Flash~\cite{google_gemini2flash}. It contains 1,000 parallel problems across 10 languages (English (En), Chinese (Zh), French (Fr), Japanese (Ja), Latvian (Lv), Swahili (Sw), Telugu (Te), Thai (Th), Afrikaans (Af), and Marathi (Mr)), totaling 10,000 training examples. Each pairs a question with a detailed reasoning trace and final solution. We additionally annotate each with a \texttt{parallel\_id} linking translations of the same problem across languages and a \texttt{language} tag for fine-tuning.

\paragraph{Evaluation benchmarks.} We evaluate on four benchmarks covering multilingual mathematical reasoning and science question answering. MGSM (Multilingual Grade School Math)~\cite{shi2023language} extends GSM8K~\cite{cobbe2021trainingverifierssolvemath} with 250 parallel grade-school math word problems requiring multi-step arithmetic. AIME~2024~\cite{aime2024} and AIME~2025 each contain 30 competition-level problems from the American Invitational Mathematics Examination. GPQA~Diamond~\cite{rein2024gpqa} is a graduate-level, Google-proof benchmark of 198 multiple-choice questions across physics, chemistry, and biology, designed to challenge domain experts. The multilingual versions of AIME~2024, AIME~2025, and GPQA~Diamond are from XReasoning~\cite{qi-etal-2025-models}. All four cover 11 languages: English (En), Spanish (Es), French (Fr), German (De), Russian (Ru), Chinese (Zh), Japanese (Ja), Thai (Th), Swahili (Sw), Bengali (Bn), and Telugu (Te). Since only 7 of the 10 M-s1k languages overlap with the evaluation set (En, Fr, Ja, Sw, Te, Th, Zh), we report main results on these seven. For the representation analysis (\S\ref{sec:repr-analysis}), we sample 50 instances per language from MGSM and GPQA~Diamond and use all 30 per subset for AIME~2024 and AIME~2025; the main experiments (\S\ref{sec:experiments}) use all instances. Expanding to all 11 languages would require translating long s1k CoT traces, with non-trivial API costs and checks for mathematical fidelity. We leave this expansion to future work.

\subsection{Baselines}\label{appendix:baselines}
\paragraph{Soft Thinking.} We follow the formulation in \S\ref{sec:background} and apply inference-only soft thinking~\cite{zhang2025soft} to base models. We use the same decoding hyperparameters as our evaluation setup (\S\ref{appendix:eval}).

\paragraph{English-pivot SFT.} We fine-tune Qwen3-4B on M-s1k, pairing each non-English question with its English reasoning trace via \texttt{parallel\_id} (\S\ref{appendix:datasets}). Questions, dataset size, reasoning content, and hyperparameters match those of the SFT baseline (\S\ref{appendix:training}); only the response language changes.

\paragraph{MidAlign.} A middle-layer contrastive alignment objective that encourages cross-lingual similarity by maximizing cosine similarity between mean-pooled hidden states of parallel translation pairs while minimizing similarity to non-translations within each batch~\cite{liu-niehues-2025-middle}. We adapt MidAlign from its original LoRA-based setup to our full-parameter SFT setting and apply the contrastive alignment loss at the middle layer (18 for Qwen3-4B) jointly with the cross-entropy loss. MidAlign aligns discrete hidden states at intermediate layers, whereas SOLAR aligns soft-token embeddings.

\paragraph{MAPO.} An optimization framework that uses a pre-trained machine translation model to score the consistency between non-English and English reasoning chains, treating these scores as signals for preference training~\cite{she-etal-2024-mapo}. MAPO optimizes cross-lingual consistency at the output level via preference learning over discrete reasoning, whereas our method aligns soft-token embeddings at the representation level during SFT. For each of the 9{,}000 non-English questions in M-s1k, we sample $n{=}20$ candidate outputs from Qwen3-4B with sampling temperature $1.0$. We then score each output using NLLB-200-distilled-600M~\cite{nllbteam2022languageleftbehindscaling}. The alignment score for each non-English output is computed as $1 / \mathcal{L}_{\text{CE}}$, where $\mathcal{L}_{\text{CE}}$ is the cross-entropy loss of the NLLB model translating the non-English output to the corresponding English reference answer. Preference pairs are constructed by selecting, for each question, a chosen response that is correct and has a higher alignment score, paired with a rejected response that has a lower alignment score. This yields 6{,}305 preference pairs from 1{,}148 questions. We perform full-parameter DPO~\cite{NEURIPS2023_a85b405e} with $\beta{=}0.1$.

\paragraph{AlignX.} A two-stage framework for multilingual representation alignment~\cite{bu-etal-2025-alignx}. Stage~1 combines three objectives: (1)~an InfoNCE-style contrastive loss that aligns mean-pooled hidden states at an intermediate layer across parallel translation pairs using cosine similarity (with temperature $0.1$), (2)~a language-matching classifier at the final layer to reduce off-target generation, and (3)~the standard next-token prediction loss. Stage~2 performs multilingual instruction fine-tuning with only the NTP loss. We apply contrastive alignment at the middle layer of the model. The language-matching classifier is a 2-layer MLP with intermediate dimension 128. We use $\alpha_1{=}0.3$ and $\alpha_2{=}0.4$ as the loss weights for the contrastive and language-matching losses, respectively, following the original paper.

\subsection{Measured training cost}\label{appendix:training-cost}
We compare Qwen3-4B SFT and SOLAR under the settings of \S\ref{appendix:training}, with $4\times$A40, per-device batch size 1, accumulation over 8 micro-batches, FlashAttention-2, Liger kernels, and gradient checkpointing. The measurement run uses 20 optimizer steps with 5 warm-up steps. SFT/SOLAR peak allocated memory is 28.81/42.21 GiB ($+46.5\%$), and peak reserved memory is 34.23/43.96 GiB ($+28.4\%$), measured in-process with PyTorch as maxima across ranks. Our SOLAR implementation materializes vocabulary logits before top-$k{=}30$ filtering and computes cross-entropy separately, while SFT uses Liger's fused linear cross-entropy. This difference may increase peak memory usage.

%% file: appendix/3-additional-results.tex
\section{Experimental results}\label{appendix:experimental_results}
\paragraph{Per-language results.} Tables~\ref{tab:mgsm-per-lang}--\ref{tab:gpqa-per-lang} report accuracy and CLC across four benchmarks. SOLAR's gains over the base model are largest for low-resource languages on MGSM and AIME. Swahili MGSM gains are +24.4/+22.0 points for Qwen3-4B/8B; Telugu gains +24.0 points on AIME~2024 at 4B. On GPQA, all seven Qwen3-4B languages improve, including English. Gemma3 shows similar low-resource gains, with SOLAR leading in average accuracy and CLC at both scales on MGSM and GPQA. On MGSM, Gemma3-12B exceeds Qwen3-8B (89.50\% vs.\ 83.42\%), while Gemma3-4B trails Qwen3-4B (53.31\% vs.\ 74.01\%). Base Gemma3 GPQA accuracy is near or below 25\% chance (16.34\% at 4B, 26.61\% at 12B); SOLAR beats SFT at both scales. We omit Gemma3 from AIME~2024/2025 given the 4B model's near-chance GPQA performance. Accuracy tracks CLC across the main-comparison model variants ($r\ge0.992$ per benchmark, $r=0.996$ pooled; Table~\ref{tab:acc-clc-corr}).

\paragraph{Alignment baselines.}\label{appendix:alignment-baseline-comparison}  For Qwen3-4B, MidAlign ranks second on AIME~2024. MAPO achieves high single-language MGSM accuracy without consistent CLC gains, indicating that preference optimization need not align cross-lingual reasoning. Its English accuracy is below that of the base model on MGSM and both AIME sets, consistent with non-English-only preference training. Its limited supervision (6{,}305 pairs from 1{,}148 questions) may constrain low-resource gains.
The comparison spans two families and two scales. On Gemma3-4B, SOLAR exceeds the strongest of MidAlign, MAPO, and AlignX by 1.02 percentage points in accuracy and 0.72 in CLC on MGSM, and by 0.66 and 0.70, respectively, on GPQA~Diamond (Tables~\ref{tab:mgsm-per-lang} and~\ref{tab:gpqa-per-lang}). Three of four margins are smaller than for Qwen3-4B; MGSM accuracy is the exception (1.02 vs.\ 0.58). On Qwen3-8B, SOLAR exceeds all three baselines on every benchmark in both accuracy and CLC, leading the strongest baseline, MAPO, by 0.99 percentage points in accuracy and 0.90 in CLC on the four-benchmark average. On MGSM, it is the strongest alignment method in all seven languages, with the largest margins on the low-resource ones (Telugu $+$1.60, Thai $+$0.96, Swahili $+$0.56). At 8B, MidAlign and MAPO exceed SFT on all four benchmarks, while AlignX remains at the SFT level.

\paragraph{English-pivot SFT results.}\label{appendix:english-pivot-results} English-pivot SFT reaches 65.29\% average accuracy and 59.97\% CLC on MGSM, falling below plain SFT (76.25\% / 72.29\%) and below the untuned Qwen3-4B (74.01\% / 69.31\%) in all seven languages (Tables~\ref{tab:mgsm-per-lang-main} and~\ref{tab:mgsm-per-lang}). Swahili, where SOLAR gains most, drops to 26.24\%, below the base model's 29.04\%. Both approaches pivot on English, but at different levels: this control pivots at the \emph{data} level, replacing the native trace with an English one, whereas SOLAR pivots at the \emph{representation} level and keeps each language supervised by its own native trace. The substitution discards exactly the target-language reasoning supervision that SOLAR preserves.

\begin{table*}[!t]
\centering
\caption{Per-language accuracy, average accuracy, and cross-lingual consistency (\%) on MGSM for Qwen3 and Gemma3 models and their fine-tuned variants.}
\label{tab:mgsm-per-lang}
\footnotesize
\resizebox{\textwidth}{!}{%
\begin{tabular}{lcccccccccc}
\toprule
\textbf{Model} & \textbf{En} & \textbf{Fr} & \textbf{Ja} & \textbf{Sw} & \textbf{Te} & \textbf{Th} & \textbf{Zh} & \textbf{Avg}{\scriptsize($\uparrow$)} & \textbf{CLC}{\scriptsize($\uparrow$)}\\
\midrule
Qwen3-4B & 93.36 & 81.36 & 80.56 & 29.04 & 64.24 & \sbest{83.36} & 86.16 & 74.01 & 69.31 \\
~~+ Soft Thinking & \sbest{93.40} & 83.52 & \best{81.84} & 28.24 & 61.60 & \best{84.12} & 86.92 & 74.23 & 70.37 \\
~~+ SFT & \best{94.00} & 85.60 & 77.20 & 47.20 & 63.52 & 79.36 & 86.88 & 76.25 & 72.29 \\
~~+ English-pivot SFT & 91.20 & 78.32 & 64.16 & 26.24 & 51.76 & 70.40 & 74.96 & 65.29 & 59.97 \\
~~+ MidAlign & 93.24 & 85.16 & 79.92 & 48.72 & 63.04 & 77.60 & 87.44 & 76.45 & 73.11 \\
~~+ MAPO & 92.24 & \sbest{85.76} & 80.00 & \sbest{51.28} & \best{65.92} & 79.04 & \best{88.40} & 77.52 & \sbest{73.90} \\
~~+ AlignX & 93.12 & 84.56 & 80.32 & \best{53.48} & 64.32 & 80.40 & 87.60 & \sbest{77.69} & 73.00\\
~~+ SOLAR & 93.32 & \best{86.32} & \sbest{81.00} & \best{53.48} & \sbest{65.36} & 80.56 & \sbest{87.84} & \best{78.27} & \best{75.61} \\
\midrule
Qwen3-8B & 94.88 & 86.80 & \best{84.96} & 59.36 & 80.08 & 89.36 & 88.48 & 83.42 & 80.51 \\
~~+ Soft Thinking & \sbest{94.96} & 86.80 & \sbest{84.88} & 58.32 & 81.44 & 88.00 & 86.88 & 83.04 & 80.01 \\
~~+ SFT & \best{95.44} & 89.60 & 82.24 & 79.76 & 80.56 & 91.20 & 89.52 & 86.90 & 84.71 \\
~~+ MidAlign & 94.80 & 90.88 & 82.40 & \sbest{80.80} & 81.20 & 90.64 & 89.44 & 87.17 & 85.20 \\
~~+ MAPO & 94.56 & \sbest{91.04} & 82.16 & 80.00 & 80.96 & \sbest{91.84} & \sbest{91.44} & \sbest{87.43} & \sbest{85.68} \\
~~+ AlignX & 94.48 & 89.84 & 82.64 & 79.68 & \sbest{81.60} & 90.32 & 88.56 & 86.73 & 84.16 \\
~~+ SOLAR & \sbest{94.96} & \best{91.28} & 82.72 & \best{81.36} & \best{83.20} & \best{92.80} & \best{91.60} & \best{88.27} & \best{87.11} \\
\midrule
Gemma3-4B & 68.64 & 59.04 & 49.44 & 33.92 & 48.64 & 54.08 & 59.44 & 53.31 & 45.29 \\
~~+ Soft Thinking & 68.24 & 58.56 & 50.32 & 35.84 & 49.36 & 53.04 & 57.84 & 53.31 & 45.01 \\
~~+ SFT & 71.36 & 66.64 & 54.08 & 45.92 & 48.96 & 57.84 & 57.76 & 57.51 & 53.12\\
~~+ MidAlign & 71.36 & 67.52 & 54.24 & \sbest{47.52} & 51.28 & 58.96 & 58.72 & 58.51 & 53.28\\
~~+ MAPO & \best{72.16} & 67.20 & 54.32 & 46.88 & \sbest{51.68} & 58.16 & \sbest{62.00} & \sbest{58.91} & \sbest{54.32}\\
~~+ AlignX & \best{72.16} & \sbest{67.92} & \best{54.56} & 47.20 & 51.60 & \best{59.52} & 59.44 & \sbest{58.91} & 53.84\\
~~+ SOLAR & \sbest{71.60} & \best{68.56} & \sbest{54.40} & \best{48.40} & \best{53.36} & \sbest{59.28} & \best{63.92} & \best{59.93} & \best{55.04}\\
\midrule
Gemma3-12B & \best{97.12} & 91.36 & \best{87.84} & 83.60 & 86.64 & 89.84 & 90.08 & 89.50 & 86.95 \\
~~+ Soft Thinking & 96.48 & \sbest{91.60} & \sbest{87.44} & 84.00 & 88.48 & \sbest{90.40} & 89.28 & 89.67 & 86.75 \\
~~+ SFT & 96.80 & \best{94.16} & 87.20 & \sbest{86.88} & \best{90.80} & 89.92 & \sbest{92.32} & \sbest{91.15} & \sbest{87.76}\\
~~+ SOLAR & \sbest{97.04} & \best{94.16} & \sbest{87.44} & \best{87.52} & \sbest{89.76} & \best{91.44} & \best{93.20} & \best{91.51} & \best{89.52}\\
\bottomrule
\end{tabular}}
\end{table*}

\begin{table*}[!t]
\centering
\caption{Per-language accuracy, average accuracy, and cross-lingual consistency (\%) on AIME~2024 for Qwen3 models and their fine-tuned variants.}
\label{tab:aime-per-lang}
\footnotesize
\begin{tabular}{lcccccccccc}
\toprule
\textbf{Model} & \textbf{En} & \textbf{Fr} & \textbf{Ja} & \textbf{Sw} & \textbf{Te} & \textbf{Th} & \textbf{Zh} & \textbf{Avg}{\scriptsize($\uparrow$)} & \textbf{CLC}{\scriptsize($\uparrow$)}\\
\midrule
Qwen3-4B & \best{53.33} & 43.67 & \best{36.67} & 11.00 & 5.67 & 26.00 & 41.33 & 31.10 & 23.89 \\
~~+ Soft Thinking & \best{53.33} & \best{51.33} & 30.00 & 10.00 & 6.33 & 26.00 & \best{46.00} & 31.86 & 24.44 \\
~~+ SFT & 45.33 & 46.33 & 34.33 & 22.00 & 20.67 & 36.00 & 38.67 & 34.76 & 25.17 \\
~~+ MidAlign & 44.33 & 49.67 & \sbest{35.67} & \sbest{27.67} & \sbest{28.67} & \sbest{36.33} & 34.33 & \sbest{36.67} & \sbest{27.11} \\
~~+ MAPO & \sbest{46.33} & 45.33 & 30.67 & 22.33 & 23.00 & 36.00 & 39.67 & 34.76 & 25.89 \\
~~+ AlignX & 45.67 & 46.33 & \sbest{35.67} & 26.33 & 24.00 & 32.33 & \sbest{41.67} & 36.00 & 26.88 \\
~~+ SOLAR  & 45.33 & \sbest{50.67} & 34.33 & \best{32.67} & \best{29.67} & \best{38.67} & 38.33 & \best{38.52} & \best{29.67} \\
\midrule
Qwen3-8B & \sbest{75.33} & 73.67 & \best{70.33} & 37.33 & 52.67 & 68.67 & \sbest{56.67} & 62.10 & 56.06 \\
~~+ Soft Thinking & \best{76.00} & 73.00 & \best{70.33} & 37.67 & 54.00 & 69.67 & \best{57.00} & 62.52 & 56.91 \\
~~+ SFT & 72.00 & \sbest{77.33} & 68.33 & \sbest{55.33} & \sbest{61.67} & \sbest{72.00} & 53.00 & \sbest{65.67} & \sbest{58.67} \\
~~+ SOLAR & 72.67 & \best{78.33} & \sbest{68.67} & \best{58.33} & \best{63.00} & \best{75.00} & \sbest{56.67} & \best{67.52} & \best{60.61} \\
\bottomrule
\end{tabular}%
\end{table*}

\begin{table*}[!t]
\centering
\caption{Per-language accuracy, average accuracy, and cross-lingual consistency (\%) on AIME~2025 for Qwen3 models and their fine-tuned variants.}
\label{tab:aime2025-per-lang}
\footnotesize
\begin{tabular}{lcccccccccc}
\toprule
\textbf{Model} & \textbf{En} & \textbf{Fr} & \textbf{Ja} & \textbf{Sw} & \textbf{Te} & \textbf{Th} & \textbf{Zh} & \textbf{Avg}{\scriptsize($\uparrow$)} & \textbf{CLC}{\scriptsize($\uparrow$)}\\
\midrule
Qwen3-4B & \best{36.67} & 31.67 & 27.00 & 9.33 & 3.33 & 22.33 & 31.00 & 23.05 & 17.61 \\
~~+ Soft Thinking & 30.00 & \sbest{36.67} & \best{30.33} & 6.67 & 3.33 & 25.00 & 31.33 & 23.33 & 18.22 \\
~~+ SFT & \sbest{35.67} & 33.33 & 26.00 & 21.00 & 17.00 & 30.00 & 32.33 & 27.90 & 20.44 \\
~~+ MidAlign & 31.67 & 34.33 & \sbest{27.67} & \sbest{23.67} & \sbest{20.00} & \best{32.00} & 34.00 & \sbest{29.05} & \sbest{23.22} \\
~~+ MAPO & 32.67 & \sbest{36.67} & 24.33 & 13.33 & 9.67 & 28.67 & \sbest{35.33} & 25.81 & 18.94 \\
~~+ AlignX & \best{36.67} & 34.67 & 25.00 & 22.67 & 17.33 & 26.00 & 32.00 & 27.76 & 20.10\\
~~+ SOLAR & 35.33 & \best{37.33} & 26.67 & \best{26.33} & \best{23.33} & \sbest{31.67} & \best{36.00} & \best{30.95} & \best{24.51} \\
\midrule
Qwen3-8B & \best{63.00} & 66.67 & 60.00 & 41.67 & 43.67 & 55.00 & 54.33 & 54.91 & 45.06 \\
~~+ Soft Thinking & \sbest{61.00} & 65.67 & \best{61.00} & 40.33 & 49.00 & 53.67 & 52.33 & 54.71 & 44.22 \\
~~+ SFT  & 60.67 & \sbest{67.33} & \sbest{60.67} & \sbest{48.67} & \sbest{52.33} & \sbest{63.00} & \sbest{56.00} & \sbest{58.38} & \sbest{50.13} \\
~~+ SOLAR & \sbest{61.00} & \best{69.00} & 59.33 & \best{51.00} & \best{55.67} & \best{65.67} & \best{59.00} & \best{60.10} & \best{53.02} \\
\bottomrule
\end{tabular}%
\end{table*}

\begin{table*}[!t]
\centering
\caption{Per-language accuracy, average accuracy, and cross-lingual consistency (\%) on GPQA~Diamond for Qwen3 and Gemma3 models and their fine-tuned variants.}
\label{tab:gpqa-per-lang}
\footnotesize
\begin{tabular}{lcccccccccc}
\toprule
\textbf{Model} & \textbf{En} & \textbf{Fr} & \textbf{Ja} & \textbf{Sw} & \textbf{Te} & \textbf{Th} & \textbf{Zh} & \textbf{Avg}{\scriptsize($\uparrow$)} & \textbf{CLC}{\scriptsize($\uparrow$)}\\
\midrule
Qwen3-4B & 45.45 & 29.80 & 16.57 & 19.80 & 14.24 & 30.30 & 36.46 & 27.52 & 14.56 \\
~~+ Soft Thinking & 48.48 & 29.90 & 14.55 & 19.39 & 13.03 & 29.19 & 36.87 & 27.34 & 14.31 \\
~~+ SFT & \best{58.08} & \sbest{50.61} & \best{35.86} & \sbest{37.88} & 30.10 & \sbest{44.24} & 49.49 & \sbest{43.75} & \sbest{33.15} \\
~~+ MidAlign & 54.75 & 49.60 & 34.55 & 36.46 & \best{34.55} & 40.61 & 50.51 & 43.00 & 32.47 \\
~~+ MAPO & 47.17 & 47.58 & 34.14 & 34.65 & 30.61 & 41.41 & 46.67 & 40.32 & 30.15 \\
~~+ AlignX & \sbest{56.06} & 49.49 & \sbest{34.75} & 34.55 & \sbest{33.94} & 40.61 & \sbest{51.82} & 43.03 & 32.97\\
~~+ SOLAR & \best{58.08} & \best{53.84} & 34.24 & \best{39.09} & 33.23 & \best{44.95} & \best{53.23} & \best{45.24} & \best{36.26} \\
\midrule
Qwen3-8B & 58.38 & 57.37 & 52.93 & 45.66 & 48.69 & 54.14 & 51.62 & 52.68 & 42.69 \\
~~+ Soft Thinking & 61.31 & 57.88 & 53.03 & 44.24 & 47.68 & 54.75 & 52.53 & 53.06 & 43.42 \\
~~+ SFT & \best{69.49} & \sbest{64.14} & \sbest{58.59} & \best{62.63} & \sbest{61.31} & \sbest{62.73} & \sbest{66.57} & \sbest{63.64} & \sbest{55.73} \\
~~+ SOLAR & \sbest{68.99} & \best{66.16} & \best{61.11} & \sbest{62.02} & \best{63.23} & \best{64.65} & \best{68.59} & \best{64.96} & \best{57.27} \\
\midrule
Gemma3-4B & 17.47 & 17.88 & 14.65 & 13.84 & 14.14 & 18.48 & 17.88 & 16.34 & 5.37 \\
~~+ Soft Thinking & 16.26 & 16.97 & 16.67 & 14.85 & 14.95 & 17.58 & 16.57 & 16.26 & 5.35 \\
~~+ SFT & \best{33.74} & 31.21 & 28.59 & \sbest{27.98} & \best{30.71} & 28.59 & 27.27 & 29.73 & 15.45\\
~~+ MidAlign & \sbest{33.54} & 31.31 & 28.38 & 27.68 & \sbest{30.51} & \sbest{29.09} & \best{28.38} & \sbest{29.84} & \sbest{15.56}\\
~~+ MAPO & 32.22 & 30.81 & \sbest{28.89} & 27.07 & 28.99 & \best{29.60} & \sbest{27.68} & 29.32 & 14.85\\
~~+ AlignX & 32.83 & \sbest{31.41} & 27.37 & 25.56 & 29.09 & 28.28 & 26.87 & 28.77 & 14.14\\
~~+ SOLAR & 33.43 & \best{33.33} & \best{30.10} & \best{28.89} & 29.80 & \best{29.60} & \best{28.38} & \best{30.50} & \best{16.26}\\
\midrule
Gemma3-12B & 30.51 & 30.51 & 23.23 & 28.59 & 23.03 & 27.58 & \sbest{22.83} & 26.61 & 13.43 \\
~~+ Soft Thinking & 29.60 & 28.08 & 21.21 & 28.79 & 25.15 & 28.48 & 22.32 & 26.23 & 12.27 \\
~~+ SFT & \sbest{43.13} & \sbest{44.55} & \sbest{29.80} & \sbest{37.68} & \sbest{26.16} & \sbest{39.60} & \best{36.77} & \sbest{36.81} & \sbest{30.61}\\
~~+ SOLAR & \best{47.07} & \best{46.97} & \best{31.62} & \best{39.70} & \best{30.51} & \best{43.74} & \best{36.77} & \best{39.48} & \best{34.85}\\
\bottomrule
\end{tabular}%
\end{table*}

\begin{table*}[!t]
\centering
\caption{Pearson correlations between average accuracy and CLC across the main-comparison model variants.}
\label{tab:acc-clc-corr}
\footnotesize
\begin{tabular}{lcc}
\toprule
\textbf{Benchmark} & \textbf{Pearson $r$} & \textbf{$p$-value} \\
\midrule
MGSM      & 0.998 & $6.6 \times 10^{-31}$ \\
AIME~2024 & 0.999 & $6.9 \times 10^{-17}$ \\
AIME~2025 & 0.997 & $2.0 \times 10^{-14}$ \\
GPQA      & 0.992 & $2.6 \times 10^{-22}$ \\
\midrule
All       & 0.996 & $5.1 \times 10^{-81}$ \\
\bottomrule
\end{tabular}
\end{table*}

\paragraph{Languages not covered by M-s1k.}
Spanish (Es), German (De), and Russian (Ru) appear in all four benchmarks but not M-s1k (Appendix~\ref{appendix:datasets}); neither SFT nor SOLAR is fine-tuned on them. Results for the existing Qwen3-4B/8B checkpoints are in Table~\ref{tab:unseen-avg}, with per-language MGSM and AIME~2024 breakdowns in Tables~\ref{tab:unseen-mgsm} and~\ref{tab:unseen-aime24}.
On these unseen languages, SFT trails the base model in seven of eight model-benchmark combinations by 0.34--2.34 points, with Qwen3-4B on GPQA the only exception. SOLAR improves over SFT in seven combinations, with the largest gain on GPQA at 4B (+2.22 points). Its only decline is on AIME~2025 at 8B (58.35\% vs.\ 59.22\%). These gains suggest that alignment benefits extend to unseen languages, although SOLAR generally remains below the base model and recovers only part of the performance lost during fine-tuning.

\begin{table*}[!t]
\centering
\caption{Unseen-language accuracy (\%) on the four benchmarks, averaged over Spanish, German, and Russian, for models trained without supervision in these languages.}
\label{tab:unseen-avg}
\footnotesize
\begin{tabular}{lcccc}
\toprule
\textbf{Model} & \textbf{MGSM} & \textbf{AIME~2024} & \textbf{AIME~2025} & \textbf{GPQA} \\
\midrule
Qwen3-4B   & \best{84.00} & \best{42.22} & \best{31.44} & 36.73 \\
~~+ SFT    & 83.36 & 40.56 & 30.22 & \sbest{47.10} \\
~~+ SOLAR  & \sbest{83.68} & \sbest{41.00} & \sbest{31.00} & \best{49.32} \\
\midrule
Qwen3-8B   & \best{87.52} & \best{69.56} & \best{59.56} & \best{58.18} \\
~~+ SFT    & 85.60 & 67.22 & \sbest{59.22} & 56.37 \\
~~+ SOLAR  & \sbest{86.43} & \sbest{67.89} & 58.35 & \sbest{57.31} \\
\bottomrule
\end{tabular}
\end{table*}

\begin{table*}[!t]
\centering
\caption{Per-language MGSM accuracy (\%) on the unseen languages Es/De/Ru.}
\label{tab:unseen-mgsm}
\footnotesize
\begin{tabular}{lcccc}
\toprule
\textbf{Model} & \textbf{Es} & \textbf{De} & \textbf{Ru} & \textbf{Avg.} \\
\midrule
Qwen3-4B   & \best{85.12} & \best{83.12} & \best{83.76} & \best{84.00} \\
~~+ SFT    & 84.88 & 82.64 & 82.56 & 83.36 \\
~~+ SOLAR  & \sbest{85.04} & \sbest{83.04} & \sbest{82.96} & \sbest{83.68} \\
\midrule
Qwen3-8B   & \best{87.52} & \best{87.04} & \best{88.00} & \best{87.52} \\
~~+ SFT    & \sbest{85.60} & 85.52 & 85.68 & 85.60 \\
~~+ SOLAR  & 85.36 & \sbest{86.96} & \sbest{86.96} & \sbest{86.43} \\
\bottomrule
\end{tabular}
\end{table*}

\begin{table*}[!t]
\centering
\caption{Per-language AIME~2024 accuracy (\%) on the unseen languages Es/De/Ru.}
\label{tab:unseen-aime24}
\footnotesize
\begin{tabular}{lcccc}
\toprule
\textbf{Model} & \textbf{Es} & \textbf{De} & \textbf{Ru} & \textbf{Avg.} \\
\midrule
Qwen3-4B   & \best{42.67} & \best{41.00} & \best{43.00} & \best{42.22} \\
~~+ SFT    & 41.33 & 38.67 & \sbest{41.67} & 40.56 \\
~~+ SOLAR  & \sbest{41.67} & \sbest{39.67} & \sbest{41.67} & \sbest{41.00} \\
\midrule
Qwen3-8B   & \best{73.67} & \best{68.33} & \best{66.67} & \best{69.56} \\
~~+ SFT    & \sbest{71.00} & 64.00 & \best{66.67} & 67.22 \\
~~+ SOLAR  & 70.67 & \sbest{66.67} & \sbest{66.33} & \sbest{67.89} \\
\bottomrule
\end{tabular}
\end{table*}

%% file: appendix/4-additional-analysis.tex
\section{Analysis}\label{appendix:analysis}
\subsection{Ablation study}\label{appendix:ablation}

\paragraph{Alignment weight $\lambda$.} Table~\ref{tab:lambda-ablation-all} compares $\lambda \in \{0.25, 0.50, 0.75, 1.00\}$ on four benchmarks at $\tau{=}1.0$. Both $\lambda{=}0.25$ and $0.50$ exceed the strongest baseline in accuracy and CLC on each benchmark; their accuracies differ by at most 1.09 points, with $0.25$ leading on MGSM.
We use $\lambda{=}0.50$ for all main experiments. Parameter search used all four Qwen3-4B benchmarks without a separate validation split; choosing $0.25$ from MGSM alone is a post-hoc sensitivity check, not an independent held-out evaluation. The default configuration ($\lambda{=}0.50$, $\tau{=}1.0$, top-$k{=}30$) transfers unchanged to Qwen3-8B and Gemma3-4B/12B (Appendix~\ref{appendix:training}).
Larger weights hurt performance: $\lambda{=}0.75$ trails MAPO and AlignX on MGSM, while $\lambda{=}1.00$ falls below the base model (69.25\% vs.\ 74.01\%).

\paragraph{Soft-token temperature $\tau$.} Table~\ref{tab:secondary-ablations} (a) compares $\tau \in \{0.6, 1.0, 2.0\}$ at $\lambda{=}0.50$. Lower $\tau$ sharpens soft-token distributions, while higher $\tau$ smooths them. At $\tau{=}0.6$, average accuracy and CLC fall by 17.8 and 20.2 points relative to $\tau{=}1.0$, with Telugu accuracy dropping to 14.48\%. At $\tau{=}2.0$, the losses are smaller ($-$1.9 accuracy, $-$2.5 CLC), although Swahili and Telugu reach their best accuracies across the three temperatures. At $\tau{=}1.0$, soft tokens may balance distinctness from discrete embeddings with the semantic specificity needed to compare parallel responses across languages.

\paragraph{Alignment representation space.} Table~\ref{tab:secondary-ablations} (b) compares aligning mean-pooled soft-token embeddings in $\mathbb{R}^d$ (SOLAR) against mean-pooled probability distributions in $\mathbb{R}^{|V|}$, which bypasses the embedding layer and minimizes cosine distance between $\bar{p}_{\text{en}}$ and $\bar{p}_{\ell}$ (Eq.~\ref{eq:filtered-prob}). Embedding-space alignment wins by a wide margin: in probability space accuracy drops 18.0 points and CLC 30.9 points, falling below SFT and the base model. The especially severe CLC drop may reflect largely disjoint token subsets activated by equivalent reasoning across languages in $\mathbb{R}^{|V|}$, which could make cosine distance less informative for alignment. The embedding projection may instead capture cross-script correspondences in a shared space. We leave alternative distance metrics, such as KL and Jensen--Shannon divergence, to future work.

\paragraph{Pivot language $\ell^*$.} Table~\ref{tab:secondary-ablations} (c) reports per-language results for Qwen3-4B trained with different pivot languages. All other hyperparameters are fixed at $\lambda{=}0.50$ and $\tau{=}1.0$.
English yields the best aggregate performance (78.27\% accuracy, 75.61\% CLC), followed by Chinese (74.50\%, 70.97\%) and French (72.87\%, 66.67\%). The performance gap is concentrated in low-resource languages, where Swahili drops from 53.48\% (En) to 40.56\% (Zh) and 36.80\% (Fr), consistent with the hypothesis that transfer effectiveness depends on the quality of the pivot's pre-trained representations. English, as the dominant language in pre-training corpora, provides the strongest anchor for alignment.

\begin{table*}[!t]
\centering
\caption{Per-language accuracy, average accuracy, and cross-lingual consistency (\%) across four benchmarks for Qwen3-4B SOLAR fine-tuned variants with different $\lambda$.}
\label{tab:lambda-ablation-all}
\footnotesize
\begin{tabular}{llcccccccccc}
\toprule
\textbf{Benchmark} & \textbf{$\lambda$} & \textbf{En} & \textbf{Fr} & \textbf{Ja} & \textbf{Sw} & \textbf{Te} & \textbf{Th} & \textbf{Zh} & \textbf{Avg.}{\scriptsize($\uparrow$)} & \textbf{CLC}{\scriptsize($\uparrow$)}\\
\midrule
\multirow{4}{*}{MGSM}
& 0.25 & \sbest{92.56} & 85.92 & \best{83.44} & 52.88 & \best{65.68} & \sbest{80.16} & \best{89.60} & \best{78.61} & \best{75.73} \\
& 0.50 & \best{93.32} & \best{86.32} & \sbest{81.00} & \sbest{53.48} & \sbest{65.36} & \best{80.56} & \sbest{87.84} & \sbest{78.27} & \sbest{75.61} \\
& 0.75 & 92.40 & \sbest{86.00} & 76.72 & \best{53.52} & 62.72 & 74.24 & 85.20 & 75.83 & 73.01 \\
& 1.00 & 87.52 & 83.76 & 72.32 & 42.56 & 53.68 & 64.80 & 80.08 & 69.25 & 63.67 \\
\cmidrule(lr){1-11}
\multirow{4}{*}{AIME~2024}
& 0.25 & \sbest{45.00} & \best{51.33} & \sbest{32.33} & \sbest{30.67} & \sbest{27.33} & \sbest{35.00} & \best{40.33} & \sbest{37.43} & \sbest{27.83} \\
& 0.50 & \best{45.33} & \sbest{50.67} & \best{34.33} & \best{32.67} & \best{29.67} & \best{38.67} & \sbest{38.33} & \best{38.52} & \best{29.67} \\
& 0.75 & 42.33 & 46.33 & 31.67 & 24.33 & 17.33 & 31.00 & 33.67 & 32.38 & 24.44 \\
& 1.00 & 40.33 & 41.67 & 30.33 & 10.00 & 3.00 & 19.33 & 32.33 & 25.28 & 19.22 \\
\cmidrule(lr){1-11}
\multirow{4}{*}{AIME~2025}
& 0.25 & \sbest{35.00} & \sbest{36.00} & \best{29.33} & \best{26.33} & \sbest{21.33} & \sbest{29.00} & \sbest{35.67} & \sbest{30.38} & \sbest{23.81} \\
& 0.50 & \best{35.33} & \best{37.33} & \sbest{26.67} & \best{26.33} & \best{23.33} & \best{31.67} & \best{36.00} & \best{30.95} & \best{24.51} \\
& 0.75 & 32.67 & 31.67 & 22.33 & \sbest{20.67} & 16.00 & 26.00 & 32.67 & 26.00 & 20.06 \\
& 1.00 & 30.00 & 28.33 & 20.67 & 9.00 & 7.00 & 21.67 & 25.67 & 20.33 & 15.83 \\
\cmidrule(lr){1-11}
\multirow{4}{*}{GPQA}
& 0.25 & \best{58.69} & \sbest{53.43} & \sbest{33.13} & \sbest{35.35} & \sbest{32.12} & \best{45.86} & \sbest{50.51} & \sbest{44.16} & \sbest{35.47} \\
& 0.50 & \sbest{58.08} & \best{53.84} & \best{34.24} & \best{39.09} & \best{33.23} & \sbest{44.95} & \best{53.23} & \best{45.24} & \best{36.26} \\
& 0.75 & 55.25 & 49.90 & 28.28 & 34.44 & 25.56 & 37.98 & 50.20 & 40.23 & 31.87 \\
& 1.00 & 50.51 & 39.80 & 20.61 & 18.69 & 14.04 & 27.27 & 33.23 & 29.16 & 17.02 \\
\bottomrule
\end{tabular}
\end{table*}

\begin{table*}[!t]
\centering
\caption{Per-language results for the secondary ablations of SOLAR with Qwen3-4B on MGSM: soft-token temperature $\tau$, representation space, and pivot language $\ell^*$.}
\label{tab:secondary-ablations}
\footnotesize
\begin{tabular}{lccccccccc}
\toprule
\textbf{Configuration} & \textbf{En} & \textbf{Fr} & \textbf{Ja} & \textbf{Sw} & \textbf{Te} & \textbf{Th} & \textbf{Zh} & \textbf{Avg.}{\scriptsize($\uparrow$)} & \textbf{CLC}{\scriptsize($\uparrow$)}\\
\midrule
\multicolumn{10}{l}{\textit{(a) Soft-token temperature $\tau$}} \\
\quad 0.6 & 89.90 & 75.76 & 70.72 & 43.36 & 14.48 & 62.48 & 66.64 & 60.48 & 55.43 \\
\quad 1.0 & \best{93.32} & \best{86.32} & \best{81.00} & \sbest{53.48} & \sbest{65.36} & \best{80.56} & \best{87.84} & \best{78.27} & \best{75.61} \\
\quad 2.0 & \sbest{91.44} & \sbest{79.28} & \sbest{79.76} & \best{56.32} & \best{67.84} & \sbest{74.08} & \sbest{86.16} & \sbest{76.41} & \sbest{73.13} \\
\midrule
\multicolumn{10}{l}{\textit{(b) Representation space}} \\
\quad $\mathbb{R}^d$ (emb.) & \best{93.32} & \best{86.32} & \best{81.00} & \best{53.48} & \best{65.36} & \best{80.56} & \best{87.84} & \best{78.27} & \best{75.61}\\
\quad $\mathbb{R}^{|V|}$ (prob.) & \sbest{89.28} & \sbest{75.07} & \sbest{64.64} & \sbest{33.44} & \sbest{20.88} & \sbest{66.56} & \sbest{72.00} & \sbest{60.27} & \sbest{44.76} \\
\midrule
\multicolumn{10}{l}{\textit{(c) Pivot language $\ell^*$}} \\
\quad English (En) & \best{93.32} & \best{86.32} & \best{81.00} & \best{53.48} & \best{65.36} & \best{80.56} & \sbest{87.84} & \best{78.27} & \best{75.61}\\
\quad Chinese (Zh) & \sbest{89.68} & 80.80 & \sbest{79.84} & \sbest{40.56} & 62.40 & \sbest{80.08} & \best{88.16} & \sbest{74.50} & \sbest{70.97}\\
\quad French (Fr) & 84.24 & \sbest{82.72} & 79.68 & 36.80 & \sbest{64.08} & 78.48 & 84.08 & 72.87 & 66.67\\
\bottomrule
\end{tabular}
\end{table*}

\begin{table*}[!t]
\centering
\caption{Alignment-target ablation for SOLAR with Qwen3-4B and Qwen3-8B.}
\label{tab:ablation-target-full}
\resizebox{\textwidth}{!}{%
\begin{tabular}{lcccccccccc}
\toprule
& \multicolumn{2}{c}{\textbf{MGSM}} & \multicolumn{2}{c}{\textbf{AIME~2024}} & \multicolumn{2}{c}{\textbf{AIME~2025}} & \multicolumn{2}{c}{\textbf{GPQA}} & \multicolumn{2}{c}{\textbf{Avg.}}\\
\cmidrule(lr){2-3} \cmidrule(lr){4-5} \cmidrule(lr){6-7} \cmidrule(lr){8-9} \cmidrule(lr){10-11}
\textbf{Alignment target} & \textbf{Accuracy} & \textbf{CLC} & \textbf{Accuracy} & \textbf{CLC} & \textbf{Accuracy} & \textbf{CLC} & \textbf{Accuracy} & \textbf{CLC} & \textbf{Accuracy} & \textbf{CLC}\\
\midrule
Qwen3-4B $+$ SFT      & 76.25 & 72.29 & 34.76 & 25.17 & 27.90 & 20.44 & 43.75 & 33.15 & 45.67 & 37.76\\
~~+ Discrete          & 76.37 & 73.50 & 35.63 & 26.06 & \sbest{28.12} & 21.41 & 43.86 & 33.47 & 46.00 & 38.61\\
~~+ Final-hidden      & 76.97 & 73.79 & 35.47 & 25.91 & 27.87 & 20.05 & \sbest{44.05} & \sbest{34.51} & 46.09 & 38.57\\
~~+ Untruncated       & \sbest{77.19} & \sbest{74.72} & \sbest{37.03} & \sbest{27.63} & 27.93 & \sbest{22.93} & 44.04 & 34.46 & \sbest{46.55} & \sbest{39.94}\\
~~+ SOLAR ($k{=}30$)  & \best{78.27} & \best{75.61} & \best{38.52} & \best{29.67} & \best{30.95} & \best{24.51} & \best{45.24} & \best{36.26} & \best{48.25} & \best{41.51}\\
\midrule
Qwen3-8B $+$ SFT      & 86.90 & 84.71 & 65.67 & 58.67 & 58.38 & 50.13 & 63.64 & 55.73 & 68.65 & 62.31\\
~~+ Discrete          & 87.29 & 85.45 & 65.00 & 56.89 & 58.86 & 51.05 & 62.99 & 55.04 & 68.54 & 62.11\\
~~+ Final-hidden      & 87.05 & 85.09 & 66.51 & 58.89 & 59.06 & 51.12 & 63.41 & 55.65 & 69.01 & 62.69\\
~~+ Untruncated       & \sbest{87.92} & \sbest{85.73} & \sbest{66.94} & \sbest{59.02} & \sbest{60.08} & \sbest{51.94} & \sbest{64.09} & \sbest{56.13} & \sbest{69.76} & \sbest{63.21}\\
~~+ SOLAR ($k{=}30$)  & \best{88.27} & \best{87.11} & \best{67.52} & \best{60.61} & \best{60.10} & \best{53.02} & \best{64.96} & \best{57.27} & \best{70.21} & \best{64.50}\\
\bottomrule
\end{tabular}}
\end{table*}

\paragraph{Full alignment-target results.}
Table~\ref{tab:ablation-target-full} extends the alignment-target ablation of \S\ref{sec:ablation-design} to Qwen3-8B alongside Qwen3-4B. At both scales SOLAR is the best target and a single hard token (Discrete) stays at the SFT level, while the intermediate targets (Final-hidden, Untruncated) fall within about a point of one another and are not strictly ordered across benchmarks. The consistent SOLAR-best pattern across scales indicates that the soft mixture, rather than model size, drives the gain.

\subsection{Representation analysis}\label{appendix:repr-analysis}
\paragraph{Setup.} Each hidden state $\mathbf{h}_i^\ell$ is mean-pooled over every token of the full chat-templated sequence, namely the input question followed by the teacher-forced gold final answer, with prompt tokens included; it is obtained from a single teacher-forced forward pass, so no model-generated text enters the pooled span. To analyze cross-lingual representation similarity~\citep{lim2025languagespecific}, we use the same pooling definition for Qwen3-4B and Qwen3-8B. We sample 50 instances per language for MGSM and GPQA and all 30 from each AIME subset. We extract hidden states in two passes. Pass~1 runs a standard forward pass, mean-pools hidden states at each target layer, and builds soft tokens via Eq.~\ref{eq:soft-embed} with $\tau = 0.6$ and $k = 30$. Pass~2 feeds these soft tokens back into the same model and extracts hidden states at the same layers. We probe three depths: early (layer~1), middle ($\sim$50\%), and final (layer~$-1$), reporting $\Delta$CLS and $\Delta$SS relative to discrete tokens. Positive $\Delta$CLS or negative $\Delta$SS indicates improved cross-lingual mixing.

\paragraph{Results.} Tables~\ref{tab:full-mgsm}--\ref{tab:full-gpqa-dpsk} present the layer-wise cross-lingual alignment analysis for all 10 models across four benchmarks and 11 languages. The inverted-U pattern from \S\ref{sec:repr-analysis} holds across all models and benchmarks: CLS peaks at middle layers and drops sharply at the final layer. Soft thinking improves final-layer CLS for most models and benchmarks, with magnitude varying by family: Gemma3 shows the largest gains ($\Delta$CLS up to +0.68), Qwen3 improves more modestly, and DeepSeek-R1-Distill shows the smallest effect, with small negative $\Delta$CLS for DeepSeek-R1-Distill-Llama-8B on AIME~2024/2025 and GPQA.
Table~\ref{tab:repr-after-training-combined} reports CLS and SS at three depths for Qwen3-4B, Qwen3-8B, and their SFT and SOLAR variants, extending the final-layer summary in \S\ref{sec:repr-analysis} to early and middle layers. The 8B results corroborate the 4B patterns, with gains concentrated at the final layer: SOLAR raises final-layer discrete CLS from 0.23 to 0.76 ($\Delta$CLS\,=\,+0.53), and soft thinking lifts it further to 0.87 (+0.64). For 4B, SOLAR's discrete gain (+0.36) exceeds its cumulative soft-thinking gain (+0.28), so SOLAR has largely internalized what soft thinking provides. On 8B, soft thinking still adds on top of SOLAR (+0.64 vs.\ +0.53), suggesting the larger model retains capacity for further inference-time alignment. CLS gains are concentrated at the final layer, although SS can worsen at the middle layer, as observed for Qwen3-4B under soft thinking.

\begin{table*}[!t]
\centering
\caption{Cross-lingual alignment on MGSM, discrete vs.\ soft thinking (Qwen3).}
\label{tab:full-mgsm}
\resizebox{\textwidth}{!}{%
\begin{tabular}{lcccccccccccc}
\toprule
& \multicolumn{3}{c}{\textbf{Qwen3-4B-Base}} & \multicolumn{3}{c}{\textbf{Qwen3-4B}} & \multicolumn{3}{c}{\textbf{Qwen3-8B-Base}} & \multicolumn{3}{c}{\textbf{Qwen3-8B}} \\
\cmidrule(lr){2-4} \cmidrule(lr){5-7} \cmidrule(lr){8-10} \cmidrule(lr){11-13}
\textbf{Metric} & \textbf{Early} & \textbf{Mid} & \textbf{Final} & \textbf{Early} & \textbf{Mid} & \textbf{Final} & \textbf{Early} & \textbf{Mid} & \textbf{Final} & \textbf{Early} & \textbf{Mid} & \textbf{Final} \\
\midrule
Disc-CLS{\scriptsize($\uparrow$)} & 0.84 & 0.98 & 0.28 & 0.78 & 0.94 & 0.14 & 0.85 & 0.94 & 0.73 & 0.82 & 0.97 & 0.23 \\
Soft-CLS{\scriptsize($\uparrow$)} & 0.83 & 0.96 & 0.43 & 0.78 & 0.92 & 0.24 & 0.85 & 0.91 & 0.79 & 0.83 & 0.94 & 0.36 \\
$\Delta$CLS & \dbad{-0.01} & \dbad{-0.02} & \dgood{+0.15} & \dgood{+0.00} & \dbad{-0.02} & \dgood{+0.10} & \dgood{+0.00} & \dbad{-0.03} & \dgood{+0.06} & \dgood{+0.01} & \dbad{-0.03} & \dgood{+0.13} \\
\midrule
Disc-SS{\scriptsize($\downarrow$)} & 0.53 & 0.05 & 0.84 & 0.58 & 0.23 & 0.89 & 0.50 & 0.14 & 0.82 & 0.53 & 0.15 & 0.87 \\
Soft-SS{\scriptsize($\downarrow$)} & 0.60 & 0.13 & 0.83 & 0.67 & 0.32 & 0.88 & 0.57 & 0.10 & 0.77 & 0.60 & 0.27 & 0.83 \\
$\Delta$SS & \dbad{+0.07} & \dbad{+0.08} & \dgood{-0.01} & \dbad{+0.09} & \dbad{+0.09} & \dgood{-0.01} & \dbad{+0.07} & \dgood{-0.04} & \dgood{-0.05} & \dbad{+0.07} & \dbad{+0.12} & \dgood{-0.04} \\
\bottomrule
\end{tabular}}
\end{table*}

\begin{table*}[!t]
\centering
\caption{Cross-lingual alignment on AIME~2024, discrete vs.\ soft thinking (Qwen3).}
\label{tab:full-aime2024}
\resizebox{\textwidth}{!}{%
\begin{tabular}{lcccccccccccc}
\toprule
& \multicolumn{3}{c}{\textbf{Qwen3-4B-Base}} & \multicolumn{3}{c}{\textbf{Qwen3-4B}} & \multicolumn{3}{c}{\textbf{Qwen3-8B-Base}} & \multicolumn{3}{c}{\textbf{Qwen3-8B}} \\
\cmidrule(lr){2-4} \cmidrule(lr){5-7} \cmidrule(lr){8-10} \cmidrule(lr){11-13}
\textbf{Metric} & \textbf{Early} & \textbf{Mid} & \textbf{Final} & \textbf{Early} & \textbf{Mid} & \textbf{Final} & \textbf{Early} & \textbf{Mid} & \textbf{Final} & \textbf{Early} & \textbf{Mid} & \textbf{Final} \\
\midrule
Disc-CLS{\scriptsize($\uparrow$)} & 0.88 & 0.98 & 0.42 & 0.87 & 0.96 & 0.39 & 0.89 & 0.95 & 0.85 & 0.89 & 0.98 & 0.45 \\
Soft-CLS{\scriptsize($\uparrow$)} & 0.87 & 0.96 & 0.58 & 0.86 & 0.95 & 0.51 & 0.89 & 0.91 & 0.90 & 0.90 & 0.95 & 0.60 \\
$\Delta$CLS & \dbad{-0.01} & \dbad{-0.02} & \dgood{+0.16} & \dbad{-0.01} & \dbad{-0.01} & \dgood{+0.12} & \dgood{+0.00} & \dbad{-0.04} & \dgood{+0.05} & \dgood{+0.01} & \dbad{-0.03} & \dgood{+0.15} \\
\midrule
Disc-SS{\scriptsize($\downarrow$)} & 0.42 & -0.05 & 0.70 & 0.42 & 0.01 & 0.70 & 0.39 & 0.04 & 0.48 & 0.38 & -0.06 & 0.65 \\
Soft-SS{\scriptsize($\downarrow$)} & 0.45 & 0.01 & 0.68 & 0.46 & 0.08 & 0.67 & 0.41 & 0.01 & 0.48 & 0.40 & -0.02 & 0.62 \\
$\Delta$SS & \dbad{+0.03} & \dbad{+0.06} & \dgood{-0.02} & \dbad{+0.04} & \dbad{+0.07} & \dgood{-0.03} & \dbad{+0.02} & \dgood{-0.03} & \dgood{+0.00} & \dbad{+0.02} & \dbad{+0.04} & \dgood{-0.03} \\
\bottomrule
\end{tabular}}
\end{table*}

\begin{table*}[!t]
\centering
\caption{Cross-lingual alignment on AIME~2025, discrete vs.\ soft thinking (Qwen3).}
\label{tab:full-aime2025}
\resizebox{\textwidth}{!}{%
\begin{tabular}{lcccccccccccc}
\toprule
& \multicolumn{3}{c}{\textbf{Qwen3-4B-Base}} & \multicolumn{3}{c}{\textbf{Qwen3-4B}} & \multicolumn{3}{c}{\textbf{Qwen3-8B-Base}} & \multicolumn{3}{c}{\textbf{Qwen3-8B}} \\
\cmidrule(lr){2-4} \cmidrule(lr){5-7} \cmidrule(lr){8-10} \cmidrule(lr){11-13}
\textbf{Metric} & \textbf{Early} & \textbf{Mid} & \textbf{Final} & \textbf{Early} & \textbf{Mid} & \textbf{Final} & \textbf{Early} & \textbf{Mid} & \textbf{Final} & \textbf{Early} & \textbf{Mid} & \textbf{Final} \\
\midrule
Disc-CLS{\scriptsize($\uparrow$)} & 0.88 & 0.98 & 0.45 & 0.88 & 0.96 & 0.43 & 0.90 & 0.95 & 0.87 & 0.90 & 0.97 & 0.48 \\
Soft-CLS{\scriptsize($\uparrow$)} & 0.88 & 0.96 & 0.59 & 0.87 & 0.95 & 0.54 & 0.90 & 0.92 & 0.91 & 0.90 & 0.94 & 0.63 \\
$\Delta$CLS & \dgood{+0.00} & \dbad{-0.02} & \dgood{+0.14} & \dbad{-0.01} & \dbad{-0.01} & \dgood{+0.11} & \dgood{+0.00} & \dbad{-0.03} & \dgood{+0.04} & \dgood{+0.00} & \dbad{-0.03} & \dgood{+0.15} \\
\midrule
Disc-SS{\scriptsize($\downarrow$)} & 0.40 & -0.13 & 0.69 & 0.41 & 0.00 & 0.70 & 0.38 & 0.02 & 0.50 & 0.38 & -0.09 & 0.65 \\
Soft-SS{\scriptsize($\downarrow$)} & 0.43 & -0.04 & 0.68 & 0.46 & 0.08 & 0.66 & 0.41 & -0.01 & 0.50 & 0.40 & -0.04 & 0.61 \\
$\Delta$SS & \dbad{+0.03} & \dbad{+0.09} & \dgood{-0.01} & \dbad{+0.05} & \dbad{+0.08} & \dgood{-0.04} & \dbad{+0.03} & \dgood{-0.03} & \dgood{+0.00} & \dbad{+0.02} & \dbad{+0.05} & \dgood{-0.04} \\
\bottomrule
\end{tabular}}
\end{table*}

\begin{table*}[!t]
\caption{Cross-lingual alignment on GPQA, discrete vs.\ soft thinking (Qwen3).}
\label{tab:full-gpqa}
\centering
\resizebox{\textwidth}{!}{%
\begin{tabular}{lcccccccccccc}
\toprule
& \multicolumn{3}{c}{\textbf{Qwen3-4B-Base}} & \multicolumn{3}{c}{\textbf{Qwen3-4B}} & \multicolumn{3}{c}{\textbf{Qwen3-8B-Base}} & \multicolumn{3}{c}{\textbf{Qwen3-8B}} \\
\cmidrule(lr){2-4} \cmidrule(lr){5-7} \cmidrule(lr){8-10} \cmidrule(lr){11-13}
\textbf{Metric} & \textbf{Early} & \textbf{Mid} & \textbf{Final} & \textbf{Early} & \textbf{Mid} & \textbf{Final} & \textbf{Early} & \textbf{Mid} & \textbf{Final} & \textbf{Early} & \textbf{Mid} & \textbf{Final} \\
\midrule
Disc-CLS{\scriptsize($\uparrow$)} & 0.88 & 0.97 & 0.39 & 0.87 & 0.95 & 0.37 & 0.89 & 0.95 & 0.83 & 0.90 & 0.97 & 0.37 \\
Soft-CLS{\scriptsize($\uparrow$)} & 0.87 & 0.96 & 0.54 & 0.86 & 0.94 & 0.46 & 0.89 & 0.92 & 0.88 & 0.90 & 0.94 & 0.54 \\
$\Delta$CLS & \dbad{-0.01} & \dbad{-0.01} & \dgood{+0.15} & \dbad{-0.01} & \dbad{-0.01} & \dgood{+0.09} & \dgood{+0.00} & \dbad{-0.03} & \dgood{+0.05} & \dgood{+0.00} & \dbad{-0.03} & \dgood{+0.17} \\
\midrule
Disc-SS{\scriptsize($\downarrow$)} & 0.47 & -0.00 & 0.72 & 0.48 & 0.07 & 0.73 & 0.44 & 0.08 & 0.60 & 0.44 & -0.01 & 0.70 \\
Soft-SS{\scriptsize($\downarrow$)} & 0.51 & 0.06 & 0.71 & 0.53 & 0.15 & 0.72 & 0.47 & 0.12 & 0.62 & 0.46 & 0.06 & 0.66 \\
$\Delta$SS & \dbad{+0.04} & \dbad{+0.06} & \dgood{-0.01} & \dbad{+0.05} & \dbad{+0.08} & \dgood{-0.01} & \dbad{+0.03} & \dbad{+0.04} & \dbad{+0.02} & \dbad{+0.02} & \dbad{+0.07} & \dgood{-0.04} \\
\bottomrule
\end{tabular}}
\end{table*}

\begin{table*}[!t]
\centering
\caption{Cross-lingual alignment on MGSM, discrete vs.\ soft thinking (Gemma3).}
\label{tab:full-mgsm-gemma}
\resizebox{\textwidth}{!}{%
\begin{tabular}{lcccccccccccc}
\toprule
& \multicolumn{3}{c}{\textbf{Gemma3-1B}} & \multicolumn{3}{c}{\textbf{Gemma3-4B}} & \multicolumn{3}{c}{\textbf{Gemma3-12B}} & \multicolumn{3}{c}{\textbf{Gemma3-27B}} \\
\cmidrule(lr){2-4} \cmidrule(lr){5-7} \cmidrule(lr){8-10} \cmidrule(lr){11-13}
\textbf{Metric} & \textbf{Early} & \textbf{Mid} & \textbf{Final} & \textbf{Early} & \textbf{Mid} & \textbf{Final} & \textbf{Early} & \textbf{Mid} & \textbf{Final} & \textbf{Early} & \textbf{Mid} & \textbf{Final} \\
\midrule
Disc-CLS{\scriptsize($\uparrow$)} & 0.98 & 1.00 & 0.27 & 0.99 & 1.00 & 0.28 & 0.99 & 1.00 & 0.25 & 0.99 & 1.00 & 0.30 \\
Soft-CLS{\scriptsize($\uparrow$)} & 0.96 & 1.00 & 0.75 & 0.99 & 1.00 & 0.90 & 0.93 & 1.00 & 0.93 & 0.95 & 1.00 & 0.90 \\
$\Delta$CLS & \dbad{-0.02} & \dgood{+0.00} & \dgood{+0.48} & \dgood{+0.00} & \dgood{+0.00} & \dgood{+0.62} & \dbad{-0.06} & \dgood{+0.00} & \dgood{+0.68} & \dbad{-0.04} & \dgood{+0.00} & \dgood{+0.60} \\
\midrule
Disc-SS{\scriptsize($\downarrow$)} & 0.41 & 0.23 & 0.87 & 0.26 & 0.07 & 0.84 & 0.31 & 0.21 & 0.82 & -0.03 & -0.15 & 0.82 \\
Soft-SS{\scriptsize($\downarrow$)} & 0.60 & 0.26 & 0.34 & 0.55 & 0.15 & 0.22 & 0.54 & 0.08 & 0.09 & 0.56 & 0.12 & 0.13 \\
$\Delta$SS & \dbad{+0.19} & \dbad{+0.03} & \dgood{-0.53} & \dbad{+0.29} & \dbad{+0.08} & \dgood{-0.62} & \dbad{+0.23} & \dgood{-0.13} & \dgood{-0.73} & \dbad{+0.59} & \dbad{+0.27} & \dgood{-0.69} \\
\bottomrule
\end{tabular}}
\end{table*}

\begin{table*}[!t]
\centering
\caption{Cross-lingual alignment on AIME~2024, discrete vs.\ soft thinking (Gemma3).}
\label{tab:full-aime2024-gemma}
\resizebox{\textwidth}{!}{%
\begin{tabular}{lcccccccccccc}
\toprule
& \multicolumn{3}{c}{\textbf{Gemma3-1B}} & \multicolumn{3}{c}{\textbf{Gemma3-4B}} & \multicolumn{3}{c}{\textbf{Gemma3-12B}} & \multicolumn{3}{c}{\textbf{Gemma3-27B}} \\
\cmidrule(lr){2-4} \cmidrule(lr){5-7} \cmidrule(lr){8-10} \cmidrule(lr){11-13}
\textbf{Metric} & \textbf{Early} & \textbf{Mid} & \textbf{Final} & \textbf{Early} & \textbf{Mid} & \textbf{Final} & \textbf{Early} & \textbf{Mid} & \textbf{Final} & \textbf{Early} & \textbf{Mid} & \textbf{Final} \\
\midrule
Disc-CLS{\scriptsize($\uparrow$)} & 0.99 & 1.00 & 0.44 & 1.00 & 1.00 & 0.46 & 0.99 & 1.00 & 0.42 & 0.99 & 1.00 & 0.46 \\
Soft-CLS{\scriptsize($\uparrow$)} & 0.98 & 1.00 & 0.83 & 0.99 & 1.00 & 0.93 & 0.96 & 1.00 & 0.94 & 0.97 & 1.00 & 0.92 \\
$\Delta$CLS & \dbad{-0.01} & \dgood{+0.00} & \dgood{+0.39} & \dbad{-0.01} & \dgood{+0.00} & \dgood{+0.47} & \dbad{-0.03} & \dgood{+0.00} & \dgood{+0.52} & \dbad{-0.02} & \dgood{+0.00} & \dgood{+0.46} \\
\midrule
Disc-SS{\scriptsize($\downarrow$)} & 0.19 & 0.09 & 0.66 & 0.14 & 0.01 & 0.67 & 0.13 & 0.08 & 0.67 & -0.09 & -0.19 & 0.69 \\
Soft-SS{\scriptsize($\downarrow$)} & 0.26 & 0.10 & 0.14 & 0.21 & 0.04 & 0.05 & 0.27 & -0.01 & 0.03 & 0.24 & 0.01 & -0.01 \\
$\Delta$SS & \dbad{+0.07} & \dbad{+0.01} & \dgood{-0.52} & \dbad{+0.07} & \dbad{+0.03} & \dgood{-0.62} & \dbad{+0.14} & \dgood{-0.09} & \dgood{-0.64} & \dbad{+0.33} & \dbad{+0.20} & \dgood{-0.70} \\
\bottomrule
\end{tabular}}
\end{table*}

\begin{table*}[!t]
\centering
\caption{Cross-lingual alignment on AIME~2025, discrete vs.\ soft thinking (Gemma3).}
\label{tab:full-aime2025-gemma}
\resizebox{\textwidth}{!}{%
\begin{tabular}{lcccccccccccc}
\toprule
& \multicolumn{3}{c}{\textbf{Gemma3-1B}} & \multicolumn{3}{c}{\textbf{Gemma3-4B}} & \multicolumn{3}{c}{\textbf{Gemma3-12B}} & \multicolumn{3}{c}{\textbf{Gemma3-27B}} \\
\cmidrule(lr){2-4} \cmidrule(lr){5-7} \cmidrule(lr){8-10} \cmidrule(lr){11-13}
\textbf{Metric} & \textbf{Early} & \textbf{Mid} & \textbf{Final} & \textbf{Early} & \textbf{Mid} & \textbf{Final} & \textbf{Early} & \textbf{Mid} & \textbf{Final} & \textbf{Early} & \textbf{Mid} & \textbf{Final} \\
\midrule
Disc-CLS{\scriptsize($\uparrow$)} & 0.99 & 1.00 & 0.46 & 1.00 & 1.00 & 0.48 & 0.99 & 1.00 & 0.43 & 0.99 & 1.00 & 0.48 \\
Soft-CLS{\scriptsize($\uparrow$)} & 0.98 & 1.00 & 0.85 & 0.99 & 1.00 & 0.93 & 0.96 & 1.00 & 0.93 & 0.97 & 1.00 & 0.92 \\
$\Delta$CLS & \dbad{-0.01} & \dgood{+0.00} & \dgood{+0.39} & \dbad{-0.01} & \dgood{+0.00} & \dgood{+0.45} & \dbad{-0.03} & \dgood{+0.00} & \dgood{+0.50} & \dbad{-0.02} & \dgood{+0.00} & \dgood{+0.44} \\
\midrule
Disc-SS{\scriptsize($\downarrow$)} & 0.15 & 0.08 & 0.65 & 0.12 & 0.00 & 0.68 & 0.10 & 0.08 & 0.69 & -0.10 & -0.16 & 0.70 \\
Soft-SS{\scriptsize($\downarrow$)} & 0.25 & 0.07 & 0.10 & 0.18 & 0.03 & 0.02 & 0.27 & -0.03 & -0.04 & 0.23 & -0.00 & -0.00 \\
$\Delta$SS & \dbad{+0.10} & \dgood{-0.01} & \dgood{-0.55} & \dbad{+0.06} & \dbad{+0.03} & \dgood{-0.66} & \dbad{+0.17} & \dgood{-0.11} & \dgood{-0.73} & \dbad{+0.33} & \dbad{+0.16} & \dgood{-0.70} \\
\bottomrule
\end{tabular}}
\end{table*}

\begin{table*}[!t]
\centering
\caption{Cross-lingual alignment on GPQA, discrete vs.\ soft thinking (Gemma3).}
\label{tab:full-gpqa-gemma}
\resizebox{\textwidth}{!}{%
\begin{tabular}{lcccccccccccc}
\toprule
& \multicolumn{3}{c}{\textbf{Gemma3-1B}} & \multicolumn{3}{c}{\textbf{Gemma3-4B}} & \multicolumn{3}{c}{\textbf{Gemma3-12B}} & \multicolumn{3}{c}{\textbf{Gemma3-27B}} \\
\cmidrule(lr){2-4} \cmidrule(lr){5-7} \cmidrule(lr){8-10} \cmidrule(lr){11-13}
\textbf{Metric} & \textbf{Early} & \textbf{Mid} & \textbf{Final} & \textbf{Early} & \textbf{Mid} & \textbf{Final} & \textbf{Early} & \textbf{Mid} & \textbf{Final} & \textbf{Early} & \textbf{Mid} & \textbf{Final} \\
\midrule
Disc-CLS{\scriptsize($\uparrow$)} & 0.99 & 1.00 & 0.41 & 1.00 & 1.00 & 0.40 & 0.99 & 1.00 & 0.34 & 0.99 & 1.00 & 0.37 \\
Soft-CLS{\scriptsize($\uparrow$)} & 0.98 & 1.00 & 0.78 & 0.99 & 1.00 & 0.92 & 0.95 & 1.00 & 0.92 & 0.97 & 1.00 & 0.89 \\
$\Delta$CLS & \dbad{-0.01} & \dgood{+0.00} & \dgood{+0.37} & \dbad{-0.01} & \dgood{+0.00} & \dgood{+0.52} & \dbad{-0.04} & \dgood{+0.00} & \dgood{+0.58} & \dbad{-0.02} & \dgood{+0.00} & \dgood{+0.52} \\
\midrule
Disc-SS{\scriptsize($\downarrow$)} & 0.25 & 0.07 & 0.73 & 0.16 & 0.01 & 0.74 & 0.14 & 0.07 & 0.75 & -0.03 & -0.15 & 0.77 \\
Soft-SS{\scriptsize($\downarrow$)} & 0.36 & 0.22 & 0.31 & 0.30 & 0.16 & 0.19 & 0.35 & 0.04 & 0.06 & 0.35 & 0.06 & 0.03 \\
$\Delta$SS & \dbad{+0.11} & \dbad{+0.15} & \dgood{-0.42} & \dbad{+0.14} & \dbad{+0.15} & \dgood{-0.55} & \dbad{+0.21} & \dgood{-0.03} & \dgood{-0.69} & \dbad{+0.38} & \dbad{+0.21} & \dgood{-0.74} \\
\bottomrule
\end{tabular}}
\end{table*}

\begin{table*}[!t]
\centering
\caption{Cross-lingual alignment on MGSM, discrete vs.\ soft thinking (DeepSeek-R1-Distill).}
\label{tab:full-mgsm-dpsk}
\footnotesize
\begin{tabular}{lcccccc}
\toprule
& \multicolumn{3}{c}{\textbf{DeepSeek-R1-Distill-Qwen-7B}} & \multicolumn{3}{c}{\textbf{DeepSeek-R1-Distill-Llama-8B}} \\
\cmidrule(lr){2-4} \cmidrule(lr){5-7}
\textbf{Metric} & \textbf{Early} & \textbf{Mid} & \textbf{Final} & \textbf{Early} & \textbf{Mid} & \textbf{Final} \\
\midrule
Disc-CLS{\scriptsize($\uparrow$)} & 0.74 & 1.00 & 0.24 & 0.80 & 0.88 & 0.20 \\
Soft-CLS{\scriptsize($\uparrow$)} & 0.73 & 0.97 & 0.25 & 0.85 & 0.90 & 0.21 \\
$\Delta$CLS & \dbad{-0.01} & \dbad{-0.03} & \dgood{+0.01} & \dgood{+0.05} & \dgood{+0.02} & \dgood{+0.01} \\
\midrule
Disc-SS{\scriptsize($\downarrow$)} & 0.55 & 0.11 & 0.83 & 0.48 & 0.16 & 0.77 \\
Soft-SS{\scriptsize($\downarrow$)} & 0.59 & 0.20 & 0.82 & 0.50 & 0.26 & 0.80 \\
$\Delta$SS & \dbad{+0.04} & \dbad{+0.09} & \dgood{-0.01} & \dbad{+0.02} & \dbad{+0.10} & \dbad{+0.03} \\
\bottomrule
\end{tabular}
\end{table*}

\begin{table*}[!t]
\centering
\caption{Cross-lingual alignment on AIME~2024, discrete vs.\ soft thinking (DeepSeek-R1-Distill).}
\label{tab:full-aime2024-dpsk}
\footnotesize
\begin{tabular}{lcccccc}
\toprule
& \multicolumn{3}{c}{\textbf{DeepSeek-R1-Distill-Qwen-7B}} & \multicolumn{3}{c}{\textbf{DeepSeek-R1-Distill-Llama-8B}} \\
\cmidrule(lr){2-4} \cmidrule(lr){5-7}
\textbf{Metric} & \textbf{Early} & \textbf{Mid} & \textbf{Final} & \textbf{Early} & \textbf{Mid} & \textbf{Final} \\
\midrule
Disc-CLS{\scriptsize($\uparrow$)} & 0.81 & 1.00 & 0.33 & 0.86 & 0.89 & 0.36 \\
Soft-CLS{\scriptsize($\uparrow$)} & 0.81 & 0.97 & 0.34 & 0.87 & 0.90 & 0.34 \\
$\Delta$CLS & \dgood{+0.00} & \dbad{-0.03} & \dgood{+0.01} & \dgood{+0.01} & \dgood{+0.01} & \dbad{-0.02} \\
\midrule
Disc-SS{\scriptsize($\downarrow$)} & 0.39 & 0.01 & 0.60 & 0.33 & 0.08 & 0.66 \\
Soft-SS{\scriptsize($\downarrow$)} & 0.39 & 0.07 & 0.64 & 0.32 & 0.13 & 0.66 \\
$\Delta$SS & \dgood{-0.00} & \dbad{+0.06} & \dbad{+0.04} & \dgood{-0.01} & \dbad{+0.05} & \dgood{+0.00} \\
\bottomrule
\end{tabular}
\end{table*}

\begin{table*}[!t]
\centering
\caption{Cross-lingual alignment on AIME~2025, discrete vs.\ soft thinking (DeepSeek-R1-Distill).}
\label{tab:full-aime2025-dpsk}
\footnotesize
\begin{tabular}{lcccccc}
\toprule
& \multicolumn{3}{c}{\textbf{DeepSeek-R1-Distill-Qwen-7B}} & \multicolumn{3}{c}{\textbf{DeepSeek-R1-Distill-Llama-8B}} \\
\cmidrule(lr){2-4} \cmidrule(lr){5-7}
\textbf{Metric} & \textbf{Early} & \textbf{Mid} & \textbf{Final} & \textbf{Early} & \textbf{Mid} & \textbf{Final} \\
\midrule
Disc-CLS{\scriptsize($\uparrow$)} & 0.83 & 0.99 & 0.35 & 0.87 & 0.89 & 0.39 \\
Soft-CLS{\scriptsize($\uparrow$)} & 0.82 & 0.97 & 0.35 & 0.88 & 0.89 & 0.37 \\
$\Delta$CLS & \dbad{-0.01} & \dbad{-0.02} & \dgood{+0.00} & \dgood{+0.01} & \dgood{+0.00} & \dbad{-0.02} \\
\midrule
Disc-SS{\scriptsize($\downarrow$)} & 0.38 & -0.09 & 0.61 & 0.29 & 0.04 & 0.66 \\
Soft-SS{\scriptsize($\downarrow$)} & 0.37 & 0.03 & 0.64 & 0.28 & 0.06 & 0.65 \\
$\Delta$SS & \dgood{-0.01} & \dbad{+0.12} & \dbad{+0.03} & \dgood{-0.01} & \dbad{+0.02} & \dgood{-0.01} \\
\bottomrule
\end{tabular}
\end{table*}

\begin{table*}[!t]
\centering
\caption{Cross-lingual alignment on GPQA, discrete vs.\ soft thinking (DeepSeek-R1-Distill).}
\label{tab:full-gpqa-dpsk}
\footnotesize
\begin{tabular}{lcccccc}
\toprule
& \multicolumn{3}{c}{\textbf{DeepSeek-R1-Distill-Qwen-7B}} & \multicolumn{3}{c}{\textbf{DeepSeek-R1-Distill-Llama-8B}} \\
\cmidrule(lr){2-4} \cmidrule(lr){5-7}
\textbf{Metric} & \textbf{Early} & \textbf{Mid} & \textbf{Final} & \textbf{Early} & \textbf{Mid} & \textbf{Final} \\
\midrule
Disc-CLS{\scriptsize($\uparrow$)} & 0.81 & 0.99 & 0.29 & 0.86 & 0.88 & 0.32 \\
Soft-CLS{\scriptsize($\uparrow$)} & 0.80 & 0.96 & 0.30 & 0.86 & 0.88 & 0.31 \\
$\Delta$CLS & \dbad{-0.01} & \dbad{-0.03} & \dgood{+0.01} & \dgood{+0.00} & \dgood{+0.00} & \dbad{-0.01} \\
\midrule
Disc-SS{\scriptsize($\downarrow$)} & 0.46 & -0.02 & 0.72 & 0.40 & 0.13 & 0.69 \\
Soft-SS{\scriptsize($\downarrow$)} & 0.47 & 0.11 & 0.73 & 0.37 & 0.19 & 0.72 \\
$\Delta$SS & \dbad{+0.01} & \dbad{+0.13} & \dbad{+0.01} & \dgood{-0.03} & \dbad{+0.06} & \dbad{+0.03} \\
\bottomrule
\end{tabular}
\end{table*}

\begin{table*}[!t]
\centering
\caption{Cross-lingual representation analysis across three layer depths of Qwen3-4B/8B and their fine-tuned variants. $\Delta$ is computed against each model's base (discrete-thinking) configuration.}
\label{tab:repr-after-training-combined}
\footnotesize
\begin{tabular}{lllcccc}
\toprule
\textbf{Model} & \textbf{Inference} & \textbf{Layer} & \textbf{CLS{\scriptsize($\uparrow$)}} & \textbf{SS{\scriptsize($\downarrow$)}} & \textbf{$\Delta$CLS} & \textbf{$\Delta$SS} \\
\midrule
\multirow{6}{*}{Qwen3-4B} & \multirow{3}{*}{Discrete Thinking} & Early & 0.78 & 0.58 & 0.00 & 0.00 \\
 & & Mid & 0.94 & 0.23 & 0.00 & 0.00 \\
 & & Final & 0.14 & 0.89 & 0.00 & 0.00 \\
\cmidrule{2-7}
& \multirow{3}{*}{Soft Thinking} & Early & 0.78 & 0.67 & \dgood{+0.00} & \dbad{+0.09} \\
 & & Mid & 0.92 & 0.32 & \dbad{-0.02} & \dbad{+0.09} \\
 & & Final & 0.24 & 0.88 & \dgood{+0.10} & \dgood{-0.01} \\
\midrule
\multirow{6}{*}{~~+ SFT} & \multirow{3}{*}{Discrete Thinking} & Early & 0.80 & 0.57 & \dgood{+0.02} & \dgood{-0.01} \\
 & & Mid & 0.95 & 0.17 & \dgood{+0.01} & \dgood{-0.06} \\
 & & Final & 0.30 & 0.86 & \dgood{+0.16} & \dgood{-0.03} \\
\cmidrule{2-7}
& \multirow{3}{*}{Soft Thinking} & Early & 0.80 & 0.66 & \dgood{+0.02} & \dbad{+0.08} \\
 & & Mid & 0.93 & 0.26 & \dbad{-0.01} & \dbad{+0.03} \\
 & & Final & 0.40 & 0.85 & \dgood{+0.26} & \dgood{-0.04} \\
\midrule
\multirow{6}{*}{~~+ SOLAR} & \multirow{3}{*}{Discrete Thinking} & Early & 0.80 & 0.57 & \dgood{+0.02} & \dgood{-0.01} \\
 & & Mid & 0.95 & 0.21 & \dgood{+0.01} & \dgood{-0.02} \\
 & & Final & 0.50 & 0.77 & \goodc{+0.36}+0.36 & \goodc{-0.12}-0.12 \\
\cmidrule{2-7}
& \multirow{3}{*}{Soft Thinking} & Early & 0.78 & 0.65 & \dgood{+0.00} & \dbad{+0.07} \\
 & & Mid & 0.90 & 0.42 & \dbad{-0.04} & \dbad{+0.19} \\
 & & Final & 0.42 & 0.81 & \goodc{+0.28}+0.28 & \goodc{-0.08}-0.08 \\
\midrule
\multirow{6}{*}{Qwen3-8B} & \multirow{3}{*}{Discrete Thinking} & Early & 0.82 & 0.53 & 0.00 & 0.00 \\
 && Mid & 0.97 & 0.15 & 0.00 & 0.00 \\
 && Final & 0.23 & 0.87 & 0.00 & 0.00 \\
\cmidrule{2-7}
& \multirow{3}{*}{Soft Thinking} & Early & 0.83 & 0.60 & \dgood{+0.01} & \dbad{+0.07} \\
 & & Mid & 0.94 & 0.27 & \dbad{-0.03} & \dbad{+0.12} \\
 && Final & 0.36 & 0.83 & \dgood{+0.13} & \dgood{-0.04} \\
\midrule
\multirow{6}{*}{~~+ SFT} & \multirow{3}{*}{Discrete Thinking} & Early & 0.82 & 0.53 & \dgood{+0.00} & \dgood{+0.00} \\
 && Mid & 0.97 & 0.19 & \dgood{+0.00} & \dbad{+0.04} \\
 && Final & 0.50 & 0.86 & \dgood{+0.27} & \dgood{-0.01} \\
\cmidrule{2-7}
& \multirow{3}{*}{Soft Thinking} & Early & 0.81 & 0.59 & \dbad{-0.01} & \dbad{+0.06} \\
 && Mid & 0.94 & 0.22 & \dbad{-0.03} & \dbad{+0.07} \\
 && Final & 0.56 & 0.86 & \dgood{+0.33} & \dgood{-0.01} \\
\midrule
\multirow{6}{*}{~~+ SOLAR} & \multirow{3}{*}{Discrete Thinking} & Early & 0.85 & 0.52 & \dgood{+0.03} & \dgood{-0.01} \\
 && Mid & 0.97 & 0.09 & \dgood{+0.00} & \dgood{-0.06} \\
 && Final & 0.76 & 0.77 & \goodc{+0.53}+0.53 & \goodc{-0.10}-0.10 \\
\cmidrule{2-7}
& \multirow{3}{*}{Soft Thinking} & Early & 0.85 & 0.62 & \dgood{+0.03} & \dbad{+0.09} \\
 && Mid & 0.93 & 0.19 & \dbad{-0.04} & \dbad{+0.04} \\
 && Final & 0.87 & 0.76 & \goodc{+0.64}+0.64 & \goodc{-0.11}-0.11 \\
\bottomrule
\end{tabular}
\end{table*}

\subsection{Behavioral analysis}\label{appendix:behavioral_analysis}
We examine whether fine-tuning changes the model's reasoning language when the prompt is non-English. We evaluate Qwen3-4B on MGSM across four non-Latin-script languages spanning high-, mid-, and low-resource settings (Zh, Ja, Th, Te) and analyze both the \emph{token} level (generated tokens) and the \emph{distribution} level (top-30 tokens in the output distribution during decoding). 
To quantify language use in reasoning traces and top-$k$ distributions, we classify every token into one of three categories by Unicode code-point range: (1)~\textbf{target-script}: characters belonging to the prompt language's native writing system (CJK Unified Ideographs \texttt{U+4E00--9FFF} and extensions for Chinese; CJK plus Hiragana \texttt{U+3040--309F} and Katakana \texttt{U+30A0--30FF} for Japanese; Thai \texttt{U+0E00--0E7F}; Telugu \texttt{U+0C00--0C7F}); (2)~\textbf{Latin}: characters whose Unicode name contains \textsc{latin}, a proxy for English; and (3)~\textbf{neutral}: digits, punctuation, mathematical symbols, and whitespace (Unicode categories \texttt{N}, \texttt{P}, \texttt{S}, and \texttt{Z}), which are script-independent and shared across languages. We average over five MGSM runs. The \textit{target-script ratio} is defined as $\text{target} / (\text{target} + \text{Latin})$, which excludes neutral tokens from the comparison.

\paragraph{Token-level composition.}\label{appendix:reasoning_language}
Tables~\ref{tab:reasoning-language-mgsm}--\ref{tab:reasoning-language-gpqa} report the target-script ratio of generated reasoning traces, overall and by answer correctness. On MGSM, all 4B variants reason almost entirely in the target script (96--99\%, average correct/incorrect $\Delta < 0.5\%$). The 8B base splits sharply: Chinese stays high (99.15\%), but Japanese, Thai, and Telugu fall below 5\%, with the model reasoning in English. SFT and SOLAR both restore target-script ratios above 97\% in all four languages. On harder benchmarks (AIME~2024, AIME~2025, GPQA), target-script ratios drop for all models. The 4B base declines the most (75--80\% average). Fine-tuned models stay more robust (roughly 77--99\%, Telugu highest). The 8B base retains its English-thinking pattern (Ja/Th/Te below 1.1\%), while SFT and SOLAR again restore target-script ratios to 86--89\%.

\paragraph{Top-$k$ composition.}\label{appendix:topk-composition}
We perform teacher-forced decoding and classify the top-30 tokens at each position as \emph{target-script}, \emph{Latin}, or \emph{neutral}. Figure~\ref{fig:combined-composition-main} (top row) shows the composition breakdown and Table~\ref{tab:target-script-ratio} the target-script ratio, revealing two findings. First, the 8B base has a stronger English preference than the 4B base. SFT and SOLAR raise its MGSM average from 38.05\% to above 87\%. Second, the two methods shift this preference differently: SFT increases target-language token density, while SOLAR suppresses Latin competition without proportionally adding target tokens. The contrast is clearest for Telugu, where SOLAR reaches 98.37\% (4B) while SFT barely moves the base ratio (65.21\% vs.\ 65.37\%), consistent with SOLAR's alignment objective reducing cross-lingual hedging. These patterns extend to the harder benchmarks. The 4B base now shows the distribution-level collapse previously seen only in the 8B base on MGSM. Fine-tuning again recovers most of the loss, and SOLAR keeps the highest Telugu ratio on every benchmark.

\paragraph{Top-$p$ composition.}\label{appendix:topp-composition}
Because the fixed budget ($k{=}30$) may include low-probability tokens the model would never sample, we repeat the composition analysis under nucleus (top-$p$) sampling with $p{=}0.9$, retaining the minimal token set whose cumulative probability exceeds the threshold. Table~\ref{tab:target-script-ratio} reports the top-$p$ target-script ratio, and Figure~\ref{fig:combined-composition-main} (bottom row) the full breakdown. The 8B base's English dominance persists under top-$p$ (below 33\% for Japanese, Thai, and Telugu), confirming that the top-$k$ preference is not an artifact of low-probability tail tokens. SFT and SOLAR again raise its average from 41.37\% to above 97\%. After fine-tuning, the primary generation intent is strongly target-language, and the residual cross-lingual competition sits in the low-probability tail. The same trends hold on AIME~2024, AIME~2025, and GPQA: base models retain their collapse while trained models reach high top-$p$ ratios. The top-$p$/top-$k$ gap narrows on the harder benchmarks, indicating that on hard problems English alternatives sit higher in the probability mass. SOLAR attains the highest Telugu ratios throughout (99.34\% / 99.47\% top-$p$ for 4B / 8B).

\begin{table*}[!t]
\centering
\caption{Target-script ratio (\%) on MGSM reasoning traces.}
\label{tab:reasoning-language-mgsm}
\footnotesize
\begin{tabular}{llccccc}
\toprule
\textbf{Model} & & \textbf{Zh} & \textbf{Ja} & \textbf{Th} & \textbf{Te} & \textbf{Avg.} \\
\midrule
\multirow{3}{*}{Qwen3-4B}
& Overall& \sbest{98.56} & \sbest{99.24} & \sbest{99.41} & \sbest{98.01} & \sbest{98.81} \\
& Correct& \best{98.64} & 99.21 & 99.39 & 97.66 & 98.72 \\
& Incorrect & 98.01 & \best{99.38} & \best{99.51} & \best{98.69} & \best{98.90} \\
\midrule
\multirow{3}{*}{~~+ SFT}
& Overall& \sbest{97.77} & \sbest{97.71} & \sbest{98.54} & \sbest{99.53} & \sbest{98.39} \\
& Correct& \best{97.79} & 97.49 & 98.45 & 99.48 & 98.30 \\
& Incorrect & 97.52 & \best{98.45} & \best{98.89} & \best{99.62} & \best{98.62} \\
\midrule
\multirow{3}{*}{~~+ SOLAR}
& Overall& \sbest{97.29} & \sbest{96.98} & \sbest{98.46} & \sbest{99.77} & \sbest{98.13} \\
& Correct& \best{97.68} & \best{97.14} & 98.29 & 99.76 & \best{98.22} \\
& Incorrect & 95.76 & 96.77 & \best{98.80} & \best{99.78} & 97.77 \\
\midrule
\multirow{3}{*}{Qwen3-8B}
& Overall& \sbest{99.15} & \sbest{2.26} & \sbest{4.17} & \sbest{4.14} & \sbest{27.43} \\
& Correct& \best{99.23} & 2.10 & 3.97 & 3.81 & 27.28 \\
& Incorrect & 98.52 & \best{3.13} & \best{5.80} & \best{5.47} & \best{28.23} \\
\midrule
\multirow{3}{*}{~~+ SFT}
& Overall& \sbest{97.76} & \sbest{97.64} & \sbest{98.50} & \sbest{99.52} & \sbest{98.36} \\
& Correct& \best{98.04} & \best{97.66} & \best{98.52} & \best{99.59} & \best{98.45} \\
& Incorrect & 95.75 & 97.60 & 98.43 & 99.33 & 97.77 \\
\midrule
\multirow{3}{*}{~~+ SOLAR}
& Overall& \sbest{97.63} & \sbest{97.48} & \sbest{97.75} & \sbest{99.80} & \sbest{98.16} \\
& Correct& \best{97.97} & \best{97.68} & \best{97.97} & \best{99.81} & \best{98.36} \\
& Incorrect & 97.09 & 97.04 & 97.29 & 99.79 & 97.80 \\
\bottomrule
\end{tabular}
\end{table*}

\begin{table*}[!t]
\centering
\caption{Target-script ratio (\%) on AIME~2024 reasoning traces.}
\label{tab:reasoning-language-aime-2024}
\footnotesize
\begin{tabular}{llccccc}
\toprule
\textbf{Model} & & \textbf{Zh} & \textbf{Ja} & \textbf{Th} & \textbf{Te} & \textbf{Avg.} \\
\midrule
\multirow{3}{*}{Qwen3-4B}
& Overall& \sbest{60.96} & \sbest{80.58} & \sbest{90.35} & \sbest{86.87} & \best{79.69} \\
& Correct & \best{66.59} & 78.93 & 87.92 & 69.03 & 75.62 \\
& Incorrect & 56.71 & \best{81.48} & \best{91.11} & \best{88.77} & \sbest{79.52} \\
\midrule
\multirow{3}{*}{~~+ SFT}
& Overall& \sbest{80.98} & \sbest{88.32} & \sbest{92.45} & \sbest{95.27} & \sbest{89.26} \\
& Correct & 77.54 & 85.15 & 92.14 & 91.53 & 86.59 \\
& Incorrect & \best{83.39} & \best{90.28} & \best{92.58} & \best{96.27} & \best{90.63} \\
\midrule
\multirow{3}{*}{~~+ SOLAR}
& Overall& \sbest{79.96} & \sbest{89.41} & \sbest{94.20} & \sbest{97.51} & \sbest{90.27} \\
& Correct & 73.01 & 78.69 & 91.56 & 92.12 & 83.84 \\
& Incorrect & \best{82.39} & \best{90.77} & \best{94.57} & \best{97.77} & \best{91.37} \\
\midrule
\multirow{3}{*}{Qwen3-8B}
& Overall& \sbest{80.41} & \sbest{0.41} & \sbest{1.01} & \sbest{0.98} & \sbest{20.70} \\
& Correct & 80.04 & 0.40 & \best{1.04} & \best{1.12} & 20.65 \\
& Incorrect & \best{81.33} & \best{0.45} & 0.91 & 0.88 & \best{20.89} \\
\midrule
\multirow{3}{*}{~~+ SFT}
& Overall& \sbest{79.62} & \sbest{86.11} & \sbest{91.92} & \sbest{94.68} & \sbest{88.08} \\
& Correct & 79.35 & 84.34 & 91.47 & 91.90 & 86.77 \\
& Incorrect & \best{80.04} & \best{87.99} & \best{92.26} & \best{95.44} & \best{88.94} \\
\midrule
\multirow{3}{*}{~~+ SOLAR}
& Overall& \sbest{77.40} & \sbest{88.93} & \sbest{88.43} & \sbest{98.09} & \sbest{88.21} \\
& Correct & 73.96 & 86.14 & \best{89.09} & 94.81 & 86.00 \\
& Incorrect & \best{77.66} & \best{89.32} & 88.26 & \best{98.22} & \best{88.37} \\
\bottomrule
\end{tabular}
\end{table*}

\begin{table*}[!t]
\centering
\caption{Target-script ratio (\%) on AIME~2025 reasoning traces.}
\label{tab:reasoning-language-aime-2025}
\footnotesize
\begin{tabular}{llccccc}
\toprule
\textbf{Model} & & \textbf{Zh} & \textbf{Ja} & \textbf{Th} & \textbf{Te} & \textbf{Avg.} \\
\midrule
\multirow{3}{*}{Qwen3-4B}
& Overall& \sbest{57.95} & \sbest{81.24} & \sbest{91.91} & \sbest{89.95} & \sbest{80.26} \\
& Correct & \best{63.57} & 80.17 & 88.11 & 11.74 & 60.90 \\
& Incorrect & 55.55 & \best{81.83} & \best{93.07} & \best{92.41} & \best{80.72} \\
\midrule
\multirow{3}{*}{~~+ SFT}
& Overall& \sbest{83.38} & \sbest{88.62} & \sbest{93.98} & \sbest{95.32} & \sbest{90.32} \\
& Correct & 80.17 & 86.27 & 92.27 & 92.73 & 87.86 \\
& Incorrect & \best{85.13} & \best{89.57} & \best{94.84} & \best{95.84} & \best{91.34} \\
\midrule
\multirow{3}{*}{~~+ SOLAR}
& Overall& \sbest{81.14} & \sbest{89.24} & \sbest{95.26} & \sbest{97.20} & \sbest{90.71} \\
& Correct & 79.58 & 83.67 & 93.44 & \best{97.60} & 88.57 \\
& Incorrect & \best{81.97} & \best{89.95} & \best{95.66} & \sbest{97.20} & \best{91.19} \\
\midrule
\multirow{3}{*}{Qwen3-8B}
& Overall& \sbest{81.69} & \sbest{0.20} & \sbest{0.54} & \sbest{0.42} & \sbest{20.71} \\
& Correct & 80.78 & 0.10 & 0.29 & 0.40 & 20.39 \\
& Incorrect & \best{83.03} & \best{0.36} & \best{0.87} & \best{0.43} & \best{21.17} \\
\midrule
\multirow{3}{*}{~~+ SFT}
& Overall& \sbest{80.88} & \sbest{86.95} & \sbest{92.79} & \sbest{94.42} & \sbest{88.76} \\
& Correct & 79.69 & 86.00 & 91.74 & 92.41 & 87.46 \\
& Incorrect & \best{81.84} & \best{87.43} & \best{93.44} & \best{94.95} & \best{89.41} \\
\midrule
\multirow{3}{*}{~~+ SOLAR}
& Overall& \sbest{80.23} & \sbest{88.27} & \sbest{88.27} & \sbest{98.04} & \sbest{88.70} \\
& Correct & 78.80 & 88.25 & \best{90.30} & 95.23 & 88.15 \\
& Incorrect & \best{80.68} & \best{88.33} & 87.96 & \best{98.09} & \best{88.77} \\
\bottomrule
\end{tabular}
\end{table*}

\begin{table*}[!t]
\centering
\caption{Target-script ratio (\%) on GPQA reasoning traces.}
\label{tab:reasoning-language-gpqa}
\footnotesize
\begin{tabular}{llccccc}
\toprule
\textbf{Model} & & \textbf{Zh} & \textbf{Ja} & \textbf{Th} & \textbf{Te} & \textbf{Avg.} \\
\midrule
\multirow{3}{*}{Qwen3-4B}
& Overall& \sbest{70.31} & \sbest{78.30} & \sbest{82.48} & \sbest{70.08} & \sbest{75.29} \\
& Correct & 69.89 & \best{80.13} & 80.98 & 39.49 & 67.62 \\
& Incorrect & \best{70.57} & 77.97 & \best{83.12} & \best{75.22} & \best{76.72} \\
\midrule
\multirow{3}{*}{~~+ SFT}
& Overall& \sbest{83.81} & \sbest{90.04} & \sbest{89.75} & \sbest{96.02} & \sbest{89.90} \\
& Correct & 82.26 & 87.45 & 89.12 & 94.59 & 88.36 \\
& Incorrect & \best{85.29} & \best{91.56} & \best{90.22} & \best{96.64} & \best{90.93} \\
\midrule
\multirow{3}{*}{~~+ SOLAR}
& Overall& \sbest{83.16} & \sbest{89.07} & \sbest{90.75} & \sbest{98.17} & \sbest{90.28} \\
& Correct & 80.73 & 87.74 & 90.31 & 96.86 & 88.91 \\
& Incorrect & \best{84.26} & \best{89.33} & \best{90.86} & \best{98.34} & \best{90.70} \\
\midrule
\multirow{3}{*}{Qwen3-8B}
& Overall& \sbest{85.20} & \sbest{0.31} & \sbest{0.67} & \sbest{0.37} & \sbest{21.64} \\
& Correct & 83.95 & 0.26 & \best{0.72} & 0.34 & 21.32 \\
& Incorrect & \best{86.55} & \best{0.37} & 0.62 & \best{0.40} & \best{21.98} \\
\midrule
\multirow{3}{*}{~~+ SFT}
& Overall& \sbest{83.84} & \sbest{89.30} & \sbest{89.63} & \sbest{95.55} & \sbest{89.58} \\
& Correct & 82.18 & 87.36 & 89.47 & 94.85 & 88.47 \\
& Incorrect & \best{85.52} & \best{91.07} & \best{89.74} & \best{95.94} & \best{90.57} \\
\midrule
\multirow{3}{*}{~~+ SOLAR}
& Overall& \sbest{77.51} & \sbest{88.02} & \sbest{82.96} & \sbest{98.84} & \sbest{86.83} \\
& Correct & 77.30 & 86.18 & \best{87.57} & 98.64 & \best{87.42} \\
& Incorrect & \best{77.59} & \best{88.52} & 81.48 & \best{98.88} & 86.62 \\
\bottomrule
\end{tabular}
\end{table*}

\begin{table*}[!t]
\centering
\caption{Target-script ratio (\%), top-$k$ ($k{=}30$) and top-$p$ ($p{=}0.9$) token selection on MGSM, AIME~2024, AIME~2025, and GPQA.}
\label{tab:target-script-ratio}
\footnotesize
\resizebox{\textwidth}{!}{%
\begin{tabular}{llcccccccccc}
\toprule
 & & \multicolumn{5}{c}{\textbf{Top-$k$ ($k{=}30$)}} & \multicolumn{5}{c}{\textbf{Top-$p$ ($p{=}0.9$)}} \\
\cmidrule(lr){3-7} \cmidrule(lr){8-12}
\textbf{Benchmark} & \textbf{Model} & \textbf{Zh} & \textbf{Ja} & \textbf{Th} & \textbf{Te} & \textbf{Avg.} & \textbf{Zh} & \textbf{Ja} & \textbf{Th} & \textbf{Te} & \textbf{Avg.} \\
\midrule
\multirow{6}{*}{MGSM} & Qwen3-4B & \sbest{82.26} & 78.87 & 87.70 & \sbest{65.37} & 78.55 & 94.57 & 95.84 & 97.22 & 94.37 & 95.50 \\
 & ~~+ SFT & \best{91.29} & \best{92.26} & \best{91.91} & 65.21 & \sbest{85.17} & \sbest{95.69} & \best{97.35} & \sbest{98.41} & \sbest{98.75} & \sbest{97.55} \\
 & ~~+ SOLAR & 79.34 & \sbest{81.14} & \sbest{91.21} & \best{98.37} & \best{87.51} & \best{95.75} & \sbest{97.16} & \best{98.46} & \best{99.34} & \best{97.68} \\
\cmidrule(l){2-12}
 & Qwen3-8B & \sbest{92.49} & 19.73 & 27.99 & 11.97 & 38.05 & \best{97.99} & 21.75 & 32.59 & 13.15 & 41.37 \\
 & ~~+ SFT & \best{93.09} & \best{93.09} & \best{94.41} & \sbest{69.05} & \sbest{87.41} & \sbest{96.21} & \best{97.14} & \best{98.47} & \sbest{98.78} & \best{97.65} \\
 & ~~+ SOLAR & 82.83 & \sbest{77.79} & \sbest{91.75} & \best{98.99} & \best{87.84} & 95.68 & \sbest{96.05} & \sbest{97.38} & \best{99.47} & \sbest{97.14} \\
\midrule
\multirow{6}{*}{AIME~2024} & Qwen3-4B & \best{83.58} & 3.80 & 1.30 & 1.72 & 22.60 & \best{83.30} & 2.36 & 1.66 & 1.58 & 22.22 \\
 & ~~+ SFT & \sbest{76.78} & \best{81.25} & \sbest{86.45} & \sbest{55.74} & \sbest{75.05} & \sbest{78.64} & \sbest{84.62} & \best{93.08} & \sbest{88.01} & \sbest{86.09} \\
 & ~~+ SOLAR & 67.13 & \sbest{74.73} & \best{88.34} & \best{95.26} & \best{81.36} & 78.10 & \best{87.63} & \sbest{91.40} & \best{90.61} & \best{86.94} \\
\cmidrule(l){2-12}
 & Qwen3-8B & \best{84.17} & 3.03 & 2.30 & 3.01 & 23.13 & \best{82.75} & 2.38 & 2.91 & 3.09 & 22.78 \\
 & ~~+ SFT & \sbest{81.02} & \best{84.87} & \best{89.11} & \sbest{60.62} & \sbest{78.91} & \sbest{80.27} & \best{86.93} & \best{92.03} & \sbest{89.26} & \best{87.12} \\
 & ~~+ SOLAR & 67.48 & \sbest{67.91} & \sbest{88.78} & \best{98.02} & \best{80.55} & 78.81 & \sbest{84.49} & \sbest{91.42} & \best{92.60} & \sbest{86.83} \\
\midrule
\multirow{6}{*}{AIME~2025} & Qwen3-4B & \best{84.55} & 3.67 & 1.16 & 1.13 & 22.63 & \best{84.59} & 1.68 & 1.53 & 0.75 & 22.14 \\
 & ~~+ SFT & \sbest{76.86} & \best{82.63} & \sbest{87.00} & \sbest{54.14} & \sbest{75.16} & 76.79 & \best{86.16} & \best{93.13} & \sbest{87.75} & \best{85.96} \\
 & ~~+ SOLAR & 66.07 & \sbest{72.84} & \best{87.84} & \best{94.06} & \best{80.20} & \sbest{76.90} & \sbest{85.00} & \sbest{91.61} & \best{88.24} & \sbest{85.44} \\
\cmidrule(l){2-12}
 & Qwen3-8B & \best{84.17} & 3.43 & 1.36 & 0.79 & 22.44 & \best{83.52} & 1.89 & 1.66 & 0.99 & 22.02 \\
 & ~~+ SFT & \sbest{82.36} & \best{85.49} & \best{90.14} & \sbest{59.03} & \sbest{79.26} & \sbest{81.66} & \best{87.49} & \best{92.97} & \sbest{87.73} & \best{87.46} \\
 & ~~+ SOLAR & 65.76 & \sbest{69.27} & \sbest{85.82} & \best{97.34} & \best{79.55} & 77.28 & \sbest{84.62} & \sbest{89.67} & \best{92.03} & \sbest{85.90} \\
\midrule
\multirow{6}{*}{GPQA} & Qwen3-4B & \best{86.43} & 5.80 & 5.98 & 1.18 & 24.85 & \best{87.34} & 5.10 & 6.42 & 0.90 & 24.94 \\
 & ~~+ SFT & \sbest{76.24} & \best{82.93} & \sbest{86.17} & \sbest{59.55} & \sbest{76.22} & 75.71 & \sbest{85.17} & \sbest{90.92} & \sbest{91.58} & \sbest{85.84} \\
 & ~~+ SOLAR & 69.63 & \sbest{79.06} & \best{89.21} & \best{97.95} & \best{83.96} & \sbest{82.95} & \best{90.49} & \best{92.29} & \best{95.38} & \best{90.28} \\
\cmidrule(l){2-12}
 & Qwen3-8B & \sbest{86.88} & 5.55 & 3.78 & 2.20 & 24.61 & \best{86.81} & 5.91 & 4.25 & 1.95 & 24.73 \\
 & ~~+ SFT & \best{86.94} & \best{87.30} & \best{89.68} & \sbest{63.69} & \best{81.90} & \sbest{86.77} & \best{88.84} & \best{92.64} & \sbest{92.82} & \best{90.27} \\
 & ~~+ SOLAR & 63.54 & \sbest{67.34} & \sbest{83.03} & \best{99.02} & \sbest{78.23} & 80.20 & \sbest{80.45} & \sbest{81.31} & \best{97.07} & \sbest{84.76} \\
\bottomrule
\end{tabular}%
}
\end{table*}

\clearpage
\subsection{Case study}\label{appendix:case-study}
We present two case studies where the SOLAR-trained model answers correctly while the base Qwen3-4B and the SFT-only model both fail. The examples show two qualitative advantages of SOLAR. First, it produces shorter, more structured reasoning chains, avoiding the repeated dead ends of the base model. Second, it reduces interpretation errors on multilingual mathematical expressions, translating symbolic constraints in non-English problem statements into geometric reasoning.

\paragraph{AIME~2025 Q28 (Chinese).} Figure~\ref{fig:case-study-q28} shows the full reasoning traces.
This geometry problem asks for the area of a quadrilateral inscribed in a right triangle. The base model correctly sets up the coordinate system and derives the main constraints but fails to solve the resulting nonlinear system, abandoning the derivation and guessing ($n=147$). The SFT model introduces an unsupported symmetry assumption, incorrectly setting $b=c$, which yields the wrong answer $98\sqrt{3}$. SOLAR correctly enforces the constraints, obtains $b=16\sqrt{3}$ and $c=26$, computes the vertex coordinates, and applies the shoelace formula to reach the correct answer $n=104$.

\paragraph{AIME~2025 Q7 (Japanese).}
Figure~\ref{fig:case-study-q7} shows the full reasoning traces.
This problem requires finding real values of $k$ for which a circle and a perpendicular bisector intersect exactly once. A key challenge is correctly interpreting $|z-4-k|$ and $|z-3i-k|$, which correspond to distances to $(4+k,0)$ and $(k,3)$. The base model misreads them as $(4,k)$ and $(0,k)$, placing both points on $y=k$ and producing a fixed bisector $x=2$, which leads it to conclude that the line cannot intersect the circle after repeatedly revisiting the same steps. The SFT model makes a different parsing error, implicitly treating $k$ as $ki$, which yields incorrect reference points and the wrong result $m+n=8$. SOLAR derives the correct bisector $-8x+6y+8k+7=0$ and solves $|8k-73|=50$, producing the correct result $m+n=77$.

\clearpage
\begin{CJK}{UTF8}{gbsn}
\begin{tcolorbox}[
  colback=gray!5, colframe=black!75,
  title={\footnotesize\textbf{Case Study: AIME~2025 Q28 (Chinese)}},
  fonttitle=\footnotesize\bfseries,
  breakable, enhanced jigsaw,
  left=4pt, right=4pt, top=4pt, bottom=4pt,
  boxrule=0.5pt
]
\footnotesize
\textbf{Question:} 设 $\triangle ABC$ 为一个直角三角形，$\angle A = 90^\circ$ 且 $BC = 38$。在三角形内部存在点 $K$ 和 $L$，使得 $AK = AL = BK = CL = KL = 14$。四边形 $BKLC$ 的面积可以表示为 $n\sqrt{3}$，其中 $n$ 是某个正整数。求 $n$。(Let $\triangle ABC$ be a right triangle with $\angle A = 90^\circ$ and $BC = 38$. There exist points $K$ and $L$ inside the triangle such that $AK = AL = BK = CL = KL = 14.$ The area of the quadrilateral $BKLC$ can be expressed as $n\sqrt{3}$ for some positive integer $n$. Find $n$.)

\textbf{Ground Truth:} 104

\noindent\rule{\linewidth}{0.4pt}

\textbf{Qwen3-4B} \hfill \textcolor{red}{\texttimes~Incorrect (Answer: 147)}

我们已知：$\triangle ABC$ 是直角三角形，$\angle A = 90^\circ$，斜边 $BC = 38$。点 $K$ 和 $L$ 在三角形内部，满足 $AK = AL = BK = CL = KL = 14$，四边形 $BKLC$ 的面积为 $n\sqrt{3}$，求 $n$。

\textbf{第一步：理解点 $K$ 和 $L$ 的位置。}
$AK = AL = 14$：点 $K$ 和 $L$ 到点 $A$ 的距离相等。$BK = 14$，$CL = 14$。$KL = 14$。所以，$K$ 在以 $A$ 为圆心、14 为半径的圆上，也在以 $B$ 为圆心、14 为半径的圆上。同理，$L$ 在以 $A$ 为圆心、14 为半径的圆上，也在以 $C$ 为圆心、14 为半径的圆上。$K$ 是圆 $(A,14)$ 和圆 $(B,14)$ 的交点。$L$ 是圆 $(A,14)$ 和圆 $(C,14)$ 的交点。又因为 $KL = 14$，且 $K, L$ 都在以 $A$ 为圆心的圆上，所以 $K$ 和 $L$ 都在以 $A$ 为中心、半径 14 的圆上，且它们之间的距离是 14。

\textbf{第二步：建立坐标系。}
设直角在 $A$，令 $A = (0,0)$，$B = (b, 0)$，$C = (0, c)$。则斜边 $BC = 38$，由距离公式：$BC^2 = b^2 + c^2 = 38^2 = 1444$。

\textbf{第三步：求点 $K$。}
点 $K$ 满足 $AK = 14$（$x^2 + y^2 = 196$）和 $BK = 14$（$(x - b)^2 + y^2 = 196$）。两式相减：$-2bx + b^2 = 0 \Rightarrow x = \frac{b}{2}$。代入得 $y^2 = 196 - \frac{b^2}{4}$。同理，对点 $L$：$AL = 14$ 和 $CL = 14$ 相减得 $y = \frac{c}{2}$，$x^2 = 196 - \frac{c^2}{4}$。

\textbf{第四步：利用 $KL = 14$。}
$K = \left(\frac{b}{2}, y_K\right)$，$L = \left(x_L, \frac{c}{2}\right)$。$KL^2 = \left(\frac{b}{2} - x_L\right)^2 + \left(y_K - \frac{c}{2}\right)^2 = 196$。

\textbf{第五步：考虑对称性。}
猜测 $b = c$，则 $2b^2 = 1444$，$b^2 = 722$。验证：$y_K^2 = 196 - 180.5 = 15.5$，$x_L^2 = 15.5$。$KL^2 = 2(\sqrt{180.5} - \sqrt{15.5})^2 \approx 180.46 \neq 196$。\textbf{对称情况 $b = c$ 不满足 $KL = 14$。}

\textbf{第六步：尝试几何解释。}
$K$ 和 $L$ 都在以 $A$ 为中心、半径 14 的圆上，$KL = 14$，由余弦定理：$14^2 = 14^2 + 14^2 - 2 \cdot 14 \cdot 14 \cdot \cos \theta$，解得 $\cos \theta = \frac{1}{2}$，$\theta = 60^\circ$。所以 $\angle KAL = 60^\circ$。

\textbf{第七步：利用 $\angle KAL = 60^\circ$。}
$AK = AL = 14$，$\angle KAL = 60^\circ$，所以 $\triangle AKL$ 是等边三角形。

\textbf{第八步。}
向量 $\vec{AK} \cdot \vec{AL} = 14 \cdot 14 \cdot \cos 60^\circ = 98$，即 $\frac{b}{2} x_L + \frac{c}{2} y_K = 98$，$b x_L + c y_K = 196$。代入 $x_L = \sqrt{196 - \frac{c^2}{4}}$，$y_K = \sqrt{196 - \frac{b^2}{4}}$：$b \sqrt{196 - \frac{c^2}{4}} + c \sqrt{196 - \frac{b^2}{4}} = 196$。

\textbf{第九步至第十步。}
设 $b = 2x$，$c = 2y$，得 $x \sqrt{196 - y^2} + y \sqrt{196 - x^2} = 98$。尝试 $x = y$ 得 $u^2 - 196u + 2401 = 0$，判别式 $28812$ 不是完全平方，所以 $b \ne c$。

\textbf{第十一步：放弃坐标，用面积公式。}
$\triangle AKL$ 面积 $= \frac{1}{2} \cdot 14 \cdot 14 \cdot \sin 60^\circ = 49\sqrt{3}$。Given the complexity and that the answer is to be in the form $n\sqrt{3}$, and from the only trigonometric part, likely n is 147 or 49 or 196. From the triangle AKL, area is $49\sqrt{3}$. Perhaps the area of BKLC is $147\sqrt{3}$. I think the intended answer is $\boxed{147}$.

\textit{\textbf{English translation.}}

We know: $\triangle ABC$ is a right triangle with $\angle A = 90^\circ$ and hypotenuse $BC = 38$. Points $K$ and $L$ lie inside the triangle with $AK = AL = BK = CL = KL = 14$; the area of quadrilateral $BKLC$ is $n\sqrt{3}$. Find $n$.

Step 1: Understanding the positions of $K$ and $L$.
$AK = AL = 14$: $K$ and $L$ are equidistant from $A$. $BK = 14$, $CL = 14$. $KL = 14$. Thus, $K$ lies on the circle centered at $A$ with radius 14 and also on the circle centered at $B$ with radius 14. Similarly, $L$ lies on the circle centered at $A$ with radius 14 and also on the circle centered at $C$ with radius 14. $K$ is an intersection of circles $(A,14)$ and $(B,14)$. $L$ is an intersection of circles $(A,14)$ and $(C,14)$. Also, since $KL = 14$ and $K, L$ both lie on the circle centered at $A$ with radius 14, $K$ and $L$ both lie on the circle centered at $A$ with radius 14, and the distance between them is 14.

Step 2: Set up coordinates.
Let the right angle be at $A$, and set $A = (0,0)$, $B = (b, 0)$, and $C = (0, c)$. Then the hypotenuse is $BC = 38$. By the distance formula, $BC^2 = b^2 + c^2 = 38^2 = 1444$.

Step 3: Find $K$.
Point $K$ satisfies $AK = 14$ ($x^2 + y^2 = 196$) and $BK = 14$ ($(x - b)^2 + y^2 = 196$). Subtracting the two equations gives $-2bx + b^2 = 0 \Rightarrow x = \frac{b}{2}$. Substitution gives $y^2 = 196 - \frac{b^2}{4}$. Similarly, for point $L$, subtracting the equations for $AL = 14$ and $CL = 14$ gives $y = \frac{c}{2}$ and $x^2 = 196 - \frac{c^2}{4}$.

Step 4: Use $KL = 14$.
$K = \left(\frac{b}{2}, y_K\right)$, $L = \left(x_L, \frac{c}{2}\right)$. $KL^2 = \left(\frac{b}{2} - x_L\right)^2 + \left(y_K - \frac{c}{2}\right)^2 = 196$.

Step 5: Try symmetry.
Guess that $b = c$. Then $2b^2 = 1444$, so $b^2 = 722$. Checking gives $y_K^2 = 196 - 180.5 = 15.5$ and $x_L^2 = 15.5$. $KL^2 = 2(\sqrt{180.5} - \sqrt{15.5})^2 \approx 180.46 \neq 196$. The symmetric case $b = c$ does not satisfy $KL = 14$.

Step 6: Try a geometric interpretation.
$K$ and $L$ both lie on the circle centered at $A$ with radius 14, and $KL = 14$. By the law of cosines, $14^2 = 14^2 + 14^2 - 2 \cdot 14 \cdot 14 \cdot \cos \theta$. Solving gives $\cos \theta = \frac{1}{2}$ and $\theta = 60^\circ$. Thus, $\angle KAL = 60^\circ$.

Step 7: Use $\angle KAL = 60^\circ$.
$AK = AL = 14$ and $\angle KAL = 60^\circ$, so $\triangle AKL$ is equilateral.

Step 8.
The dot product is $\vec{AK} \cdot \vec{AL} = 14 \cdot 14 \cdot \cos 60^\circ = 98$, so $\frac{b}{2} x_L + \frac{c}{2} y_K = 98$ and $b x_L + c y_K = 196$. Substituting $x_L = \sqrt{196 - \frac{c^2}{4}}$ and $y_K = \sqrt{196 - \frac{b^2}{4}}$ gives $b \sqrt{196 - \frac{c^2}{4}} + c \sqrt{196 - \frac{b^2}{4}} = 196$.

Steps 9--10.
Substituting $b = 2x$, $c = 2y$ yields $x\sqrt{196 - y^2} + y\sqrt{196 - x^2} = 98$. Trying $x = y$ gives $u^2 - 196u + 2401 = 0$ with non-square discriminant $28812$, so $b \neq c$.

Step 11: Abandon coordinates, use area formulas.
Area of $\triangle AKL = \frac{1}{2}\cdot 14\cdot 14 \cdot \sin 60^\circ = 49\sqrt{3}$. Given the complexity and that the answer is to be in the form $n\sqrt{3}$, and from the only trigonometric part, likely n is 147 or 49 or 196. From the triangle AKL, area is $49\sqrt{3}$. Perhaps the area of BKLC is $147\sqrt{3}$. I think the intended answer is $\boxed{147}$.

\noindent\rule{\linewidth}{0.4pt}

\textbf{SFT} \hfill \textcolor{red}{\texttimes~Incorrect (Answer: 98)}

给定一个直角三角形 $\triangle ABC$，其中 $\angle A = 90^\circ$ 且 $BC = 38$。在三角形内部存在点 $K$ 和 $L$，使得 $AK = AL = BK = CL = KL = 14$。我们需要求出四边形 $BKLC$ 的面积，该面积表示为 $n\sqrt{3}$ 并确定 $n$。

\textbf{1. 坐标设置}：将 $A$ 置于原点 $(0,0)$，$B$ 置于 $(b,0)$，$C$ 置于 $(0,c)$。根据勾股定理，$b^2 + c^2 = 38^2 = 1444$。

\textbf{2. 点 $K$ 和 $L$ 的坐标}：点 $K$ 位于 $A$ 和 $B$ 的垂直平分线上，坐标为 $(b/2, y)$。点 $L$ 位于 $A$ 和 $C$ 的垂直平分线上，坐标为 $(e, c/2)$。

\textbf{3. 方程求解}：根据距离条件，我们得到以下方程：$(b/2)^2 + y^2 = 196$，$e^2 + (c/2)^2 = 196$，$b \sqrt{196 - (c^2)/4} + c \sqrt{196 - (b^2)/4} = 196$。通过代入和简化，我们发现 $b$ 和 $c$ 的关系使得三角形 $ABC$ 成为一个等腰直角三角形，从而 $b = c$。

\textbf{4. 面积计算}：四边形 $BKLC$ 可以通过坐标几何计算面积。使用鞋带公式计算出 $98\sqrt{3}$。因此，$n$ 的值是 $\boxed{98}$。

\textit{\textbf{English translation.}}

Consider a right triangle $\triangle ABC$ with $\angle A = 90^\circ$ and $BC = 38$. Inside the triangle there exist points $K$ and $L$ with $AK = AL = BK = CL = KL = 14$. We need to find the area of $BKLC$, written as $n\sqrt{3}$, and determine $n$.

1. Coordinate setup: Place $A$ at the origin $(0,0)$, $B$ at $(b,0)$, $C$ at $(0,c)$. By the Pythagorean theorem, $b^2 + c^2 = 38^2 = 1444$.

2. Coordinates of $K$ and $L$: $K$ lies on the perpendicular bisector of $AB$ at $(b/2, y)$; $L$ lies on the perpendicular bisector of $AC$ at $(e, c/2)$.

3. Solving the equations: The distance conditions give $(b/2)^2 + y^2 = 196$, $e^2 + (c/2)^2 = 196$, and $b\sqrt{196 - c^2/4} + c\sqrt{196 - b^2/4} = 196$. Through substitution and simplification, we find that the relationship between $b$ and $c$ makes $\triangle ABC$ an isosceles right triangle, so $b = c$.

4. Area computation: The area of quadrilateral $BKLC$ can be calculated using coordinate geometry. Applying the shoelace formula gives $98\sqrt{3}$. Therefore, $n = \boxed{98}$.

\noindent\rule{\linewidth}{0.4pt}

\textbf{SOLAR} \hfill \textcolor{green!50!black}{\checkmark~Correct (Answer: 104)}

给定一个直角三角形 $\triangle ABC$，其中 $\angle A = 90^\circ$，且 $BC = 38$。在三角形内部存在点 $K$ 和 $L$，使得 $AK = AL = BK = CL = KL = 14$。我们需要求出四边形 $BKLC$ 的面积，其形式为 $n\sqrt{3}$，并求出 $n$。

\textbf{1. 坐标设置}：将 $A$ 放在原点 $(0, 0)$，$B$ 放在 $(b, 0)$，$C$ 放在 $(0, c)$。根据勾股定理，$b^2 + c^2 = 38^2 = 1444$。

\textbf{2. 点 K 和 L 的坐标}：点 $K$ 满足 $AK = BK = 14$，其坐标为 $\left( \frac{b}{2}, \sqrt{196 - \frac{b^2}{4}} \right)$。点 $L$ 满足 $AL = CL = 14$，其坐标为 $\left( \sqrt{196 - \frac{c^2}{4}}, \frac{c}{2} \right)$。

\textbf{3. 求解 $b$ 和 $c$}：使用 $KL = 14$ 的条件，并解出方程组：$b\sqrt{196 - \frac{c^2}{4}} + c\sqrt{196 - \frac{b^2}{4}} = 196$，$b^2 + c^2 = 1444$。解得 $b = 16\sqrt{3}$ 且 $c = 26$。

\textbf{4. 坐标计算}：点 $K$ 的坐标：$( 8\sqrt{3}, 2 )$。点 $L$ 的坐标：$( 3\sqrt{3}, 13 )$。点 $B$ 的坐标：$( 16\sqrt{3}, 0 )$。点 $C$ 的坐标：$( 0, 26 )$。

\textbf{5. 使用鞋带公式计算面积}：顶点按顺序排列：$B(16\sqrt{3}, 0)$，$K(8\sqrt{3}, 2)$，$L(3\sqrt{3}, 13)$，$C(0, 26)$。使用鞋带公式计算面积，结果为 $104\sqrt{3}$。因此，$n$ 的值是 $\boxed{104}$。

\textit{\textbf{English translation.}}

Consider a right triangle $\triangle ABC$ with $\angle A = 90^\circ$ and $BC = 38$. Inside the triangle there exist points $K$ and $L$ with $AK = AL = BK = CL = KL = 14$. We need to find the area of $BKLC$, of the form $n\sqrt{3}$, and determine $n$.

1. Coordinate setup: Place $A$ at the origin $(0,0)$, $B$ at $(b,0)$, $C$ at $(0,c)$. By the Pythagorean theorem, $b^2 + c^2 = 38^2 = 1444$.

2. Coordinates of $K$ and $L$: $K$ satisfies $AK = BK = 14$, giving $K = \left(\frac{b}{2}, \sqrt{196 - \frac{b^2}{4}}\right)$. $L$ satisfies $AL = CL = 14$, giving $L = \left(\sqrt{196 - \frac{c^2}{4}}, \frac{c}{2}\right)$.

3. Solving for $b$ and $c$: Using $KL = 14$ and solving the system $b\sqrt{196 - \frac{c^2}{4}} + c\sqrt{196 - \frac{b^2}{4}} = 196$, $b^2 + c^2 = 1444$, we obtain $b = 16\sqrt{3}$ and $c = 26$.

4. Coordinate computation: $K = (8\sqrt{3}, 2)$, $L = (3\sqrt{3}, 13)$, $B = (16\sqrt{3}, 0)$, $C = (0, 26)$.

5. Area via shoelace formula: Ordering the vertices $B(16\sqrt{3}, 0)$, $K(8\sqrt{3}, 2)$, $L(3\sqrt{3}, 13)$, $C(0, 26)$ and applying the shoelace formula yields an area of $104\sqrt{3}$. Therefore $n = \boxed{104}$.

\end{tcolorbox}
\captionof{figure}{Case study on AIME~2025 Q28 (Chinese).}
\label{fig:case-study-q28}
\end{CJK}

\newpage
\begin{CJK}{UTF8}{min}
\begin{tcolorbox}[
  colback=gray!5, colframe=black!75,
  title={\footnotesize\textbf{Case Study: AIME~2025 Q7 (Japanese)}},
  fonttitle=\footnotesize\bfseries,
  breakable, enhanced jigsaw,
  left=4pt, right=4pt, top=4pt, bottom=4pt,
  boxrule=0.5pt
]
\footnotesize

\textbf{Question:} $k$を実数とし、系 $|25+20i-z|=5$ と $|z-4-k|=|z-3i-k|$ が正確に一つの複素数解 $z$ を持つとします。すべての可能な $k$ の値の合計は $\frac{m}{n}$ の形で表され、ここで $m$ と $n$ は互いに素な正の整数です。$m+n$ を求めなさい。ここで $i=\sqrt{-1}$ です。(Let $k$ be a real number such that the system $|25+20i-z|=5$ and $|z-4-k|=|z-3i-k|$ has exactly one complex solution $z$. The sum of all possible values of $k$ can be written as $\frac{m}{n}$, where $m$ and $n$ are relatively prime positive integers. Find $m+n$. Here $i=\sqrt{-1}$.)

\textbf{Ground Truth:} 77

\noindent\rule{\linewidth}{0.4pt}

\textbf{Qwen3-4B} \hfill \textcolor{red}{\texttimes~Incorrect (Answer: 1)}

問題を丁寧にステップごとに解いていきましょう。

\textbf{ステップ1：問題の理解}

与えられた条件は以下の2つ：

1. $ |25 + 20i - z| = 5 $

2. $ |z - 4 - k| = |z - 3i - k| $

この2つの条件が\textbf{正確に一つの複素数解 $ z $}を持つとき、すべての可能な $ k $ の値の\textbf{合計}を求め、その合計が $ \frac{m}{n} $ の形で表され、$ m $ と $ n $ が互いに素なら、$ m+n $ を求めよ。

\textbf{ステップ2：条件1の解釈}

条件1：$ |25 + 20i - z| = 5 $

これは、複素平面上で、$ z $ が点 $ 25 + 20i $ から距離 5 の点にある、つまり\textbf{円}を表します。

つまり、$ z $ は中心 $ (25, 20) $、半径 5 の円上にある。

\textbf{ステップ3：条件2の解釈}

条件2：$ |z - (4 + k)| = |z - (3i + k)| $

左辺は $ z $ と点 $ 4 + k $ の距離、右辺は $ z $ と点 $ 3i + k $ の距離。

この式は、\textbf{$ z $ が点 $ 4 + k $ と点 $ 3i + k $ の等距離にある}、つまり \textbf{$ z $ はその2点の垂直二等分線（またはその複素平面上の対応）上にある}。

この条件は、$ z $ が2つの点の等距離にあるという、\textbf{垂直二等分線}を表します。

\textbf{ステップ4：両方の条件が「正確に一つの解」を持つ}

つまり、\textbf{円と直線（垂直二等分線）がちょうど1点で交わる}という条件。

これは、\textbf{円と直線が接する}という条件です。

したがって、問題は：円 $ |z - (25 + 20i)| = 5 $ と、点 $ A = 4 + k $、点 $ B = 3i + k $ の垂直二等分線が、\textbf{ちょうど1点で交わる}ような実数 $ k $ をすべて求め、その和を求めよ。

\textbf{ステップ5：垂直二等分線の性質}

点 $ A = 4 + k $、点 $ B = 3i + k $ を複素平面上に見ると：

$ A $ の座標は $ (4, k) $、$ B $ の座標は $ (0, k) $

なぜなら：$ 4 + k = 4 + 0i + k \Rightarrow (4, k) $、$ 3i + k = 0 + 3i + k \Rightarrow (0, k) $

つまり、点 $ A $ と点 $ B $ は、\textbf{同じ $ y $ 値 $ k $}を持つ。

したがって、2点 $ (4, k) $ と $ (0, k) $ は、\textbf{水平線 $ y = k $}上にある。

この2点の中点は：
$$\left( \frac{4+0}{2}, \frac{k+k}{2} \right) = (2, k)$$

2点の間のベクトルは $ (-4, 0) $、つまり水平方向。

よって、垂直二等分線は\textbf{垂直方向（$ x $ 方向に垂直）}で、つまり \textbf{$ x = 2 $} の直線。

なぜなら、水平線の垂直二等分線は、その中点を通る垂直線。

したがって、条件2の解は、\textbf{直線 $ x = 2 $}上の点。

\textbf{ステップ6：円と直線の接点条件}

円：$ |z - (25 + 20i)| = 5 $。これは、中心 $ C = (25, 20) $、半径 $ r = 5 $ の円。

直線：$ x = 2 $

この2つが\textbf{ちょうど1点で交わる}（つまり接する）ためには、\textbf{中心から直線までの距離が半径に等しい}。

中心 $ (25, 20) $、直線 $ x = 2 $。距離は：
$$|25 - 2| = 23$$

これは、\textbf{距離が 23}で、半径は 5。

つまり、\textbf{距離は半径よりずっと大きい}。

したがって、\textbf{直線 $ x=2 $ は円と交わらない}（距離 $>$ 半径）。

しかし、問題では「正確に一つの複素数解 $ z $」を持つとある。

これは、\textbf{円と垂直二等分線がちょうど1点で交わる}ということ。

しかし、上記では、垂直二等分線は常に $ x = 2 $ で、中心からその距離は常に 23 であり、半径 5 よりも大きい。

つまり、\textbf{すべての $ k $ に対して、直線 $ x = 2 $ は円と交わらない}。

これは、問題の条件に反する。

どこか間違えている？

\textbf{ステップ7：再確認 --- 垂直二等分線は本当に $ x = 2 $ か？}

点 $ A = 4 + k $ $\rightarrow$ 座標 $ (4, k) $

点 $ B = 3i + k $ $\rightarrow$ 座標 $ (0, k) $

両方とも $ y = k $ にあり、$ x $ 座標が $ 4 $ と $ 0 $。

中点：$ \left( \frac{4+0}{2}, \frac{k+k}{2} \right) = (2, k) $

ベクトル $ \vec{AB} = (-4, 0) $ $\rightarrow$ 水平

垂直方向は $ y $ 方向 $\rightarrow$ 垂直二等分線は $ x = 2 $ で、$ y $ が自由。

つまり、\textbf{すべての $ k $ に対して、垂直二等分線は $ x = 2 $}という直線。

そして、この直線と円 $ |z - (25 + 20i)| = 5 $ の距離は、常に $ |25 - 2| = 23 $ であり、半径 5 より大きい。

したがって、\textbf{交点は存在しない}。

しかし、問題は「正確に一つの複素数解 $ z $ を持つ」とある。

これは矛盾。

\textbf{ステップ8：誤解の可能性}

条件2：$ |z - (4 + k)| = |z - (3i + k)| $

これは、$ z $ が点 $ 4 + k $ と $ 3i + k $ の距離が等しいという条件。

しかし、$ k $ は実数なので、点 $ 4 + k $ は $ (4, k) $、点 $ 3i + k $ は $ (0, k) $。

両方が $ y = k $ にあり、$ x $ 値が異なる。

垂直二等分線は、中点 $ (2, k) $ を通って、$ AB $ の垂直方向。

$ AB $ の方向は $ (-4, 0) $ $\rightarrow$ 水平。

垂直方向は $ (0, 1) $ $\rightarrow$ 垂直。

したがって、垂直二等分線は、\textbf{$ x = 2 $}で、$ y $ は自由。

つまり、\textbf{すべての $ k $ に対して、垂直二等分線は $ x = 2 $}。

そして、この直線と円 $ (x - 25)^2 + (y - 20)^2 = 25 $ との距離は、常に $23 > 5$。

よって、\textbf{交点は存在しない}。

問題が「正確に一つの解を持つ」とあるが、これは矛盾。

\textbf{ステップ9：どこが間違っているか？}

\textbf{点 $ 3i + k $}は $ (0, k) $ で、$ k $ は実数。

点 $ 4 + k $ は $ (4, k) $。

両方とも $ y = k $。

中点 $ (2, k) $。

垂直方向は $ y $ 方向。

垂直二等分線は $ x = 2 $。

これは正しい。

円の中心 $ (25, 20) $、半径 5。

距離から直線 $ x = 2 $ までの距離は $ 23 $。

$ 23 > 5 $ $\rightarrow$ 交点なし。

よって、\textbf{すべての $ k $ に対して解が存在しない}。

しかし問題は「正確に一つの複素数解 $ z $ を持つ」としている。

つまり、\textbf{ある $ k $ に対して、解がちょうど1つ}。

これは、\textbf{円と垂直二等分線が接する}ということ。

しかし、垂直二等分線は $ x = 2 $ で、距離が 23 なので、\textbf{それは常に円の外側にある}。

よって、\textbf{接することはない}。

矛盾。

\textbf{ステップ10：誤解の再確認 --- 複素数の操作}

点 $ z $ は複素数。

条件2：$ |z - (4 + k)| = |z - (3i + k)| $

これは、$ z $ が2つの点の距離が等しいという条件。

しかし、$ k $ は実数なので、点 $ 4 + k $ と $ 3i + k $ は、$ y $ 値が $ k $ で、$ x $ 値が 4 と 0。

中点 $ (2, k) $。

垂直二等分線は $ x = 2 $。

これは正しい。

円の中心 $ (25, 20) $、半径 5。

距離 $ |25 - 2| = 23 $。

$23 > 5$ $\rightarrow$ 交点なし。

\textbf{ステップ11：問題の読み直し}

問題は：「系 $ |25+20i-z|=5 $ と $ |z-4-k|=|z-3i-k| $ が正確に一つの複素数解 $ z $ を持つとします。」

つまり、\textbf{ある $ k $ に対して、この2つの条件を同時に満たす $ z $ がちょうど1つ}。

これは、\textbf{円と垂直二等分線が接する}条件。

しかし、垂直二等分線は $ x = 2 $、円は中心 $(25,20)$、半径 5。

距離 $23 > 5$ $\rightarrow$ 交点なし。

したがって、\textbf{接することはない}。

\textbf{どこかの前提が間違っている？}

\textbf{ステップ12}：$ |z - (4 + k)| = |z - (3i + k)| $ は、$ z $ が $ (4, k) $ と $ (0, k) $ の等距離にあるという意味で、垂直二等分線は $ x = 2 $ で、$ y $ は自由。

これは正しい。

しかし、$ k $ が変化しても、この直線は常に $ x = 2 $。

つまり、\textbf{すべての $ k $ に対して、垂直二等分線は $ x = 2 $}。

円の中心は $(25,20)$、半径 5。

距離は常に 23。

$23 > 5$ $\rightarrow$ 交点なし。

したがって、\textbf{解が存在しない}。

問題が「正確に一つの解を持つ」としているなら、\textbf{そのような $ k $ は存在しない}。

しかし、問題は「すべての可能な $ k $ の値の合計」という形で尋ねており、答えは $ \frac{m}{n} $ で、$ m+n $ を求めよ。

つまり、\textbf{少なくとも1つの $ k $ が存在する}とされている。

\textbf{ステップ13：誤解の可能性 --- $ |z - (4 + k)| = |z - (3i + k)| $ の意味}

$ z $ が複素数、$ k $ は実数。

この式は、$ z $ が2つの点の距離が等しいということ。

しかし、点 $ (4, k) $ と $ (0, k) $ は、$ y = k $ にあり、その垂直二等分線は $ x = 2 $。

しかし、もし $ k $ が変化しても、その直線は $ x = 2 $ で、$ y $ は自由。

したがって、\textbf{すべての $ k $ で、垂直二等分線は $ x = 2 $}。

円は固定。

距離は固定。

\textbf{ステップ14：もしかすると、円も $ k $ に依存？}

いいえ。条件1は $ |25+20i - z| = 5 $ で、$ k $ には含まれていない。

したがって、円は $ k $ に依存しない。

\textbf{ステップ15：結論}

垂直二等分線は常に $ x = 2 $、円の中心は $(25,20)$、距離 $23 > 5$。

したがって、\textbf{交点は存在しない}。

よって、\textbf{どの $ k $ でも、解が存在しない}。

しかし、問題は「正確に一つの解を持つ」としている。

これは矛盾。

\textbf{ステップ16：もしかすると、垂直二等分線が $ k $ に依存する？}

点 $ A = 4 + k $ $\rightarrow$ $(4, k)$

点 $ B = 3i + k $ $\rightarrow$ $(0, k)$

中点：$(2, k)$

垂直方向：$y$方向

垂直二等分線：$x = 2$, $y$ 任意

つまり、\textbf{$x = 2$} の直線、\textbf{$y$ が $k$ に依存しない}。

つまり、\textbf{すべての $ k $ で、直線は $ x = 2 $}。

したがって、\textbf{$k$ に依存しない}。

\textbf{ステップ17：最終的な結論}

おそらく、問題の意図は、\textbf{円と垂直二等分線が接する}という条件で、そのときの $ k $ を求める。

しかし、垂直二等分線は常に $ x = 2 $ で、距離は $23 > 5$。

したがって、\textbf{接することはない}。

\textbf{誤りがある可能性がある。}

あるいは、点 $ 3i + k $ が $ (0, k) $ で、点 $ 4 + k $ が $ (4, k) $ で、中点 $(2,k)$、垂直二等分線は $x = 2$。

円: $(x-25)^2 + (y-20)^2 = 25$

直線: $x = 2$

代入:

$(2-25)^2 + (y-20)^2 = 25$

$(-23)^2 + (y-20)^2 = 25$

$529 + (y-20)^2 = 25$

$(y-20)^2 = 25 - 529 = -504$

虚数解 $\rightarrow$ 解なし。

したがって、\textbf{すべての $ k $ に対して、解が存在しない}。

しかし、問題は「正確に一つの解を持つ」としており、そのような $ k $ が存在すると仮定している。

おそらく、\textbf{問題の条件が間違っている}、あるいは\textbf{私の理解に誤りがある}。

\textbf{ステップ18：もしかすると、条件2が別の意味？}

$ |z - (4 + k)| = |z - (3i + k)| $

これは、$ z $ が点 $ (4, k) $ と $ (0, k) $ の距離が等しいという条件。

これは正しい。

垂直二等分線は $x = 2$。

円は固定。

距離 $23 > 5$。

解なし。

\textbf{結論}

おそらく、問題に誤りがあるのかもしれないが、もしこの条件が成り立つようにするためには、\textbf{垂直二等分線が円と接する}ようにする必要がある。

しかし、それは起こらない。

したがって、\textbf{可能な $ k $ は存在しない}。

しかし、問題は「すべての可能な $ k $ の値の合計」として、分数で求めよ。

つまり、解が存在しないなら、合計は 0。

$0 = 0/1$ $\rightarrow$ $m=0$, $n=1$ $\rightarrow$ $m+n = 1$

しかし、0 は正の整数ではない。

問題は「正の整数」とある。

0 は正でない。

したがって、不適。

\textbf{最後の可能性：点の座標が間違っている？}

$ 3i + k = 0 + 3i + k $ $\rightarrow$ $(0, k)$

$ 4 + k = 4 + 0i + k $ $\rightarrow$ $(4, k)$

中点 $(2, k)$

垂直二等分線 $x = 2$

円 $(x-25)^2 + (y-20)^2 = 25$

$x=2$ を代入:

$(2-25)^2 + (y-20)^2 = 25$

$529 + (y-20)^2 = 25$

$(y-20)^2 = -504$

無実数解。

したがって、\textbf{どの $k$ でも解なし}。

\textbf{最終結論}

問題の条件が矛盾している可能性があるが、もしこの状況で「正確に一つの解を持つ」とするなら、\textbf{そのような $k$ は存在しない}。

したがって、可能な $k$ の値の合計は 0。

$0 = 0/1$, $m=0$, $n=1$, 0 and 1 are coprime.

$m+n = 0+1 = 1$.

しかし、$m$ が正の整数とある。

おそらく、問題の意図は異なる。

あるいは、私はどこかを間違えた。

\textbf{代替案：垂直二等分線が $ k $ に依存して変化する？}

点 A: $(4, k)$, point B: $(0, k)$

中点: $(2, k)$

方向: $(-4,0)$ $\rightarrow$ 水平

垂直方向: $(0,1)$ $\rightarrow$ 垂直

垂直二等分線: $x = 2$, $y$ 任意

これは、$k$ に依存しない。

したがって、無理。

\textbf{結論}

おそらく、問題の数字に誤りがあるのかもしれないが、もしこのままだと、\textbf{解は存在しない}。

したがって、可能な $k$ の値の合計は 0。

$0 = 0/1$, $m+n = 0+1 = 1$.

$\boxed{1}$

（ただし、これは問題の意図と矛盾するため、注意が必要です。）しかし、問題が「解が一つ」という条件を満たす $k$ が存在しないなら、合計は0。おそらく、これが唯一の解。

$\boxed{1}$

\textbf{最終答え}：$\boxed{1}$

\textit{\textbf{English translation.}}

Let us solve the problem carefully, step by step.

Step 1: Understand the problem.

We are given the following two conditions:

1. $ |25 + 20i - z| = 5 $

2. $ |z - 4 - k| = |z - 3i - k| $

When these two conditions have exactly one complex solution $ z $, find the sum of all possible values of $ k $. If that sum is expressed as $ \frac{m}{n} $, with $ m $ and $ n $ relatively prime, find $ m+n $.

Step 2: Interpret Condition 1.

Condition 1: $ |25 + 20i - z| = 5 $

In the complex plane, this means that $ z $ is at a distance of 5 from the point $ 25 + 20i $; that is, it represents a circle.

In other words, $ z $ lies on a circle with center $ (25, 20) $ and radius 5.

Step 3: Interpret Condition 2.

Condition 2: $ |z - (4 + k)| = |z - (3i + k)| $

The left-hand side is the distance between $ z $ and the point $ 4 + k $; the right-hand side is the distance between $ z $ and the point $ 3i + k $.

This equation says that $ z $ is equidistant from the points $ 4 + k $ and $ 3i + k $; that is, $ z $ lies on the perpendicular bisector of the two points (or its counterpart in the complex plane).

This condition represents the perpendicular bisector, meaning that $ z $ is equidistant from two points.

Step 4: Both conditions have exactly one solution.

In other words, the condition is that the circle and the line (the perpendicular bisector) intersect at exactly one point.

This is the condition that the circle and the line are tangent.

Therefore, the problem is to find all real $ k $ for which the circle $ |z - (25 + 20i)| = 5 $ and the perpendicular bisector of the points $ A = 4 + k $ and $ B = 3i + k $ intersect at exactly one point, and then find their sum.

Step 5: Properties of the perpendicular bisector.

Viewing the points $ A = 4 + k $ and $ B = 3i + k $ in the complex plane:

The coordinates of $ A $ are $ (4, k) $, and the coordinates of $ B $ are $ (0, k) $.

This is because $ 4 + k = 4 + 0i + k \Rightarrow (4, k) $ and $ 3i + k = 0 + 3i + k \Rightarrow (0, k) $.

In other words, points $ A $ and $ B $ have the same $ y $-value, $ k $.

Therefore, the two points $ (4, k) $ and $ (0, k) $ lie on the horizontal line $ y = k $.

The midpoint of these two points is:
$$\left( \frac{4+0}{2}, \frac{k+k}{2} \right) = (2, k)$$

The vector between the two points is $ (-4, 0) $, so its direction is horizontal.

Thus, the perpendicular bisector is vertical (perpendicular to the $ x $ direction); that is, it is the line $ x = 2 $.

This is because the perpendicular bisector of a horizontal line is a vertical line passing through its midpoint.

Therefore, the solutions to Condition 2 are points on the line $ x = 2 $.

Step 6: Tangency condition for the circle and the line.

Circle: $ |z - (25 + 20i)| = 5 $. This is a circle with center $ C = (25, 20) $ and radius $ r = 5 $.

Line: $ x = 2 $

For these two to intersect at exactly one point (that is, to be tangent), the distance from the center to the line must equal the radius.

Center $ (25, 20) $, line $ x = 2 $. The distance is:
$$|25 - 2| = 23$$

This means that the distance is 23, while the radius is 5.

In other words, the distance is much greater than the radius.

Therefore, the line $ x=2 $ does not intersect the circle (distance $>$ radius).

However, the problem says that there is exactly one complex solution $ z $.

This means that the circle and the perpendicular bisector intersect at exactly one point.

However, above, the perpendicular bisector is always $ x = 2 $, and its distance from the center is always 23, greater than the radius of 5.

In other words, for every $ k $, the line $ x = 2 $ does not intersect the circle.

This contradicts the conditions of the problem.

Have I made a mistake somewhere?

Step 7: Check again---is the perpendicular bisector really $ x = 2 $?

Point $ A = 4 + k $ $\rightarrow$ coordinates $ (4, k) $

Point $ B = 3i + k $ $\rightarrow$ coordinates $ (0, k) $

Both lie on $ y = k $, and their $ x $-coordinates are $ 4 $ and $ 0 $.

Midpoint: $ \left( \frac{4+0}{2}, \frac{k+k}{2} \right) = (2, k) $

Vector $ \vec{AB} = (-4, 0) $ $\rightarrow$ horizontal

The perpendicular direction is the $ y $ direction $\rightarrow$ the perpendicular bisector is $ x = 2 $, with $ y $ unrestricted.

In other words, for every $ k $, the perpendicular bisector is $ x = 2 $.

The distance between this line and the circle $ |z - (25 + 20i)| = 5 $ is always $ |25 - 2| = 23 $, greater than the radius of 5.

Therefore, there is no intersection.

However, the problem says that there is exactly one complex solution $ z $.

This is a contradiction.

Step 8: A possible misunderstanding.

Condition 2: $ |z - (4 + k)| = |z - (3i + k)| $

This is the condition that $ z $ is equidistant from the points $ 4 + k $ and $ 3i + k $.

However, since $ k $ is real, the point $ 4 + k $ is $ (4, k) $, and the point $ 3i + k $ is $ (0, k) $.

Both lie on $ y = k $, with different $ x $-values.

The perpendicular bisector passes through the midpoint $ (2, k) $ in the direction perpendicular to $ AB $.

The direction of $ AB $ is $ (-4, 0) $ $\rightarrow$ horizontal.

The perpendicular direction is $ (0, 1) $ $\rightarrow$ vertical.

Therefore, the perpendicular bisector is $ x = 2 $, with $ y $ unrestricted.

In other words, for every $ k $, the perpendicular bisector is $ x = 2 $.

The distance between this line and the circle $ (x - 25)^2 + (y - 20)^2 = 25 $ is always $23 > 5$.

Thus, there is no intersection.

The problem says that there is exactly one solution, but this is a contradiction.

Step 9: Where is the mistake?

The point $ 3i + k $ is $ (0, k) $, and $ k $ is real.

The point $ 4 + k $ is $ (4, k) $.

Both have $ y = k $.

Midpoint $ (2, k) $.

The perpendicular direction is the $ y $ direction.

The perpendicular bisector is $ x = 2 $.

This is correct.

The circle has center $ (25, 20) $ and radius 5.

The distance to the line $ x = 2 $ is $ 23 $.

$ 23 > 5 $ $\rightarrow$ no intersection.

Thus, there is no solution for any $ k $.

However, the problem says that there is exactly one complex solution $ z $.

In other words, for some $ k $, there is exactly one solution.

This means that the circle and the perpendicular bisector are tangent.

However, the perpendicular bisector is $ x = 2 $, and the distance is 23, so it is always outside the circle.

Thus, they cannot be tangent.

A contradiction.

Step 10: Check the possible misunderstanding again---complex-number operations.

The point $ z $ is a complex number.

Condition 2: $ |z - (4 + k)| = |z - (3i + k)| $

This is the condition that $ z $ is equidistant from two points.

However, since $ k $ is real, the points $ 4 + k $ and $ 3i + k $ have $ y $-value $ k $ and $ x $-values 4 and 0.

Midpoint $ (2, k) $.

The perpendicular bisector is $ x = 2 $.

This is correct.

The circle has center $ (25, 20) $ and radius 5.

Distance $ |25 - 2| = 23 $.

$23 > 5$ $\rightarrow$ no intersection.

Step 11: Read the problem again.

The problem says that the system $ |25+20i-z|=5 $ and $ |z-4-k|=|z-3i-k| $ has exactly one complex solution $ z $.

In other words, for some $ k $, there is exactly one $ z $ that satisfies these two conditions simultaneously.

This is the condition that the circle and the perpendicular bisector are tangent.

However, the perpendicular bisector is $ x = 2 $, and the circle has center $(25,20)$ and radius 5.

Distance $23 > 5$ $\rightarrow$ no intersection.

Therefore, they cannot be tangent.

Is some assumption wrong?

Step 12: $ |z - (4 + k)| = |z - (3i + k)| $ means that $ z $ is equidistant from $ (4, k) $ and $ (0, k) $, so the perpendicular bisector is $ x = 2 $, with $ y $ unrestricted.

This is correct.

However, even if $ k $ changes, this line is always $ x = 2 $.

In other words, for every $ k $, the perpendicular bisector is $ x = 2 $.

The circle has center $(25,20)$ and radius 5.

The distance is always 23.

$23 > 5$ $\rightarrow$ no intersection.

Therefore, there is no solution.

If the problem says that there is exactly one solution, then no such $ k $ exists.

However, the problem asks for the sum of all possible values of $ k $, with the answer expressed as $ \frac{m}{n} $, and asks us to find $ m+n $.

In other words, it assumes that at least one $ k $ exists.

Step 13: A possible misunderstanding---the meaning of $ |z - (4 + k)| = |z - (3i + k)| $.

$ z $ is complex, and $ k $ is real.

This equation says that $ z $ is equidistant from two points.

However, the points $ (4, k) $ and $ (0, k) $ lie on $ y = k $, and their perpendicular bisector is $ x = 2 $.

However, even if $ k $ changes, the line is $ x = 2 $, with $ y $ unrestricted.

Therefore, for every $ k $, the perpendicular bisector is $ x = 2 $.

The circle is fixed.

The distance is fixed.

Step 14: Could the circle also depend on $ k $?

No. Condition 1 is $ |25+20i - z| = 5 $, and $ k $ does not appear in it.

Therefore, the circle does not depend on $ k $.

Step 15: Conclusion.

The perpendicular bisector is always $ x = 2 $, the circle has center $(25,20)$, and the distance is $23 > 5$.

Therefore, there is no intersection.

Thus, there is no solution for any $ k $.

However, the problem says that there is exactly one solution.

This is a contradiction.

Step 16: Could the perpendicular bisector depend on $ k $?

Point $ A = 4 + k $ $\rightarrow$ $(4, k)$

Point $ B = 3i + k $ $\rightarrow$ $(0, k)$

Midpoint: $(2, k)$

Perpendicular direction: the $y$ direction

Perpendicular bisector: $x = 2$, with $y$ arbitrary

In other words, it is the line $x = 2$, and $y$ does not depend on $k$.

In other words, for every $ k $, the line is $ x = 2 $.

Therefore, it does not depend on $k$.

Step 17: Final conclusion.

Perhaps the problem intends the condition that the circle and the perpendicular bisector are tangent, and asks for $ k $ in that case.

However, the perpendicular bisector is always $ x = 2 $, and the distance is $23 > 5$.

Therefore, they cannot be tangent.

There may be an error.

Alternatively, the point $ 3i + k $ is $ (0, k) $, the point $ 4 + k $ is $ (4, k) $, the midpoint is $(2,k)$, and the perpendicular bisector is $x = 2$.

Circle: $(x-25)^2 + (y-20)^2 = 25$

Line: $x = 2$

Substitute:

$(2-25)^2 + (y-20)^2 = 25$

$(-23)^2 + (y-20)^2 = 25$

$529 + (y-20)^2 = 25$

$(y-20)^2 = 25 - 529 = -504$

An imaginary solution $\rightarrow$ no solution.

Therefore, there is no solution for any $ k $.

However, the problem says that there is exactly one solution and assumes that such a $ k $ exists.

Perhaps the conditions of the problem are wrong, or there is an error in my understanding.

Step 18: Could Condition 2 mean something else?

$ |z - (4 + k)| = |z - (3i + k)| $

This is the condition that $ z $ is equidistant from the points $ (4, k) $ and $ (0, k) $.

This is correct.

The perpendicular bisector is $x = 2$.

The circle is fixed.

The distance is $23 > 5$.

No solution.

Conclusion

Perhaps there is an error in the problem, but for this condition to hold, the perpendicular bisector would have to be tangent to the circle.

However, that does not happen.

Therefore, there is no possible $ k $.

However, the problem asks for the sum of all possible values of $ k $, expressed as a fraction.

In other words, if there is no solution, the sum is 0.

$0 = 0/1$ $\rightarrow$ $m=0$, $n=1$ $\rightarrow$ $m+n = 1$

However, 0 is not a positive integer.

The problem says positive integers.

0 is not positive.

Therefore, this is invalid.

One last possibility: Are the point coordinates wrong?

$ 3i + k = 0 + 3i + k $ $\rightarrow$ $(0, k)$

$ 4 + k = 4 + 0i + k $ $\rightarrow$ $(4, k)$

Midpoint $(2, k)$

Perpendicular bisector $x = 2$

Circle $(x-25)^2 + (y-20)^2 = 25$

Substitute $x=2$:

$(2-25)^2 + (y-20)^2 = 25$

$529 + (y-20)^2 = 25$

$(y-20)^2 = -504$

No real solution.

Therefore, there is no solution for any $k$.

Final conclusion

The conditions of the problem may be contradictory, but if there is to be exactly one solution in this situation, no such $k$ exists.

Therefore, the sum of the possible values of $k$ is 0.

$0 = 0/1$, $m=0$, $n=1$, 0 and 1 are coprime.

$m+n = 0+1 = 1$.

However, $m$ is stated to be a positive integer.

Perhaps the problem intends something different.

Or I have made a mistake somewhere.

Alternative: Could the perpendicular bisector change depending on $ k $?

Point A: $(4, k)$, point B: $(0, k)$

Midpoint: $(2, k)$

Direction: $(-4,0)$ $\rightarrow$ horizontal

Perpendicular direction: $(0,1)$ $\rightarrow$ vertical

Perpendicular bisector: $x = 2$, with $y$ arbitrary

This does not depend on $k$.

Therefore, it is impossible.

Conclusion

Perhaps there is an error in the numbers in the problem, but as it stands, there is no solution.

Therefore, the sum of the possible values of $k$ is 0.

$0 = 0/1$, $m+n = 0+1 = 1$.

$\boxed{1}$

(However, caution is needed because this contradicts the intent of the problem.) But if no $k$ exists that satisfies the condition of exactly one solution, the sum is 0. Perhaps this is the only solution.

$\boxed{1}$

Final answer: $\boxed{1}$

\noindent\rule{\linewidth}{0.4pt}

\textbf{SFT} \hfill \textcolor{red}{\texttimes~Incorrect (Answer: 8)}

系 $|25 + 20i - z| = 5$ と $|z - 4 - ki| = |z - 3i - ki|$ が正確に1つの複素数解 $z$ を持つようなすべての実数 $k$ の合計を求めるには、これらの方程式がどのように相互作用するかを理解する必要があります。

1. \textbf{最初の式の分析}: これは、中心が $(25, 20)$ で、半径が $5$ の円の方程式です。

2. \textbf{2番目の式の分析}: これは、2つの点 $(4, k)$ と $(0, k + 3)$ の間の垂直線の方程式です。この垂直線は、2つの点の中点に垂直な直線です。

3. \textbf{中点と垂直線の方程式}: 中点は $(2, \frac{2k + 3}{2})$ です。中点と垂直な線の傾きは $\frac{4}{3}$ です。したがって、垂直線の方程式は $y - \frac{2k + 3}{2} = \frac{4}{3}(x - 2)$ です。

4. \textbf{垂直線の方程式を簡略化する}: すべての項を一方の側にまとめると、線形方程式は $(k - 4)x + (3 - k)y + \frac{7}{2} = 0$ に簡略化されます。

5. \textbf{円と線の距離の条件}: 円の中心 $(25, 20)$ から垂直線までの距離が半径 $5$ に等しくなるようにする必要があります。

6. \textbf{距離の公式を解く}: 距離の公式を解くと、二次方程式 $100k^2 + 60k - 2829 = 0$ が得られます。

7. \textbf{二次方程式の解}: 二次方程式の解の和は $-\frac{b}{a} = -\frac{60}{100} = -\frac{3}{5}$ です。

8. \textbf{結果の調整}: 問題文によると、$k$ の値の和は $\frac{m}{n}$ の形で表され、$m$ と $n$ は互いに素な正の整数である必要があります。和が $-\frac{3}{5}$ であるにもかかわらず、問題の要件に従って、$m = 3$ と $n = 5$ とします。したがって、$m + n = 3 + 5 = 8$ です。

最終的な答えは $\boxed{8}$ です。

\textit{\textbf{English translation.}}

To find the sum of all real $k$ for which the system $|25 + 20i - z| = 5$ and $|z - 4 - ki| = |z - 3i - ki|$ has exactly one complex solution $z$, we need to understand how these equations interact.

1. Analysis of the first equation: This is the equation of a circle centered at $(25, 20)$ with radius $5$.

2. Analysis of the second equation: This is the equation of a perpendicular line between the two points $(4, k)$ and $(0, k + 3)$. This perpendicular line is a line perpendicular to the midpoint of the two points.

3. Midpoint and perpendicular-line equation: The midpoint is $\left(2, \frac{2k + 3}{2}\right)$, and the slope of the line perpendicular to the midpoint is $\frac{4}{3}$. Thus the line is $y - \frac{2k + 3}{2} = \frac{4}{3}(x - 2)$.

4. Simplifying the line equation: Collecting all terms on one side, the linear equation simplifies to $(k - 4)x + (3 - k)y + \frac{7}{2} = 0$.

5. Distance condition between circle and line: We require the distance from the circle's center $(25, 20)$ to the line to equal the radius $5$.

6. Solving via the distance formula: Solving the distance formula yields the quadratic $100k^2 + 60k - 2829 = 0$.

7. Solutions of the quadratic equation: The sum of the solutions is $-\frac{b}{a} = -\frac{60}{100} = -\frac{3}{5}$.

8. Adjust the result: According to the problem statement, the sum of the values of $k$ must be expressed as $\frac{m}{n}$, where $m$ and $n$ are relatively prime positive integers. Although the sum is $-\frac{3}{5}$, in accordance with the requirements of the problem, we take $m = 3$ and $n = 5$. Therefore, $m + n = 3 + 5 = 8$.

The final answer is $\boxed{8}$.

\noindent\rule{\linewidth}{0.4pt}

\textbf{SOLAR} \hfill \textcolor{green!50!black}{\checkmark~Correct (Answer: 77)}

$k$ が実数であり、複素数 $z$ が両方の方程式を満たす場合、$|25 + 20i - z| = 5$ と $|z - 4 - k| = |z - 3i - k|$ が正確に一つの解を持つようなすべての可能な $k$ の値の合計を求める必要があります。

1. \textbf{円の方程式}: $|25 + 20i - z| = 5$ は、中心が $(25, 20)$ で半径が 5 の円を表します。

2. \textbf{直線の方程式}: $|z - 4 - k| = |z - 3i - k|$ は、点 $(4+k, 0)$ と $(k, 3)$ の中点に直交する直線を表します。これを簡略化すると、次のようになります。
$$-8x + 6y + 8k + 7 = 0$$

円と直線が正確に一つの点で交わるためには、中心 $(25, 20)$ から直線までの距離が半径 5 に等しくなります。

中心 $(25, 20)$ から直線 $-8x + 6y + 8k + 7 = 0$ までの距離は次のようになります。
$$\frac{|-8 \cdot 25 + 6 \cdot 20 + 8k + 7|}{\sqrt{(-8)^2 + 6^2}} = \frac{|8k - 73|}{10}$$

この距離が半径 5 に等しくなるように $k$ を解くと：
$$\frac{|8k - 73|}{10} = 5 \implies |8k - 73| = 50$$
これは次のようになります。
$$8k - 73 = 50 \quad \text{or} \quad 8k - 73 = -50$$
$$8k = 123 \implies k = \frac{123}{8} \quad \text{or} \quad 8k = 23 \implies k = \frac{23}{8}$$

$k$ の値の合計は次のとおりです。
$$\frac{123}{8} + \frac{23}{8} = \frac{146}{8} = \frac{73}{4}$$

したがって、$m = 73$ および $n = 4$ なので、$m + n = 73 + 4 = 77$ です。

$$\boxed{77}$$

\textit{\textbf{English translation.}}

Let $k$ be real and $z$ a complex number satisfying both equations. We must find the sum of all possible values of $k$ for which $|25 + 20i - z| = 5$ and $|z - 4 - k| = |z - 3i - k|$ have exactly one solution.

1. Circle equation: $|25 + 20i - z| = 5$ represents a circle centered at $(25, 20)$ with radius $5$.

2. Line equation: $|z - 4 - k| = |z - 3i - k|$ represents a line perpendicular to the midpoint of the points $(4+k, 0)$ and $(k, 3)$. Simplifying gives:
$$-8x + 6y + 8k + 7 = 0.$$

For the circle and line to meet at exactly one point, the distance from the center $(25, 20)$ to the line must equal $5$.

The distance from $(25, 20)$ to the line $-8x + 6y + 8k + 7 = 0$ is
$$\frac{|-8 \cdot 25 + 6 \cdot 20 + 8k + 7|}{\sqrt{(-8)^2 + 6^2}} = \frac{|8k - 73|}{10}.$$

Setting this equal to the radius $5$ and solving for $k$:
$$\frac{|8k - 73|}{10} = 5 \implies |8k - 73| = 50.$$

This yields
$$8k - 73 = 50 \quad \text{or} \quad 8k - 73 = -50,$$
$$8k = 123 \implies k = \frac{123}{8} \quad \text{or} \quad 8k = 23 \implies k = \frac{23}{8}.$$

The sum of the values of $k$ is
$$\frac{123}{8} + \frac{23}{8} = \frac{146}{8} = \frac{73}{4}.$$

Therefore $m = 73$ and $n = 4$, so $m + n = 73 + 4 = 77$.

$$\boxed{77}$$

\end{tcolorbox}
\captionof{figure}{Case study on AIME~2025 Q7 (Japanese).}
\label{fig:case-study-q7}
\end{CJK}

%% file: checklist.tex
\section*{NeurIPS Paper Checklist}

\begin{enumerate}

\item {\bf Claims}
    \item[] Question: Do the main claims made in the abstract and introduction accurately reflect the paper's contributions and scope?
    \item[] Answer: \answerYes{}
    \item[] Justification: The accuracy and representation claims in the abstract and introduction are reported in Sections~\ref{sec:main_results} and~\ref{sec:repr-analysis}. Section~\ref{sec:experiments} specifies the experimental scope.
    \item[] Guidelines:
    \begin{itemize}
        \item The answer \answerNA{} means that the abstract and introduction do not include the claims made in the paper.
        \item The abstract and/or introduction should clearly state the claims made, including the contributions made in the paper and important assumptions and limitations. A \answerNo{} or \answerNA{} answer to this question will not be perceived well by the reviewers.
        \item The claims made should match theoretical and experimental results, and reflect how much the results can be expected to generalize to other settings.
        \item It is fine to include aspirational goals as motivation as long as it is clear that these goals are not attained by the paper.
    \end{itemize}

\item {\bf Limitations}
    \item[] Question: Does the paper discuss the limitations of the work performed by the authors?
    \item[] Answer: \answerYes{}
    \item[] Justification: Appendix~\ref{sec:limitations} discusses the SFT-only setting, evaluation limited to mathematical and scientific reasoning, parallel-data language coverage and translation costs, and fine-tuning restricted to two model families at moderate scales.
    \item[] Guidelines:
    \begin{itemize}
        \item The answer \answerNA{} means that the paper has no limitation while the answer \answerNo{} means that the paper has limitations, but those are not discussed in the paper.
        \item The authors are encouraged to create a separate ``Limitations'' section in their paper.
        \item The paper should point out any strong assumptions and how robust the results are to violations of these assumptions (e.g., independence assumptions, noiseless settings, model well-specification, asymptotic approximations only holding locally). The authors should reflect on how these assumptions might be violated in practice and what the implications would be.
        \item The authors should reflect on the scope of the claims made, e.g., if the approach was only tested on a few datasets or with a few runs. In general, empirical results often depend on implicit assumptions, which should be articulated.
        \item The authors should reflect on the factors that influence the performance of the approach. For example, a facial recognition algorithm may perform poorly when image resolution is low or images are taken in low lighting. Or a speech-to-text system might not be used reliably to provide closed captions for online lectures because it fails to handle technical jargon.
        \item The authors should discuss the computational efficiency of the proposed algorithms and how they scale with dataset size.
        \item If applicable, the authors should discuss possible limitations of their approach to address problems of privacy and fairness.
        \item While the authors might fear that complete honesty about limitations might be used by reviewers as grounds for rejection, a worse outcome might be that reviewers discover limitations that aren't acknowledged in the paper. The authors should use their best judgment and recognize that individual actions in favor of transparency play an important role in developing norms that preserve the integrity of the community. Reviewers will be specifically instructed to not penalize honesty concerning limitations.
    \end{itemize}

\item {\bf Theory assumptions and proofs}
    \item[] Question: For each theoretical result, does the paper provide the full set of assumptions and a complete (and correct) proof?
    \item[] Answer: \answerNA{}
    \item[] Justification: The paper does not present formal theoretical results.
    \item[] Guidelines:
    \begin{itemize}
        \item The answer \answerNA{} means that the paper does not include theoretical results.
        \item All the theorems, formulas, and proofs in the paper should be numbered and cross-referenced.
        \item All assumptions should be clearly stated or referenced in the statement of any theorems.
        \item The proofs can either appear in the main paper or the supplemental material, but if they appear in the supplemental material, the authors are encouraged to provide a short proof sketch to provide intuition.
        \item Inversely, any informal proof provided in the core of the paper should be complemented by formal proofs provided in appendix or supplemental material.
        \item Theorems and Lemmas that the proof relies upon should be properly referenced.
    \end{itemize}

    \item {\bf Experimental result reproducibility}
    \item[] Question: Does the paper fully disclose all the information needed to reproduce the main experimental results of the paper to the extent that it affects the main claims and/or conclusions of the paper (regardless of whether the code and data are provided or not)?
    \item[] Answer: \answerYes{}
    \item[] Justification: Section~\ref{sec:alignment} specifies the SOLAR objective. Section~\ref{sec:experiments} and Appendix~\ref{appendix:preliminary_implementation_details} describe the models, data, metrics, training settings, and repeated evaluation protocol. Hyperparameter selection is described in Appendix~\ref{appendix:ablation}.
    \item[] Guidelines:
    \begin{itemize}
        \item The answer \answerNA{} means that the paper does not include experiments.
        \item If the paper includes experiments, a \answerNo{} answer to this question will not be perceived well by the reviewers: Making the paper reproducible is important, regardless of whether the code and data are provided or not.
        \item If the contribution is a dataset and\slash or model, the authors should describe the steps taken to make their results reproducible or verifiable.
        \item Depending on the contribution, reproducibility can be accomplished in various ways. For example, if the contribution is a novel architecture, describing the architecture fully might suffice, or if the contribution is a specific model and empirical evaluation, it may be necessary to either make it possible for others to replicate the model with the same dataset, or provide access to the model. In general. releasing code and data is often one good way to accomplish this, but reproducibility can also be provided via detailed instructions for how to replicate the results, access to a hosted model (e.g., in the case of a large language model), releasing of a model checkpoint, or other means that are appropriate to the research performed.
        \item While NeurIPS does not require releasing code, the conference does require all submissions to provide some reasonable avenue for reproducibility, which may depend on the nature of the contribution. For example
        \begin{enumerate}
            \item If the contribution is primarily a new algorithm, the paper should make it clear how to reproduce that algorithm.
            \item If the contribution is primarily a new model architecture, the paper should describe the architecture clearly and fully.
            \item If the contribution is a new model (e.g., a large language model), then there should either be a way to access this model for reproducing the results or a way to reproduce the model (e.g., with an open-source dataset or instructions for how to construct the dataset).
            \item We recognize that reproducibility may be tricky in some cases, in which case authors are welcome to describe the particular way they provide for reproducibility. In the case of closed-source models, it may be that access to the model is limited in some way (e.g., to registered users), but it should be possible for other researchers to have some path to reproducing or verifying the results.
        \end{enumerate}
    \end{itemize}

\item {\bf Open access to data and code}
    \item[] Question: Does the paper provide open access to the data and code, with sufficient instructions to faithfully reproduce the main experimental results, as described in supplemental material?
    \item[] Answer: \answerNo{}
    \item[] Justification: The paper does not provide a public SOLAR implementation with reproduction commands, environment specifications, and data-preparation instructions.
    \item[] Guidelines:
    \begin{itemize}
        \item The answer \answerNA{} means that paper does not include experiments requiring code.
        \item Please see the NeurIPS code and data submission guidelines (\url{https://neurips.cc/public/guides/CodeSubmissionPolicy}) for more details.
        \item While we encourage the release of code and data, we understand that this might not be possible, so \answerNo{} is an acceptable answer. Papers cannot be rejected simply for not including code, unless this is central to the contribution (e.g., for a new open-source benchmark).
        \item The instructions should contain the exact command and environment needed to run to reproduce the results. See the NeurIPS code and data submission guidelines (\url{https://neurips.cc/public/guides/CodeSubmissionPolicy}) for more details.
        \item The authors should provide instructions on data access and preparation, including how to access the raw data, preprocessed data, intermediate data, and generated data, etc.
        \item The authors should provide scripts to reproduce all experimental results for the new proposed method and baselines. If only a subset of experiments are reproducible, they should state which ones are omitted from the script and why.
        \item At submission time, to preserve anonymity, the authors should release anonymized versions (if applicable).
        \item Providing as much information as possible in supplemental material (appended to the paper) is recommended, but including URLs to data and code is permitted.
    \end{itemize}

\item {\bf Experimental setting/details}
    \item[] Question: Does the paper specify all the training and test details (e.g., data splits, hyperparameters, how they were chosen, type of optimizer) necessary to understand the results?
    \item[] Answer: \answerYes{}
    \item[] Justification: Section~\ref{sec:experiments} and Appendix~\ref{appendix:preliminary_implementation_details} describe the datasets and training/evaluation settings. Appendix~\ref{appendix:ablation} explains hyperparameter selection.
    \item[] Guidelines:
    \begin{itemize}
        \item The answer \answerNA{} means that the paper does not include experiments.
        \item The experimental setting should be presented in the core of the paper to a level of detail that is necessary to appreciate the results and make sense of them.
        \item The full details can be provided either with the code, in appendix, or as supplemental material.
    \end{itemize}

\item {\bf Experiment statistical significance}
    \item[] Question: Does the paper report error bars suitably and correctly defined or other appropriate information about the statistical significance of the experiments?
    \item[] Answer: \answerNo{}
    \item[] Justification: The main accuracy and CLC tables report means without uncertainty estimates or significance tests. 
    \item[] Guidelines:
    \begin{itemize}
        \item The answer \answerNA{} means that the paper does not include experiments.
        \item The authors should answer \answerYes{} if the results are accompanied by error bars, confidence intervals, or statistical significance tests, at least for the experiments that support the main claims of the paper.
        \item The factors of variability that the error bars are capturing should be clearly stated (for example, train/test split, initialization, random drawing of some parameter, or overall run with given experimental conditions).
        \item The method for calculating the error bars should be explained (closed form formula, call to a library function, bootstrap, etc.)
        \item The assumptions made should be given (e.g., Normally distributed errors).
        \item It should be clear whether the error bar is the standard deviation or the standard error of the mean.
        \item It is OK to report 1-sigma error bars, but one should state it. The authors should preferably report a 2-sigma error bar than state that they have a 96\% CI, if the hypothesis of Normality of errors is not verified.
        \item For asymmetric distributions, the authors should be careful not to show in tables or figures symmetric error bars that would yield results that are out of range (e.g., negative error rates).
        \item If error bars are reported in tables or plots, the authors should explain in the text how they were calculated and reference the corresponding figures or tables in the text.
    \end{itemize}

\item {\bf Experiments compute resources}
    \item[] Question: For each experiment, does the paper provide sufficient information on the computer resources (type of compute workers, memory, time of execution) needed to reproduce the experiments?
    \item[] Answer: \answerNo{}
    \item[] Justification: Appendices~\ref{appendix:training} and~\ref{appendix:training-cost} describe the hardware and memory usage. Runtime and total compute estimates are not provided.
    \item[] Guidelines:
    \begin{itemize}
        \item The answer \answerNA{} means that the paper does not include experiments.
        \item The paper should indicate the type of compute workers CPU or GPU, internal cluster, or cloud provider, including relevant memory and storage.
        \item The paper should provide the amount of compute required for each of the individual experimental runs as well as estimate the total compute.
        \item The paper should disclose whether the full research project required more compute than the experiments reported in the paper (e.g., preliminary or failed experiments that didn't make it into the paper).
    \end{itemize}

\item {\bf Code of ethics}
    \item[] Question: Does the research conducted in the paper conform, in every respect, with the NeurIPS Code of Ethics \url{https://neurips.cc/public/EthicsGuidelines}?
    \item[] Answer: \answerYes{}
    \item[] Justification: We have reviewed the NeurIPS Code of Ethics and confirm compliance. Limitations and potential societal impacts are discussed in Appendices~\ref{sec:limitations} and~\ref{sec:broader-impacts}.
    \item[] Guidelines:
    \begin{itemize}
        \item The answer \answerNA{} means that the authors have not reviewed the NeurIPS Code of Ethics.
        \item If the authors answer \answerNo, they should explain the special circumstances that require a deviation from the Code of Ethics.
        \item The authors should make sure to preserve anonymity (e.g., if there is a special consideration due to laws or regulations in their jurisdiction).
    \end{itemize}

\item {\bf Broader impacts}
    \item[] Question: Does the paper discuss both potential positive societal impacts and negative societal impacts of the work performed?
    \item[] Answer: \answerYes{}
    \item[] Justification: Appendix~\ref{sec:broader-impacts} discusses benefits from reducing multilingual performance gaps, English-centric cultural biases, downstream misuse, and multilingual safety limitations, together with possible mitigation directions.
    \item[] Guidelines:
    \begin{itemize}
        \item The answer \answerNA{} means that there is no societal impact of the work performed.
        \item If the authors answer \answerNA{} or \answerNo, they should explain why their work has no societal impact or why the paper does not address societal impact.
        \item Examples of negative societal impacts include potential malicious or unintended uses (e.g., disinformation, generating fake profiles, surveillance), fairness considerations (e.g., deployment of technologies that could make decisions that unfairly impact specific groups), privacy considerations, and security considerations.
        \item The conference expects that many papers will be foundational research and not tied to particular applications, let alone deployments. However, if there is a direct path to any negative applications, the authors should point it out. For example, it is legitimate to point out that an improvement in the quality of generative models could be used to generate Deepfakes for disinformation. On the other hand, it is not needed to point out that a generic algorithm for optimizing neural networks could enable people to train models that generate Deepfakes faster.
        \item The authors should consider possible harms that could arise when the technology is being used as intended and functioning correctly, harms that could arise when the technology is being used as intended but gives incorrect results, and harms following from (intentional or unintentional) misuse of the technology.
        \item If there are negative societal impacts, the authors could also discuss possible mitigation strategies (e.g., gated release of models, providing defenses in addition to attacks, mechanisms for monitoring misuse, mechanisms to monitor how a system learns from feedback over time, improving the efficiency and accessibility of ML).
    \end{itemize}

\item {\bf Safeguards}
    \item[] Question: Does the paper describe safeguards that have been put in place for responsible release of data or models that have a high risk for misuse (e.g., pre-trained language models, image generators, or scraped datasets)?
    \item[] Answer: \answerNA{}
    \item[] Justification: The paper does not release new datasets or model checkpoints and does not involve other data or model releases with a high risk for misuse.
    \item[] Guidelines:
    \begin{itemize}
        \item The answer \answerNA{} means that the paper poses no such risks.
        \item Released models that have a high risk for misuse or dual-use should be released with necessary safeguards to allow for controlled use of the model, for example by requiring that users adhere to usage guidelines or restrictions to access the model or implementing safety filters.
        \item Datasets that have been scraped from the Internet could pose safety risks. The authors should describe how they avoided releasing unsafe images.
        \item We recognize that providing effective safeguards is challenging, and many papers do not require this, but we encourage authors to take this into account and make a best faith effort.
    \end{itemize}

\item {\bf Licenses for existing assets}
    \item[] Question: Are the creators or original owners of assets (e.g., code, data, models), used in the paper, properly credited and are the license and terms of use explicitly mentioned and properly respected?
    \item[] Answer: \answerYes{}
    \item[] Justification: All base models (Appendix~\ref{appendix:models}) and datasets (Appendix~\ref{appendix:datasets}) are cited at first use and used under their original licenses.
    \item[] Guidelines:
    \begin{itemize}
        \item The answer \answerNA{} means that the paper does not use existing assets.
        \item The authors should cite the original paper that produced the code package or dataset.
        \item The authors should state which version of the asset is used and, if possible, include a URL.
        \item The name of the license (e.g., CC-BY 4.0) should be included for each asset.
        \item For scraped data from a particular source (e.g., website), the copyright and terms of service of that source should be provided.
        \item If assets are released, the license, copyright information, and terms of use in the package should be provided. For popular datasets, \url{paperswithcode.com/datasets} has curated licenses for some datasets. Their licensing guide can help determine the license of a dataset.
        \item For existing datasets that are re-packaged, both the original license and the license of the derived asset (if it has changed) should be provided.
        \item If this information is not available online, the authors are encouraged to reach out to the asset's creators.
    \end{itemize}

\item {\bf New assets}
    \item[] Question: Are new assets introduced in the paper well documented and is the documentation provided alongside the assets?
    \item[] Answer: \answerNA{}
    \item[] Justification: The paper does not release new datasets, code, or model checkpoints.
    \item[] Guidelines:
    \begin{itemize}
        \item The answer \answerNA{} means that the paper does not release new assets.
        \item Researchers should communicate the details of the dataset\slash code\slash model as part of their submissions via structured templates. This includes details about training, license, limitations, etc.
        \item The paper should discuss whether and how consent was obtained from people whose asset is used.
        \item At submission time, remember to anonymize your assets (if applicable). You can either create an anonymized URL or include an anonymized zip file.
    \end{itemize}

\item {\bf Crowdsourcing and research with human subjects}
    \item[] Question: For crowdsourcing experiments and research with human subjects, does the paper include the full text of instructions given to participants and screenshots, if applicable, as well as details about compensation (if any)?
    \item[] Answer: \answerNA{}
    \item[] Justification: The paper does not involve crowdsourcing or research with human subjects.
    \item[] Guidelines:
    \begin{itemize}
        \item The answer \answerNA{} means that the paper does not involve crowdsourcing nor research with human subjects.
        \item Including this information in the supplemental material is fine, but if the main contribution of the paper involves human subjects, then as much detail as possible should be included in the main paper.
        \item According to the NeurIPS Code of Ethics, workers involved in data collection, curation, or other labor should be paid at least the minimum wage in the country of the data collector.
    \end{itemize}

\item {\bf Institutional review board (IRB) approvals or equivalent for research with human subjects}
    \item[] Question: Does the paper describe potential risks incurred by study participants, whether such risks were disclosed to the subjects, and whether Institutional Review Board (IRB) approvals (or an equivalent approval/review based on the requirements of your country or institution) were obtained?
    \item[] Answer: \answerNA{}
    \item[] Justification: The paper does not involve research with human subjects.
    \item[] Guidelines:
    \begin{itemize}
        \item The answer \answerNA{} means that the paper does not involve crowdsourcing nor research with human subjects.
        \item Depending on the country in which research is conducted, IRB approval (or equivalent) may be required for any human subjects research. If you obtained IRB approval, you should clearly state this in the paper.
        \item We recognize that the procedures for this may vary significantly between institutions and locations, and we expect authors to adhere to the NeurIPS Code of Ethics and the guidelines for their institution.
        \item For initial submissions, do not include any information that would break anonymity (if applicable), such as the institution conducting the review.
    \end{itemize}

\item {\bf Declaration of LLM usage}
    \item[] Question: Does the paper describe the usage of LLMs if it is an important, original, or non-standard component of the core methods in this research? Note that if the LLM is used only for writing, editing, or formatting purposes and does \emph{not} impact the core methodology, scientific rigor, or originality of the research, declaration is not required.
    \item[] Answer: \answerYes{}
    \item[] Justification: Section~\ref{sec:alignment} describes LLM use in SOLAR. Appendices~\ref{appendix:datasets} and~\ref{appendix:baselines} document upstream data translation and MAPO scoring, respectively.
    \item[] Guidelines:
    \begin{itemize}
        \item The answer \answerNA{} means that the core method development in this research does not involve LLMs as any important, original, or non-standard components.
        \item Please refer to our LLM policy in the NeurIPS handbook for what should or should not be described.
    \end{itemize}

\end{enumerate}